%% file: main.tex
\documentclass{ecai} 

\usepackage{times}
\usepackage{soul}
\usepackage{url}
\usepackage[hidelinks]{hyperref}
\usepackage[utf8]{inputenc}
\usepackage[small]{caption}
\usepackage{graphicx}
\usepackage{amsmath,mathtools}
\usepackage{amsthm}
\usepackage{amsfonts}
\usepackage{booktabs}
\usepackage{algorithm}
\usepackage{algorithmic}
\usepackage[switch]{lineno}
\usepackage{enumitem}
\usepackage{bm}
\usepackage{makecell}
\usepackage{float}
\usepackage{siunitx}
\usepackage{tikz}
\usepackage{cleveref}

\usepackage{xcolor}
\definecolor{firebrick}{RGB}{162,49,42}
\definecolor{darkorange}{RGB}{254,140,2}
\definecolor{saddlebrown}{RGB}{129,73,32}

\begin{document}

\begin{frontmatter}




\title{stl2vec: Semantic and Interpretable Vector Representation of Temporal Logic}


\author[A, B]{\fnms{Gaia}~\snm{Saveri}}
\author[A]{\fnms{Laura}~\snm{Nenzi}}
\author[A]{\fnms{Luca}~\snm{Bortolussi}}
\author[C]{\fnms{Jan}~\snm{Křetínský}}

\address[A]{University of Trieste, Italy}
\address[B]{University of Pisa, Italy}
\address[C]{Masaryk University of Brno, Czech Republic}




\begin{abstract}
Integrating symbolic knowledge and data-driven learning algorithms is a longstanding challenge in Artificial Intelligence. Despite the recognized importance of this task, a notable gap exists due to the discreteness of symbolic representations and the continuous nature of machine-learning computations. One of the desired bridges between these two worlds would be to define semantically grounded vector representation (feature embedding) of logic formulae, thus enabling to perform continuous learning and optimization in the semantic space of formulae. We tackle this goal for knowledge expressed in \emph{Signal Temporal Logic (STL)} and devise a method to compute continuous embeddings of formulae with several desirable properties: the embedding (i) is finite-dimensional, (ii) faithfully reflects the semantics of the formulae, (iii) does not require any learning but instead is defined from basic principles, (iv) is \emph{interpretable}. 
Another significant contribution lies in demonstrating the efficacy of the approach in two tasks: learning model checking, where we predict the probability of requirements being satisfied in stochastic processes; and integrating the embeddings into a neuro-symbolic framework, to constrain the output of a deep-learning generative model to comply to a given logical specification.
\end{abstract}

\end{frontmatter}

\section{Introduction}\label{sec:intro}
The need for integrating Artificial Intelligence (AI) and symbolic (i.e.\ logical) knowledge has been claimed for a long time \cite{mccarthy}, with logic being closely related to the way in which humans represent knowledge and reasoning \cite{logic-think}.  However, a remarkable gap burdens on the integration of Machine Learning (ML) algorithms and symbolic representations: the latter are discrete objects, while ML models typically work in continuous domains. In this context, Neuro-Symbolic AI (NeSy) is emerging as a paradigm for the principled integration of sub-symbolic connectionist systems and logic knowledge \cite{nesy-survey}. As an example, NeSy models might address the following: leveraging logic knowledge for aiding the ML system improve its performance and/or learn with less data, using background knowledge expressed in symbolic form to constraint the behaviour of the ML system \cite{dl-constraints}.

\emph{Temporal logic} is a formalism suitable and since \cite{temporal-logic} widely used for describing properties and requirements of time-series related task, in particular of \emph{dynamical systems}. Here, we specifically consider 
stochastic processes, such as epidemiological models or cyber-physical systems, where \emph{Signal Temporal Logic (STL)} \cite{stl} emerges as the de-facto standard language, being concise yet rich and expressive for stating specifications of systems evolving over time \cite{stl-cps}. For example in STL one can state properties like "the temperature of the room will reach $25$ degrees within the next $10$ minutes and will stay above $22$ degrees for the next hour". In this area, one is typically interested in understanding or verifying which properties the system under analysis is compliant to (or more precisely, in the probability of observing behaviour satisfying the property). Such analysis is often tackled by  formal methods, via algorithms belonging to the world of quantitative model checking \cite{model-checking}.

\textbf{In this work, we address the challenge of incorporating knowledge in the form of temporal logic formulae inside data-driven learning algorithms.} The key step is to devise a \emph{finite-dimensional} embedding (feature mapping) of logical formulae into \emph{continuous space}, yielding their representation as vectors of real numbers. In this way, symbolic knowledge can be seamlessly integrated into distance-based or neural-based architectures, and eventually doors are opened towards gradient-based optimization techniques. To make these techniques truly effective, we additionally require that semantically similar formulae are mapped to nearby representations. We call such embeddings \emph{semantic}, allowing the efficient continuous optimization to happen in the ``semantic'' feature space of formulae.

\paragraph{Our contribution} consists in formulating a way for computing such \emph{finite-dimensional continuous semantic} embeddings of formulae of STL that are \emph{interpretable}, and proving their effectiveness in integrating logical knowledge and machine-learning algorithms. In detail, we make the following contributions:
\begin{enumerate}[label=(\roman*)]
\item We construct \emph{finite-dimensional} semantic embeddings of STL formulae starting from the kernel defined in \cite{stl-kernel}: kernel methods are indeed suitable in this context, since they efficiently allow to implicitly define a rich feature space, without the need of manually constructing it. Kernel PCA \cite{kpca} then allows us to construct suitable finite-dimensional approximations;
\item We give an \emph{interpretable description} of the geometry of such embeddings, up to a certain quantified extent, differently from state-of-art logical embedding methods. Notably, the embeddings are not learnt but defined from basic principles, and, as we show, the characterization is resilient w.r.t. the parameters of the embedding construction method, indicating the revealed structure is inherent to the logic. The extracted features foster human-understandability of the formulae representation, and thus also of the optimization;
\item We prove that the computed representations meaningfully capture the \emph{semantic} similarity of formulae, by using our finite-dimensional logical embeddings for \emph{learning model checking}, i.e.\ for predicting the probability of a given requirement being satisfied by a stochastic process, given a set of observed properties with their probabilities;
\item We demonstrate the efficacy of the representations in preserving the semantic information carried by the formulae by using them as semantic \emph{conditioning inside a NeSy deep generative framework}. We show that this improves the deep-learning process and model, critically relying in the form of our embeddings.
\end{enumerate}

\section{Preliminaries}\label{sec:background}

\paragraph{Kernel methods} \cite{rasmussen-kernel} are machine learning algorithms leveraging a positive semi-definite kernel function $k$ to map input datapoints, e.g. vectors in $\mathbb{R}^m$,  to a feature space $\mathbb{R}^D$, usually of higher dimension, i.e.\ $D\gg m$. Let $\Phi: \mathbb{R}^m\rightarrow \mathbb{R}^D$ denote this feature map, a key characteristic of kernel functions is that $\Phi$ is not explicitly calculated, but instead it is implicitly defined by computing its inner product in $\mathbb{R}^D$, formally $k: \mathbb{R}^m \times \mathbb{R}^m \rightarrow \mathbb{R}$ such that $k(\bm{x}_i, \bm{x}_j) = \langle \Phi(\bm{x}_i), \Phi(\bm{x}_j) \rangle$. The kernel trick hence allows to perform learning tasks in a feature space of higher dimension without explicitly constructing it, enabling the encoding of nonlinear manifolds without knowing the explicit feature maps, with a computational cost independent of the amount of features but only on the number of training points. 

\paragraph{Kernel Principal Component Analysis (PCA)} is a nonlinear dimensionality reduction technique that involves performing PCA \cite{pca} in the manifold identified by a kernel function. We recall that given a dataset with points described in $\mathbb{R}^D$ and an integer number $d \ll D$,  PCA consists in finding the set of $d$ orthogonal directions, called Principal Components (PC), preserving the highest amount of information (i.e.\ variance) of the original dataset, and projecting the datapoints along these vectors, reducing their dimension. In kernel PCA, such directions are provably the eigenvectors of the centered kernel matrix of the dataset, corresponding to its $d$ highest eigenvalues. 

\paragraph{Signal Temporal Logic (STL)} is a linear-time temporal logic which expresses properties on trajectories over dense time intervals \cite{stl}. We define as trajectories the functions $\xi: I\rightarrow D$, where $I\subseteq \mathbb{R}_{\geq 0}$ is the time domain and $D\subseteq \mathbb{R}^k, k\in \mathbb{N}$  is the state space.  The syntax of STL is given by:
$$\varphi:=tt\mid\pi\mid\lnot\varphi\mid \varphi_1\land\varphi_2\mid\varphi_1\mathbf{U}_{[a, b]}\varphi_2$$
where $tt$ is the Boolean \emph{true} constant; $\pi$ is an \emph{atomic predicate}, i.e.\ a function over variables $\bm{x}\in \mathbb{R}^n$ of the form $f_{\pi}(\bm{x})\geq \num{0}$ (we refer to $n$ as the number of variables of a STL formula); $\lnot$ and $\land$ are the Boolean \emph{negation} and \emph{conjunction}, respectively (from which the \emph{disjunction} $\lor$ follows by De Morgan's law); $\mathbf{U}_{[a, b]}$, with $a, b \in \mathbb{Q}, a<b$, is the \emph{until} operator, from which the \emph{eventually} $\mathbf{F}_{[a, b]}$ and the \emph{always} $\mathbf{G}_{[a, b]}$ temporal operators can be deduced. We call $\mathcal{P}$ the set of well-formed STL formulae. STL is endowed with both a \emph{qualitative} (or Boolean) semantics, giving the classical notion of satisfaction of a property over a trajectory, i.e.\ $s(\varphi, \xi, t) = \num{1}$ if the trajectory $\xi$ at time $t$ satisfies the STL formula $\varphi$ , and a  \emph{quantitative} semantics, denoted by $\rho(\varphi, \xi, t)$. The latter, also called \emph{robustness}, is a measure of how robust is the satisfaction of $\varphi$ w.r.t. perturbations of the signals. Robustness is recursively defined as:
\begin{small}
\begin{align*}
 & \rho(\pi,\xi,t) &=& f_\pi(\xi(t)) \qquad \text{for } \pi(\bm{x})=\big(f_\pi(\bm{x})\geq 0\big)\\
 & \rho(\lnot\varphi,\xi,t) &=& -\rho(\varphi,\xi,t)\\
  & \rho(\varphi_1\land\varphi_2,\xi,t) &=& \min\big(\rho(\varphi_1,\xi,t),  \rho(\varphi_2,\xi,t)\big)\\
 &\rho(\varphi_1\mathbf{U}_{[a, b]}\varphi_2,\xi,t) \hspace*{-0.5em}&=&  \max_{\!\!\mathmakebox[3em][c]{t'\in[t+a,t+b]}}\big(\min\big(\rho(\varphi_2,\xi,t'), 
 \min_{\!\!\mathmakebox[1em][c]{t''\in[t,t']}}\rho(\varphi_1,\xi,t'')\big)\big)
\end{align*}
\end{small}
Robustness is compatible with satisfaction via the following \emph{soundness} property: if $\rho(\varphi, \xi, t) > 0$ then $s(\varphi, \xi, t) = \num{1}$ and if $\rho(\varphi, \xi, t) < \num{0}$ then $s(\varphi, \xi, t) = \num{0}$. When $\rho(\varphi, \xi, t) = \num{0}$ arbitrary small perturbations of the signal might lead to changes in satisfaction value. For numerical stability reasons, we use a normalized robustness, rescaling the output signals using a sigmoid function, see Appendix \ref{app:sec:background}. When we evaluate properties at time $t=\num{0}$, we omit $t$ from the previous notations. A distribution $\mathcal{F}$ over STL formulae can be algorithmically defined by a syntax-tree random recursive growing scheme, that recursively generates the nodes of a formula given the probability $p_{\mathit{leaf}}$ of each node being an atomic predicate, and a uniform distribution over the other operator nodes.  

\paragraph{Stochastic Processes} within this context are probability spaces defined as triplets $\mathcal{M} = (\mathcal{T}, \mathcal{A}, \mu)$ of a trajectory space $\mathcal{T}$ and a probability measure $\mu$ on a $\sigma$-algebra $\mathcal{A}$ over $\mathcal{T}$.  Given a stochastic process $\mathcal{M}$, the \emph{expected robustness} is a function $R_{\mathcal{M}}: \mathcal{P}
\times I \rightarrow \mathbb{R}$ such that $R_{\mathcal{M}}(\varphi, t) = \mathbb{E}_{\mathcal{M}}[\rho(\varphi, \xi, t)] = \int_{\xi \in \mathcal{T}} \rho(\varphi, \xi, t)d\mu(\xi)$. Similarly, the \emph{satisfaction probability} $S_{\mathcal{M}}: \mathcal{P}
\times I \rightarrow \mathbb{R}$ is computed as $S_{\mathcal{M}}(\varphi, t) = \mathbb{E}_{\mathcal{M}}[s(\varphi, \xi, t)] = \int_{\xi \in \mathcal{T}} s(\varphi, \xi, t)d\mu(\xi)$. In probabilistic and statistical model checking, one is often interested in computing or estimating these quantities, see \cite{model-checking} for details. In this work we consider stochastic processes that can be simulated via the Gillespie Stochastic Simulation Algorithm (SSA) \cite{ssa}, which samples from the exact distribution $\mu$ over trajectories.

\paragraph{A kernel function for STL formulae} is defined in \cite{stl-kernel} by leveraging the quantitative semantics of STL. Indeed, robustness allows formulae to be considered as functionals mapping trajectories into real numbers, i.e.\ $\rho(\varphi,\cdot): \mathcal{T}\rightarrow \mathbb{R}$ such that $\xi\mapsto \rho(\varphi, \xi)$. Considering these as feature maps, and fixing a probability measure $\mu_0$ on $\mathcal{T}$, a kernel function capturing similarity among STL formulae on mentioned feature representations can be defined as:
\begin{equation}
\resizebox{.9\linewidth}{!}{$
k(\varphi, \psi) = \langle \rho(\varphi, \cdot), \rho(\psi, \cdot) \rangle = \int_{\xi\in \mathcal{T}} \rho(\varphi, \xi) \rho(\psi, \xi) d\mu_0(\xi)
$}
\label{eq:stl-kernel}
\end{equation}
opening the doors to the use of the scalar product in the Hilbert space $L^2$ as a kernel for $\mathcal{P}$; intuitively this results in a kernel having high positive value for formulae that behave similarly on high-probability trajectories (w.r.t. $\mu_0$), and viceversa low negative value for formulae that on those trajectories disagree. For what concerns the measure $\mu_0$ on $\mathcal{T}$, it is designed in such a way that simple signals are more probable, considering total variation and number of changes in the monotonicity as metrics for measuring the complexity of trajectories, we refer to \cite{stl-kernel} for full details. 

Note that, although the feature space $\mathbb{R}^{\mathcal{T}}$ (which we call the \emph{latent semantic space}) into which $\rho$ (and thus Equation (\ref{eq:stl-kernel})) maps formulae is infinite-dimensional, in practice the kernel trick allows to circumvent this issue. It does so by mapping each formula to a vector of dimension equal to the number of formulae which are in the training set used to evaluate the kernel (Gram) matrix. Such embeddings are continuous representations of discrete symbolic objects, and can be used to solve learning tasks such as predicting the expected robustness and the satisfaction probability of a stochastic process via continuous optimization-based ML algorithms.  

\begin{figure}[t]
\begin{minipage}[b]{.26\linewidth}
    \centering
    \vspace{-0.1cm}
     \resizebox{1.2\linewidth}{!}{
        \begin{tabular}{c|c|c}
        \hline
        \# var & $\tau$ & $\tau$\\
        \ {} & $\num{0.95}$ & $\num{0.98}$ \\
        \hline
        3 &  10 & 13\\
        4 &  11 & 16\\
        5 &  14 & 19\\
        6 &  16 & 22\\
        7 &  18 & 25\\
        8 &  20 & 28\\
        9 &  22 & 31\\
        10 & 24 & 35\\
        \hline
        \end{tabular}
    }
    \vspace{0.15cm}
    \captionof{table}{\# PC for achieving $\mathcal{X}_d$ higher than $\num{95}\%$ (resp. $\num{98}\%$), increasing the number of variables.}
    \label{tab:xai-variance}
    \vspace{0.5cm}
\end{minipage}\hfill
\begin{minipage}[b]{.68\linewidth}
        \vspace{-0.4cm}
        \includegraphics[width=1\linewidth]{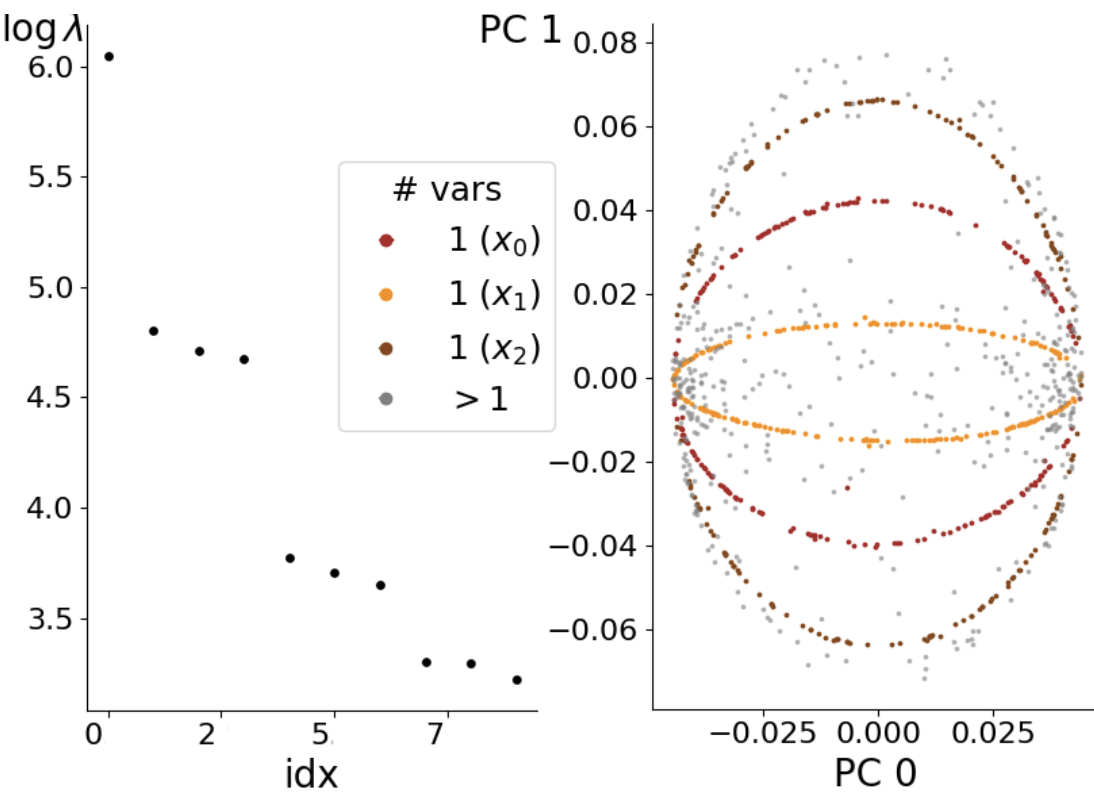}
        \vspace{0.05cm}
        \captionof{figure}{For a set of STL formulae with 3 variables (left) spectrum of the covariance matrix of its Gram matrix; (right) $1^{st}$ vs $2^{nd}$ PC, showing formulae with only one var. }
        \label{fig:spectrum}
        \vspace{0.5cm}
\end{minipage}
\end{figure}

\section{stl2vec}\label{sec:stl2vec}
We are interested in \textbf{``semantic'' embeddings}: intuitively, mapping formulae with similar semantics to nearby vectors; formally, given that the robustness $\rho$ captures the considered semantics in the infinite-dimensional latent semantic space $\mathbb R^{\mathcal T}$, the new embeddings should (approximately) preserve the distances induced by the kernel in Equation~(\ref{eq:stl-kernel}), and thus should essentially be $\rho$'s ``almost continuous'' projections.
In this work, we (i) provide an algorithmic procedure, called \emph{stl2vec}, to \emph{construct} explicit finite-dimensional  semantic embeddings of STL formulae (Section \ref{subsec:build}), (ii) explore the geometry of such representations, producing \emph{human-interpretable} explanations to a vast amount of information retained by the new representation (Section \ref{subsec:explain}), and (iii) show the effectiveness of the embeddings in integrating temporal logic knowledge inside data-driven learning algorithm (Section \ref{sec:applications}).
We remark that the explainability provides more control over producing continuous STL formulae embeddings.  
Finally, we also recall that creating finite-dimensional representations is a crucial step to make data more manageable (reducing the risk of incurring in the so-called curse of dimensionality), and help to eliminate noise and redundant information.

\subsection{Building Explicit STL Embeddings}\label{subsec:build}

The starting point of our investigation are kernel embeddings for STL formulae as defined in Section \ref{sec:background}. All reported results in this Section, unless differently specified, are obtained by keeping the default parameters used in \cite{stl-kernel}; later in the manuscript we will also report ablation studies to enforce our statements. Hence, starting from implicit infinite-dimensional embeddings constructed via Equation (\ref{eq:stl-kernel}), we derive explicit finite-dimensional numerical representations of STL formulae using kernel PCA. As we will highlight in the remainder of the paper, this transformation gives us a deep insight into the geometry of these representations, to the point of making us able to give explanations for the vast majority of information captured by the embeddings. 

In detail, the algorithm stl2vec proceeds as follows: given a fixed set of $D$ STL formulae (that we call \emph{training set}) and an integer $d \ll D$ representing the reduced dimension of the embeddings, we obtain the coordinates of the reduced dimensional space by performing the eigenvalue decomposition of the centered kernel matrix of the training set (which is $D$-dimensional) and retaining the top-$d$ eigenvectors (i.e.\ PC), which are those corresponding to the $d$ largest eigenvalues. These PC will be used to project the data into a lower-dimensional subspace. We remark that this procedure does not require any learning.

In practical applications, given the set of eigenvalues of the kernel matrix of the data  $\{\lambda_k\}_{k=1}^D$  (sorted in descending order) to select the number $d$ of dimensions to retain, it is common to look at the so-called \emph{proportion of variance explained}:  $\mathcal{X}_d = \frac{\sum_{i=1}^d \lambda_i}{\sum_{j=1}^D \lambda_j}$, choosing the smallest $d$ for which $\mathcal{X}_d\geq \tau$, for some threshold $\tau\in [\num{0}, \num{1}]$. Notably, for STL kernel embeddings built from a training set of $D=\num{1000}$ random formulae, only a few tens of components are necessary to explain more than $\num{95}\%$ of the variability in the data, as reported in Table \ref{tab:xai-variance}. 
Moreover, in Figure \ref{fig:spectrum} (left) we plot the log-spectrum (first $\num{10}$ eigenvalues) of a dataset of $D=\num{1000}$ formulae with $\num{3}$ variables, corresponding to the $\num{95}\%$ of variance explained, as per Table \ref{tab:xai-variance}.

In order to experimentally prove the independence of the individuated PC on the set of training formulae used to compute the STL kernel, we compare the coordinates found when changing the training set. In detail we sample $\num{50}$ different training sets, coming from $5$ different distributions, obtained by changing the parameter $p_{\text{leaf}}$ of the formulae sampling algorithm $\mathcal{F}$ detailed in Section~\ref{sec:background}. We vary it in the set $[0.3, 0.35, 0.4, 0.45, 0.5]$ and sample $10$ datasets for each value, each composed of $D=\num{1000}$ STL formuale with $\num{3}$ variables. We then reduce their dimension to $d=\num{13}$ (hence retaining more than the $\num{98}\%$ of information, according to Table \ref{tab:xai-variance}). Results show that, up to permutation of coordinates, the \textbf{identified principal directions are almost the same across all datasets}. Indeed, if we compute the pairwise cosine similarity between corresponding PC of each possible pair of datasets, we get that, up to the $5^{th}$ PC, all datasets share a cosine similarity of at least $\num{0.95}$, moreover similarity stays above $\num{0.69}$ for all the $\num{13}$ considered components, with both mean and median similarity being $>\num{0.9}$ in every direction, for all possible pair of datasets, see also Appendix \ref{app:sec:stl2vec}. Hence the embeddings are robust w.r.t. the choice of training formulae, at least on their most significant components.

\begin{figure*}[t]
    \begin{minipage}{0.35\linewidth}
        \centering
        \vspace{-0.5cm}
        \includegraphics[width=0.7\linewidth, keepaspectratio]{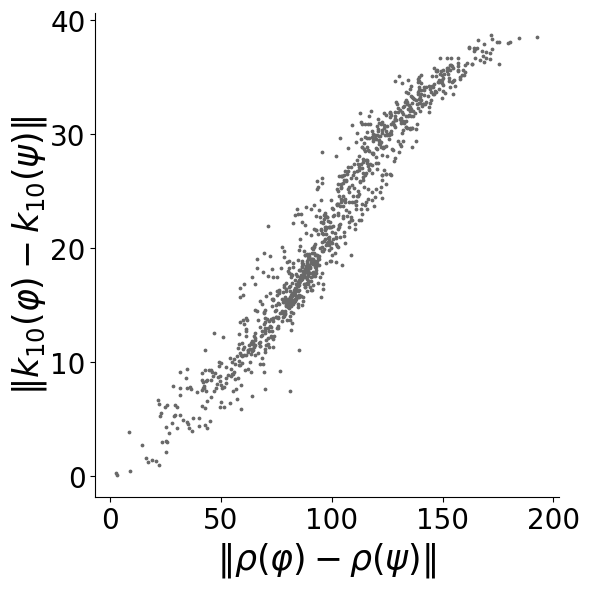}
        \vspace{0.05cm}
        \captionof{figure}{$L_2$ distance between $10$-dim embeddings of formulae vs $L_2$ distance among their respective robustness vectors.}
        \label{fig:semantic-similarity}
        \vspace{-0.8cm}
    \end{minipage}\hfill
    \begin{minipage}{0.6\linewidth}
        \centering
        \vspace{-0.3cm}
        \includegraphics[width=0.9\linewidth, keepaspectratio]{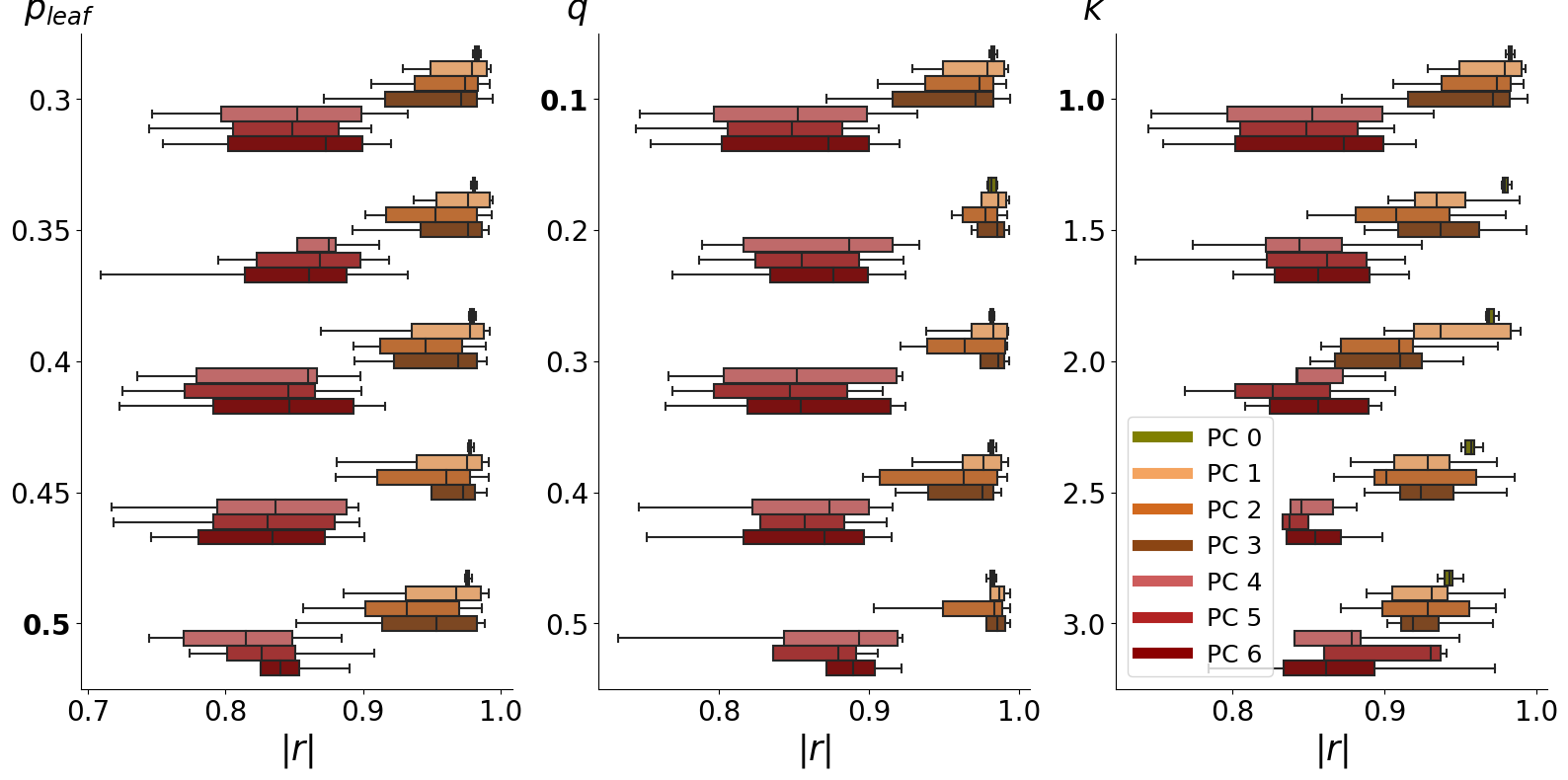}
        \vspace{0.05cm}
        \captionof{figure}{Resilience of the explanations of PC to changing of the parameters (from left to right) $p_{\mathit{leaf}}$, $q$ and $K$ in terms of absolute Pearson Correlation Coefficient ($r$). Bold labels represent default parameters.}
        \label{fig:ablations}
    \end{minipage}
\vspace{0.1cm}
\end{figure*}

Finally, we check that the embeddings are semantic, by assessing linear correlation between the distance among kernel PCA embeddings (with $d=\num{10}$) of each pair of formulae in the considered dataset, and the corresponding distance between robustness vectors, i.e.\ the vectors $\bm \rho(\varphi) = [\rho(\varphi, \xi_i)]_{i=0}^M$ of robustness of a STL formula $\varphi$ computed on $M$ (in our case $\num{10000}$) trajectories randomly sampled from $\mu_0$. The Pearson correlation coefficient among the two quantities is $\num{0.9688}$, and their correlation is graphically shown in Figure \ref{fig:semantic-similarity}; intuitively, formulae whose quantitative robustness agrees on a high number of trajectories are mapped nearby in the continuous space of their stl2vec embeddings.   

\noindent\fbox{\begin{minipage}[t]{0.47\textwidth}
In summary, (i) the principal directions of the embeddings are inherent to the STL robustness semantics (and thus it makes sense to try and \emph{explain} them), and (ii) our embeddings are also experimentally observed as semantic (and thus it makes sense to measure how well they \emph{approximate} the full semantic information as defined by robustness and reflected by the kernel).
We examine the former in Sec.~\ref{subsec:explain} and the latter in Sec.~\ref{subsec:appl-power}.
\end{minipage}}

It is worth noting that the STL kernel imposes a smoothing on the combinatorics of satisfiability, through the measure $\mu_0$, for which the semantics of formulae is captured w.r.t. the probability distribution over trajectories (i.e.\ trajectories are weighted in such a way that STL formulae which only differ on few complicated signals are essentially considered equivalent), hence all the geometrical properties of the STL embeddings presented are valid up to this statistical filter. Such filter can however be changed by using a custom measure on trajectories for computing the kernel (e.g. the data generating distribution of the problem at hand), and this adds another layer of flexibility to our methodology.

\subsection{Explaining Principal Directions}\label{subsec:explain} 

Having described how explicit embeddings for STL formulae are computed, and confirming their semantic character, we now delve into exploring the geometry of these representations. We substantiate our explanations by statistical evidence, namely strong correlations detailed in Appendix~\ref{app:subsec:ablations}.

Looking at the spectrum of the kernel matrix for formulae with $\num{3}$ variables in Figure \ref{fig:spectrum} (left) and recalling that clear gaps in the spectrum are an indication that dimensionality reduction including the components before the gap is meaningful, we immediately observe that, after a big gap between the first and the second eigenvalue, the spectrum is partitioned into groups of $\num{3}$ eigenvalues divided by gaps. This intuitively suggests that principal directions (apart from the first one) might encode properties that hold variable-wise, possibly denoting that different variables are mapped to different sub-manifolds in the latent semantic space. Following this intuition, and having in mind the way in which embeddings are computed (i.e.\ starting from Equation (\ref{eq:stl-kernel})), we are able to provide an interpretable explanation for the information carried by the first principal direction and the following two sets of components, each composed of as many values as the variables appearing in the formulae. In particular, we identify statistical properties based on the robustness of STL formulae which are linearly correlated with the PC. This is intuitively meaningful since the quantitative semantics of STL is the bridge used by the STL kernel for mapping discrete formulae into a continuous space. For this reason, we also believe that  further PC encode more refined properties related to the robustness profile of formulae, which we are not able to describe.

We stress that a clear interpretation of projections obtained by kernel PCA is far from trivial, as seen in \cite{interpet-kpca}. In this case, we work with objects and embeddings with a semantic nature, and this is reflected in the features captured by the PC, whose meaning is however not-immediate to assess. 

\paragraph{The first principal direction} PC$\num{0}$ describes the \fbox{median robustness} of each formula $\varphi$ over a random set of trajectories sampled from $\mu_0$. For the statistical evidence refer to Appendix \ref{app:subsec:ablations}.
Hence the first PC 
captures a descriptor of the satisfiability of a formula, which, from a statistical point of view, acts as the main source of variability of the robustness distribution, as computed by Equation (\ref{eq:stl-kernel}).


\paragraph{The second group of principal components} which is composed of $n$ coordinates, when considering formulae of $n$ variables, accounts instead for the \fbox{variability of the robustness} over $\mu_0$, being linearly correlated with the mean kernel similarity to formulae which exhibit high variance in robustness across signals sampled from $\mu_0$. 

In detail, the quantity which is linearly correlated with each direction belonging to this group can be computed via the following steps, given a test dataset $\mathcal{D}$ of STL formulae with $n$ variables: 
\begin{enumerate}[label=A.\arabic*]
    \item \label{a:first} Sample a random dataset $\mathcal{D}_i$ of STL formulae containing only variable $x_i$, with $i\in \mathbb{N}, 0\leq i < n$;
    \item \label{a:second} Sample an arbitrary number of trajectories $\hat{\mathcal{T}}$ from $\mu_0$;
 from the current trajectory distribution (e.g. $\mu_0$);
    \item \label{a:third} Evaluate the robustness vector $\bm\rho(\varphi_j) = \{\rho(\varphi_j, \xi)\}_{\xi\in \hat{\mathcal{T}}}$ of each formula $\varphi_j \in \mathcal{D}_i$ (on the selected trajectories);
    \item \label{a:fourth} Compute the standard deviation $\sigma_j = \text{std}(\bm\rho(\varphi_j))$ of the robustness vector of each formula $\varphi_j\in \mathcal{D}$; 
    \item \label{a:fifth} Select the indexes $j$ of each $\sigma_j$ corresponding to values above the $90^{th}$ percentile, to get a subset of formulae $\hat{\mathcal{D}}_i$;
    \item \label{a:sixth} Compute the vector of mean kernel similarity $\mathbf{\tilde{k}|x_i} = \big\{\frac{1}{|\hat{\mathcal{D}}_i|} \sum_{k=1}^{|\hat{\mathcal{D}}_i|} k(\varphi_j, \varphi_k)\big\}_{j=1}^{|\mathcal{D}|}$ between the formulae in $\mathcal{D}$ and the ones obtained by previous steps;
    \item \label{a:seventh} $\forall i, \mathbf{\tilde{k}|x_i}$ is then linearly correlated (see Appendix \ref{app:subsec:ablations}) with one of the PC having index in $[1, n]$. 
\end{enumerate}

To give an intuitive description of the behaviour of formulae in $\hat{\mathcal{D}}_i$ obtained as per steps \ref{a:first}-\ref{a:fifth}, we have experimentally verified that most of them are properties which are robustly satisfied and robustly unsatisfied on a comparable number of trajectories sampled from $\mu_0$. 
 
\paragraph{The third group of principal components} is composed of $n$ directions as well, when considering STL formulae with $n$ variables.
The information they carry represents the \fbox{importance of each variable} in determining the semantics/robustness of a formula, as it is directly proportional to the change in robustness when fixing the part of the signals involving the variable itself.
In particular, the quantity which describes each of these PC can be computed with the following steps, starting with a given a test dataset $\mathcal{D}$:
\begin{enumerate}[label=B.\arabic*]
    \item \label{b:first} Compute a set of $m$ random trajectories $\Xi = \{\xi_{k}\}_{k=1}^m$ on $n$ variables, according to the given distribution;
    \item \label{b:second} For each variable index $i\in \mathbb{N}, 0\leq i<n$, compute the set of trajectories $\Xi_i = \{\xi_{ik}\}_{k=1}^m$ by replacing the $i^{th}$ component of each signal in $\Xi$ with the constant $\bm 0$;
    \item \label{b:third} For each $\varphi\in \mathcal{D}$, compute the mean absolute difference $\{\tilde{\rho}_i(\varphi) = \frac{1}{m} \sum_{k=1}^m | \rho(\varphi, \xi_k) - \rho(\varphi, \xi_{ik})|\}_{i=0}^{n-1}$;
    \item \label{b:fourth} $\forall i, \tilde{\bm \rho}|\bm{x}_i = \{\tilde{\rho}_i(\varphi)\}_{\varphi\in \mathcal{D}}$ is then linearly correlated with one of the PC having index in $[1+ n,  2\cdot n]$.
\end{enumerate}

\begin{figure*}[t!]
\begin{minipage}{0.35\linewidth}
    \centering
    \includegraphics[width=\linewidth, keepaspectratio]{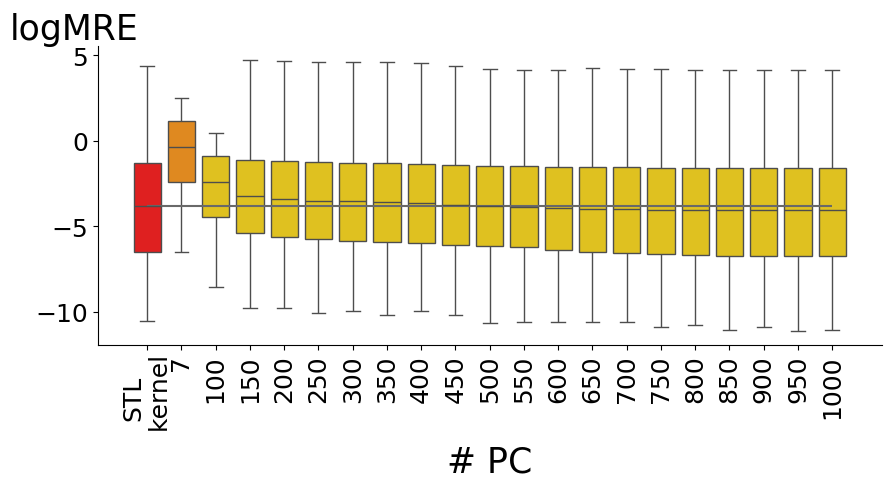}
    \captionof{figure}{Mean of the quantiles for $RE$ over $\num{100}$ regression experiments for predicting average robustness of trajectories sampled from the SIRS model, varying the number of retained PC.}
    \label{fig:regression-boxplot}
\end{minipage}\hfill
\begin{minipage}{0.62\linewidth}
\resizebox{\linewidth}{!}{
\centering
\begin{tabular}{llllll|llll}
\toprule
{} & {} & \multicolumn{4}{c}{relative error (RE) } & \multicolumn{4}{c}{absolute error (AE)} \\
\midrule
{} &  {} &  1quart &   median &   3quart &  99perc &  1quart &   median &   3quart &   99perc \\
\midrule
$\rho$ &  \makecell{ STL kernel \\ stl2vec($\num{250}$) \\ stl2vec($\num{500}$)} &   
\makecell{ 0.00772 \\ 0.01246 \\ 0.00917} &
\makecell{ 0.02582 \\ 0.03385 \\ 0.02532} &
\makecell{ 0.09225 \\ 0.10293 \\ 0.07942} &
\makecell{ 1.41988 \\ 1.26477 \\ 1.14463} &
\makecell{ 0.01362 \\ 0.02409 \\ 0.01769} &
\makecell{ 0.04376 \\ 0.06393 \\ 0.04689} &
\makecell{ 0.14283 \\ 0.17317 \\ 0.13455} &
\makecell{ 0.92352 \\ 0.83707 \\ 0.79238}
\\

\hline 

$R$ &  \makecell{ STL kernel \\ stl2vec($\num{250}$) \\ stl2vec($\num{500}$)} & 
\makecell{ 0.00629 \\ 0.01162 \\ 0.00822} &
\makecell{ 0.02209 \\ 0.03026 \\ 0.02235} &
\makecell{ 0.07593 \\ 0.08979 \\ 0.06859} &
\makecell{ 1.19013 \\ 1.28718 \\ 1.0287} &
\makecell{ 0.00608 \\ 0.01129 \\ 0.00797} &
\makecell{ 0.02052 \\ 0.02868 \\ 0.021} &
\makecell{ 0.06493 \\ 0.07669 \\ 0.05864} &
\makecell{ 0.43494 \\ 0.38096 \\ 0.34801}  \\

\hline

$S$ &  \makecell{ STL kernel \\ stl2vec($\num{250}$) \\ stl2vec($\num{500}$)} & 
\makecell{ 0.00209 \\ 0.00255 \\ 0.00212} &
\makecell{ 0.02762 \\ 0.03235 \\ 0.02821} &
\makecell{ 0.85246 \\ 1.18237 \\ 0.87827} &
\makecell{ 3.8337 \\ 4.35775 \\ 3.81897} &
\makecell{ 0.00782 \\ 0.01161 \\ 0.00825} &
\makecell{ 0.02634 \\ 0.03256 \\ 0.02679} &
\makecell{ 0.08807 \\ 0.09893 \\ 0.08823} &
\makecell{ 0.60182 \\ 0.62148 \\ 0.60294} \\
\bottomrule
\end{tabular}
}
\vspace*{0.1cm}
\captionof{table}{Mean of quantiles for RE and AE over $\num{100}$ experiments for prediction of robustness on single trajectory $\rho$ (top), average robustness $R$ (middle) and satisfaction probability $S$ (bottom), for a dataset of trajectories sampled from the SIRS model.}
\label{tab:quantile-regression}
\end{minipage}
\vspace{0.1cm}
\end{figure*}

\paragraph{An intuitive understanding of the explanations} can be given by considering simple requirements. If we take for example the following formulae of $1$-variable: $G(x_0 \geq 0) \wedge F(x_0 < 0)$ and $G(x_0\geq 0) \vee F(x_0\leq 0)$ then we immediately recognise that they are a contradiction and a tautology, respectively. This is indeed reflected in the first two components of their embeddings, which are [\textcolor{orange}{$-0.06357$}, \textcolor{cyan}{$0.0025$}] and [\textcolor{orange}{$0.0593$}, \textcolor{cyan}{$0.0058$}], i.e. for both the second component is small, witnessing a little variability of their robustness across trajectories, while the first is high (positive) for the tautology and low (negative) for the contradiction (as shown in Figure \ref{fig:spectrum} (right) the reference range of PC$0$ is $\pm 0.07$ and of $\pm 0.08$ for PC$1$ ). If we now take a slightly more complex formula in $2$ variables, namely $\varphi = (G (x_0\geq 0)) \wedge (G(x_1 \geq 0) \wedge F(x_1 < 0))$, then we recognize that it is a contradiction and that the most evident reason guiding our intuition only involves variable $x_1$, being the right conjuct of $\varphi$ a contradiction in which only $x_1$ appears. The explainable components of $\varphi$ are: [\textcolor{orange}{$-0.03219$}, \textcolor{cyan}{$-0.0272, -0.0018$}, \textcolor{magenta}{$0.0165, -0.4901$}], which lead to the following observations: a high negative value (w.r.t. above mentioned ranges) for the \textcolor{orange}{first component} together with a small value for a component belonging to the \textcolor{cyan}{second group} suggests that the formula is a contradiction, finally the fact that in the \textcolor{magenta}{third group} a component is small and positive, while the other is negative and an order of magnitude higher indicates that most of the semantic of $\varphi$ only depends on a specific variable. These examples help in getting a sense of both the intuitive meaning of the explained components, and of their usefulness in grasping the semantic of a formula when the formula is too big to be understood just visually inspecting it, or when only its embedding is available (e.g. when it is the outcome of an optimization procedure). 
 
\paragraph{Explanations of principal components are resilient} to the measure considered in the space of trajectories. 
Our reference measure $\mu_0$ (that is shown to be rather general in \cite{stl-kernel}) samples from piece-wise linear functions in the interval $\mathcal{I} = [a, b]$ by: setting the number of discretization points in the trajectory and sampling the initial point from $\mathcal{N}(0, 1)$; sampling the total variation of the trajectory $\mathit{tv} = (\mathcal{N}(0, K))^2$; sampling the local variation between each pair of consecutive points uniformly in $[0, \mathit{tv}]$ and for each such a point changing the sign of the derivative (i.e.\ the monotonicity) with probability $q$. Finally consecutive points of the discretization are linearly interpolated to make the signal continuous. Hence $\mu_0$ has the following parameters which can be tuned in order to significantly change the probability space of trajectories: (i) the mean $q$ of the Bernoulli distribution governing the number of changes in the monotonicity of each signal and (ii) the standard deviation $K$ of the Gaussian distribution from which the total variation of each trajectory is sampled. 

We test the stability of our explanations by measuring the Pearson correlation coefficient $r$ between the PC and the corresponding statistical quantities that we argue are their interpretation. For what concerns $\mu_0$, by increasing $q$ we are considering signals with an increasing number of changes in monotonocity, while by increasing $K$ we are testing trajectories with larger total variation. Besides, considering the formulae distribution $\mathcal{F}$ (see Section \ref{sec:background}), decreasing the parameter $p_{\mathit{leaf}}$ increases the syntactic complexity of formulae. In Figure \ref{fig:ablations} we show the quantiles of the distribution of the absolute linear correlation coefficient $|r|$ between the PC and our explanations, across $50$ independent datasets of STL formulae, in all the described ablation studies, verifying that it remains high in all settings, hence establishing the resilience of our interpretations. 

Moreover, we verify the stability of the explanations by changing the number $n$ of variables in formulae from $\num{3}$ to $\num{10}$: denoting the median absolute correlation coefficient $|r|$ as $\eta_{|r|}$, we have $\eta_{|r|} > \num{0.97}$ for the first PC, $\eta_{|r|} > \num{0.84}$ for the second group of PC and $\eta_{|r|} > \num{0.8}$ for the third group of PC, again proving resilience of the explanations. We remark here that, according to Table \ref{tab:xai-variance}, when the number of variables is higher then $\num{5}$ we are providing an interpretation for more than the $\num{95}\%$ of the variance in the data. Additional results and plots are reported in Appendix \ref{app:subsec:ablations}.

Finally, we test the stability of our explanations when replacing $\mu_0$ with another stochastic process, namely the SIRS epidemiological model \cite{sirs}: for the first component the median correlation is $\eta_{|r|} = \num{0.98}$, for the second group of PC $\eta_{|r|} > \num{0.53}$ and for the third group  $\eta_{|r|} > \num{0.57}$, showing moderate linear correlation, hence resilience of the explanations also for a completely different trajectory distribution. 

Interestingly, if we plot PC$\num{0}$ against PC belonging to the second group we are not only  able to individuate formulae in which only a variable appears, but also identify the involved variable (i.e. its index), as reported in Figure \ref{fig:spectrum} (right).  Intuitively, this might depend on: (i) the fact that the explanations for the second group of components hold variable-wise (suggesting that different variables are mapped to different semantic subspaces) and (ii)  the significant amount of information carried by PC$0$,  observable from the gap after PC$0$ in Figure \ref{fig:spectrum} (left). A similar behaviour is observed when considering PC belonging to the third group, as reported in Appendix \ref{app:sec:stl2vec}.
From the same plot it is possible to observe a quadratic relation between PC$\num{0}$ and PC belonging to the second group (PC$\num{1}$ in the picture). Although a clear explanation for this phenomenon is still lacking, we can interpret the behavior of formulae mapped to the extreme points of the three ellipsis: PC$\num{0}\approx \num{0}$ denotes formulae which neither robustly satisfy nor robustly unsatisfy any trajectory, or which robustly satisfy and unsatisfy a comparable number of trajectories, hence they are likely to have a highly variable robustness vector, explaining the fact that the (absolute) value for the second group of PC is high; viceversa, a formula whose variability is $\approx \num{0}$, for the opposite reason, is expected to have a high absolute median robustness value. 

\section{Applications}\label{sec:applications}

We claim and experimentally prove the high semantic expressiveness and the practical usefulness of stl2vec embeddings in two different scenarios: predicting average robustness and satisfaction probability of properties in a stochastic process (as defined in Section \ref{sec:background}) and semantically conditioning a deep learning generative model for the generation of trajectories compliant to arbitrary temporal properties.


\subsection{Predictive Power of Explicit Embeddings}
\label{subsec:appl-power}
In this suite of experiments, we use the embeddings of STL formulae as input for ridge regression in order to predict: robustness of formulae $\varphi \in \mathcal{F}$ on single trajectories $\xi\in \mathcal{T}$, i.e.\ the function $\rho:\varphi\mapsto\rho(\varphi,\xi)$; expected robustness $\mathbb E_{\xi\sim\mu_0} [\rho(\varphi,\xi)]$ and satisfaction probability $\mathbb E_{\xi\sim\mu_0} [s(\varphi,\xi)]$ of formulae $\varphi\in \mathcal{F}$, proxied by the experimental averages on a stochastic system $\{\xi_j\in \mathcal{T}\}_{j=1}^m$, i.e.\ respectively $R:\varphi\mapsto \frac{\sum_j \rho(\varphi, \xi_j)}{m}$ and $S:\varphi\mapsto \frac{\sum_j s(\varphi, \xi_j)}{m}$. We fix $\mu_0$ with its default parameters as the base measure on the space of trajectories (i.e. we use it for computing the kernel). We quantify the errors in terms both of Relative Error (RE) and Absolute Error (AE), and unless differently specified, we average results over $\num{100}$ independent experiments. We denote as stl$2$vec($d$) the embeddings obtained with our methodology, keeping the first $d$ PC. We perform the above mentioned model checking task on different scenarios: still considering $\mu_0$ as $\mathcal{T}$, but varying the dimensionality of signals; changing $\mathcal{T}$ considering trajectories coming from other stochastic processes, namely the SIRS epidemiological model ($3$-dim) and three other stochastic models (used as benchmarks also in \cite{stl-kernel}) simulated using the Python library StochPy \cite{stochpy} which are called Immigration ($1$-dim), Isomerization ($2$-dim) and Transcription ($3$-dim). We stress that in all the test cases, the STL kernel (hence the embeddings) is computed according to the base measure $\mu_0$.

As reported in Table \ref{tab:quantile-regression}, for a dataset of $D=\num{1000}$ STL formulae tested on trajectories sampled from the SIRS model, stl2vec embeddings of $\num{500}$ components, i.e.\ half the original size, achieve results comparable to those of full STL kernel ridge regression. Moreover, even if we keep just $\num{250}$ components, the predictive performance of the embeddings still is acceptable (median relative error $<\num{6}\%$ when predicting $\rho$, $<\num{1}\%$ when predicting $R$ and $<\num{2}\%$ for $S$). Interestingly, as shown in Figure \ref{fig:regression-boxplot}, where we compare against standard kernel regression monitoring performance changes as the number of retained PC is varied, the quality of predictions in terms of both errors improves until the dimensionality of the representations is $\leq \num{300}$, then it stabilizes to values comparable to those of full STL kernel ridge regression (whose quantiles are reported in red in the figure). 
In the same figure, we highlight with an orange box the errors reported when doing regression just with the components that we are able to explain ($\num{7}$ in this case, since we are working with a dataset of $\num{3}$ variables), hence in a scenario in which ridge regression can be fully interpreted. 
For what concerns the Immigration, Isomerization and Transcription models, under the same experimental assumptions, as well as experiments done on $10$-dimensional signals sampled from $\mu_0$, results in terms of median RE are reported in Table \ref{tab:pred-power-summary}. In all cases, we observe that the difference in performance between full and reduced embeddings is limited: using stl$2$vec($500$) instead of vanilla STL kernel brings at most $0.01\%$ of additional error, while using stl$2$vec($250$) brings a performance drop of at most $2.5\%$. In general, we can observe that results of these experiments are good: in all cases, the error when predicting $\rho$ is $<3.5\%$, it is $< 1.4\%$ when estimating $R$ and $<8.5\%$ for $S$. in   We remind to Appendix \ref{subsec:app:predictive-power} for more detailed results, however the same observations done for the SIRS models applies in all tested cases. Hence, in summary, the dimensions required for our embeddings to capture almost complete information are reasonably small.

\begin{table} 
    \centering
    \resizebox{\linewidth}{!}{
    \begin{tabular}{cccc}
            \toprule
            & $\rho$ & $R$ & $S$ \\
            \midrule
            Immigration &  $0.023$/$0.030$/$0.023$ & $0.014$/$0.017$/$0.013$ & $0.024$/$0.024$/$0.024$ \\
            Isomerization & $0.027$/$0.043$/$0.027$ & $0.008$/$0.015$/$0.008$ & $0.043$/$0.047$/$0.043$ \\
            Transcription & $0.033$/$0.054$/$0.033$ & $0.011$/$0.019$/$0.010$ & $0.064$/$0.085$/$0.065$ \\
            $\mu_0$ ($10$-dim) & $0.034$/$0.039$/$0.035$ & $0.003$/$0.005$/$0.003$ & $0.005$/$0.006$/$0.005$ \\
            \bottomrule
    \end{tabular}
    }
    \vspace{0.1cm}
    \captionof{table}{Median RE (across $100$ experiments) when using STLkernel/stl2vec($250$)/stl2vec($500$) in learning model checking under different test trajectory distributions.}
    \label{tab:pred-power-summary}
    \vspace{-0.3cm}
\end{table}

\subsection{Conditional Generation of Trajectories}\label{subsec:cvae}

Another context in which stl2vec might be sensibly applied is that of conditional generation of trajectories, i.e.\ inside a model whose goal is to produce synthetic multivariate signals satisfying arbitrary STL properties. To the best of our knowledge, conditioning a deep learning model on temporal logic embeddings for generating time-series has not been studied before \cite{time-series-survey}. 

Conditional Variational Autoencoders (CVAE) \cite{cvae,cond-vae} are generative models that learn a probabilistic mapping between input data and distributions on a continuous latent space, conditioning the generation process on some given additional information. More in detail, given inputs $\bm x$ with associated conditioning vectors $\bm y$, CVAE maps $\bm x$ to latent representations $\bm z$ by simultaneously learning two parametric functions: a probabilistic generation network (decoder) $p_{\theta} (\bm x| \bm y, \bm z)$ and an approximated posterior distribution (encoder) $q_{\phi}(\bm z|\bm y, \bm x)$, by maximizing the evidence lower bound (given a prior $p_{\psi}(\bm z|\bm y)$):
\begin{align}
\begin{split}
    \mathcal{L}(\phi, \theta, \psi; \bm{x}, \bm{y}) = &  \mathbb{E}_{\bm{z}\sim q_{\phi}(\bm{z}|\bm{y}, \bm{x})} [\log p_{\theta}(\bm{x}|\bm{y}, \bm{z})] \\ & - \beta \cdot KL[q_{\phi}(\bm{z}|\bm{y}, \bm{x}) \| p_{\psi}(\bm{z}|\bm{y})]
\end{split}
    \label{eq:cvae-loss}
\end{align}

where $KL[\cdot \| \cdot]$ is the Kullback-Leibler divergence, weighted by a hyperparameter $\beta\in \mathbb{R}$ controlling the balance between the reconstruction accuracy and the regularization of the learned latent space \cite{beta-vae}. 
Once trained, one might use the decoder as a generative model, by sampling vectors $\bm z$ from the prior distribution and adding conditional information $\bm y$, to obtain a point $\hat{\bm x}$ which should satisfy the given condition. 

\begin{figure}[t]
    \vspace{0.2cm}
    \centering
    \includegraphics[width=0.87\linewidth]{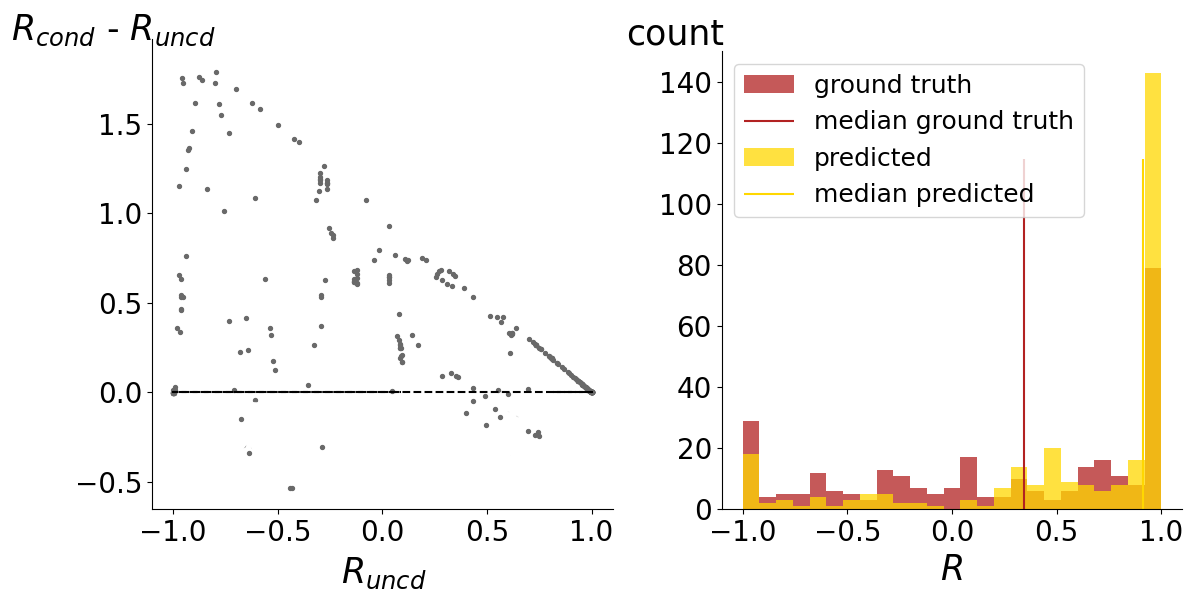}
    \vspace{0.1cm}
    \caption{Results of a random experiment for the conditional generation of trajectories using CVAE, in terms of average robustness.}
    \label{fig:cvae-results}
    \vspace{0.5cm}
\end{figure}

We devise a CVAE for multivariate time-series data, whose objective is to generate trajectories statistically similar to those of the data generating distribution and satisfying a given STL requirement, provided in the form of stl2vec embedding. More in detail, we encode signals using multiple stacked $\num{1}$D convolutional layers, and decode them using the same number of $\num{1}$D transposed convolutions; both the encoder and the decoder take as conditioning vector $\bm y$ the stl2vec representation of a property each input trajectory satisfies.  We trained the architecture on signals sampled from the SIRS model \cite{sirs} (which are $3$-dimensional time-series): we randomly sampled a set $\mathcal{D}_{\mathit{train}}$ of $\num{1000}$ formulae from $\mathcal{F}$, and $\forall \varphi \in \mathcal{D}_{\mathit{train}}$ we generated $\num{200}$ SIRS trajectories $\xi$ via SSA satisfying $\varphi$ (we do not exclude that the same signal might appear multiple times associated with different properties). The conditioning vector of each input $\xi$ is then computed with stl2vec, retaining $\num{250}$ components. 
We test the capability of the network to generate a trajectory $\xi$ compliant with a given STL property $\varphi$. Hence, for each test formula $\varphi\sim \mathcal{F}$ in the test set $\mathcal{D}_{\mathit{test}}$, represented as a $\num{250}$-dimensional semantic vector using stl2vec, we decode $\num{1000}$ signals, and compute the satisfaction probability and the average robustness of $\varphi$ on them, denoted as $S_{\mathit{cond}}$ and $R_{\mathit{cond}}$ respectively. Ideally, all the generated trajectories should satisfy (robustly) the corresponding $\varphi$; practically, we compare our results against the satisfaction probability and the average robustness of all $\varphi \in \mathcal{D}_{\mathit{test}}$ on a set of $\num{10000}$ unconstrained signals sampled from the SIRS model via SSA, denoted respectively as $S_{\mathit{uncd}}$ and $R_{\mathit{uncd}}$. Results are shown in Figure \ref{fig:cvae-results}, where we plot the difference in average robustness as a function of  $R_{\mathit{uncd}}$. 
Comparing the distribution of $R_{\mathit{uncd}}$ against that of $R_{\mathit{cond}}$, as done in Table \ref{tab:cvae-quantiles} and on the histogram of Figure \ref{fig:cvae-results}, highlights the improvement in having trajectories compliant to a given STL requirement when using a generative model. We experimented with conditioning vectors of different dimensions: retaining $[\num{10}, \num{50}, \num{100}, \num{250}, \num{500}]$ components yields a median $R_{\mathit{cond}}$ of $[\num{0.7008}, \num{0.8134}, \num{0.8737}, \num{0.903}, \num{0.9023}]$ and a median $S_{\mathit{cond}}$ of $[\num{0.8305}, \num{0.9065}, \num{0.9363}, \num{0.9515}, \num{0.951}]$, respectively (being $R_{\mathit{uncd}}$ and $S_{\mathit{uncd}}$ as in Table \ref{tab:cvae-quantiles}). In stark constrast, using implicit STL kernel embeddings of dimension $\num{1000}$ we get median $R_{\mathit{cond}}$ and $S_{\mathit{cond}}$ of $\num{0.4657}$ and $\num{0.73}$, probably because they contain redundant information which confuses the algorithm.
In conclusion, our dedicated finite-dimensional embedding are much better suited for the task than the full semantic information $\rho(\varphi,\cdot)$ even if the latter can be represented also finite-dimensionally (by the Gram matrix for the original kernel) with high enough dimension from enough data. See Appendix \ref{subsec:app:cvae-results} for more results.

\begin{table} 
    \centering
    \resizebox{\linewidth}{!}{
    \begin{tabular}{cccccc}
            \toprule
             & 1perc & 1quart & median & 3quart & 99perc\\
            \midrule
           $R_{\mathit{uncd}}$  & -0.9994 $\pm$ 0.0004 & -0.5128 $\pm$ 0.0123 & 0.0869 $\pm$ 0.0104 & 0.7321 $\pm$ 0.0046 & 1.0 $\pm$ 0 \\
           $R_{\mathit{cond}}$  & -1.0  $\pm$ 0 & -0.6157  $\pm$ 0.0086 & 0.903  $\pm$ 0.0043 & 1.0  $\pm$ 0 & 1.0  $\pm$ 0 \\
           \midrule 
           $S_{\mathit{uncd}}$  & 3.23e-04 $\pm$ 0.0003 & 0.229 $\pm$ 0.0076 & 0.5243 $\pm$ 0.0051 & 0.8122 $\pm$ 0.0022 & 1.0 $\pm$ 0 \\
           $S_{\mathit{cond}}$  & 0.0 $\pm$ 0 & 0.1923 $\pm$ 0.0045 & 0.9515 $\pm$ 0.0021 & 1.0  $\pm$ 0 & 1.0  $\pm$ 0\\
           \bottomrule
    \end{tabular}
    }
    \vspace{0.1cm}
    \captionof{table}{Mean and standard deviation of quantiles of the distributions of $R_{\mathit{uncd}}$ (resp. $S_{\mathit{uncd}}$) and $R_{\mathit{cond}}$ (resp. $S_{\mathit{cond}}$), over $\num{300}$ test formulae, averaged over $30$ experiments.}
    \label{tab:cvae-quantiles}
    \vspace{-0.3cm}
\end{table}

\section{Related Work}\label{sec:related}

\paragraph{Finding continuous embedding of logical formulae} has been an active research topic lately, with several works using Graph Neural Networks (GNN) for encoding the parsing tree of a formula to a continuous representation \cite{gnn-nesy}. Most of them, however, consider propositional and/or first-order logic \cite{pooling-logic,embedding-gnn,contrastive-logic,nesy}, hence are hard to generalize to temporal logics such as STL. In \cite{spl} a Semantic Probabilistic Layer is devised to impose properties on the output of a DL model, leveraging circuit representations of formulae. Although strictly related to ours, the approach is specific for DL model. 
Other works such as \cite{logic-machines,rule-induction} devise NeSy architectures which approximates first-order logic operations with neural networks, and then implement rules as neural operators applied to tensor representations of premises, to generate tensor representation of conclusions. Finding continuous embeddings of temporal logic formulae is addressed in: \cite{ltl-automata}, where a GNN is used to construct semantic-based embeddings of automata generated from Linear Temporal Logic (LTL) formulae and \cite{robot-stl}, where STL formulae are mapped to a continuous space by training a skip-gram and then used inside a neural network controller. The main difference between our method and the cited works is that stl2vec embeddings are not learnt, hence they are more controllable and robust, since they do not rely upon any training. 

\paragraph{Using STL formalism inside machine learning algorithm} has been exploited in: \cite{stl-intepretable-classification}, where a STL formula is learnt which abstracts the computational graph of a neural networks trained to perform interpretable classification of time-series behaviour; \cite{stlnet}, where STL is used as language to enhance the training of  a neural network model for sequence predictions compliant to a set of pre-defined properties; \cite{deepstl}, in which a tool is devised for the translation of informal requirements, given as English sentences, into STL. In all these cases, we believe that our approach can be valuably integrated for enforcing the semantics of the involved properties inside the neural architectures. 

\paragraph{Logic-based distances}  between models are typically tackled in the area of formal method using branching logic, e.g. bisimulation metrics for Markov models \cite{bisimulation-1,bisimulation-2}; the problem of computing the distance between STL specifications is instead addressed in \cite{stl-metrics} and applied to the generation of designs that exhibit desired behaviors specified in STL, in the field of synthetic genetic circuits. Differently, our work does not focus on the (dis)similarity between formulae, but instead aims at finding a semantic-preserving continuous representation of STL properties.

\section{Conclusions}
In this work we propose a constructive algorithm for computing \textit{interpretable} finite-dimensional explicit embeddings of Signal Temporal Logic (STL) formulae. We demonstrate their predictive power both as features for learning models and as semantic conditioning vectors inside other algorithms; most importantly, we provide explanations for a vast amount of information retained by the embeddings, a task which is highly non-trivial in general, but which is possible in this scenario due to the semantic nature of the objects involved. We believe that stl2vec has the potential to be a new framework for incorporating background knowledge in learning algorithms, under the umbrella of Neuro-Symbolic computing. We plan to extend this algorithm to other logics, such as Linear Temporal Logic (LTL), by defining a suitable measure in the corresponding space of signals. We also aim at using stl2vec as semantic conditioning information inside learning algorithm in other contexts, such as the synthesis of robot controllers satisfying some given (safety) properties. Most importantly, we would like to devise a way for inverting such embeddings, hence opening the doors to plenty of other applications, such as requirement mining. 

\bibliography{biblio}

\newpage
\appendix
\input{supplementary}

 \end{document}

%% file: supplementary.tex






\section{Extended Background}\label{app:sec:background}

\paragraph{A measure over trajectories} can be algorithmically defined by the following sampling algorithm (from \cite{stl-kernel}), operating on piece-wise linear functions over the interval $\mathcal{I} = [a, b]$ (which is a dense subset of the set of continuous functions over $\mathcal{I}$, denoted as $\mathcal{C}(\mathcal{I})$):
\begin{enumerate}
\item Set a discretization step $\Delta$; define $N = \frac{b-a}{\Delta}$ and $t_i = a + i\Delta$;
  	\item Sample a starting point $\xi_0\sim\mathcal{N}(m', \sigma')$ and set $\xi(t_0) = \xi_0$;
  	\item Sample $K\sim(\mathcal{N}(m'', \sigma''))^2$, that will be the total variation of $\xi$;
  	\item Sample $N-1$ points $y_1,...,y_{N-1}\sim\mathbb{U}([0, K])$ and set $y_0=0$ and $y_n=K$;
  	\item Order $y_1, ..., y_{N-1}$ and rename them such that $y_1 \leq y_2 \leq...\leq y_{N-1}$;
  	\item Sample $s_0\sim\text{Discr}(-1, 1)$;
  	\item Set iteratively $\xi(t_{i+1}) = \xi(t_i) + s_{i+1}(y_{i+1}-y_i)$ with $s_{i+1} = s_is$,\\
  	$P(s=-1) = q$ and $P(s=1) = 1-q$, for $i = 1, 2, ..., N$.
   \item Linearly interpolate between consecutive points of the discretization to make the trajectory continuous.                                 
\end{enumerate}

Default parameters of the above procedure are set as: $a=0, b=100, \Delta=1, m'=m''=0'', \sigma'=\sigma''=1, q=0.1$.


Intuitively, the measure $\mu_0$ makes \emph{simple} trajectories more probable, if considering total variation and number of changes in monotonicity as indicators of complexity of signals. We recall that the total variation of a function defined in the interval $\mathcal{I}=[a, b]$ is $V_a^b(f) = \sup_{P\in\mathbf P}\sum_{i=0}^{n_{P}-1}|f(x_{i+1})-f(x_i)|$, where $\mathbf P=\{ P=\{x_{0},\dots ,x_{n_{P}}\}\mid P{\text{ is a partition of }}[a,b]\} $.

\paragraph{STL quantitative semantics} (i.e. robustness) is recursively defined as:
\begin{align*}
 & \rho(\pi,\xi,t) &=& f_\pi(\xi(t)) \qquad \text{for } \pi(\bm{x})=\big(f_\pi(\bm{x})\geq 0\big)\\
 & \rho(\lnot\varphi,\xi,t) &=& -\rho(\varphi,\xi,t)\\
  & \rho(\varphi_1\land\varphi_2,\xi,t) &=& \min\big(\rho(\varphi_1,\xi,t),  \rho(\varphi_2,\xi,t)\big)\\
 &\rho(\varphi_1\mathbf{U}_{[a, b]}\varphi_2,\xi,t) &=&  \max_{t'\in[t+a,t+b]}\big(\min\big(\rho(\varphi_2,\xi,t'), \\ &&& \min_{t''\in[t,t']}\rho(\varphi_1,\xi,t'')\big)\big)
\end{align*}
Moreover, we derive as customary from the until operator $\mathbf{U}_{[a, b]}$ the eventually $\mathbf{F}_{[a, b]}$ and always $\mathbf{G}_{[a, b]}$ (equivalently called globally) operators, whose robust semantics is as follows:
\begin{align*}
    & \rho(\mathbf{F}_{[a, b]}\varphi, \xi, t) &=& \max_{t'\in[t+a,t+b]} \rho(\varphi,\xi,t)\\ 
    & \rho(\mathbf{G}_{[a, b]}\varphi, \xi, t) &=& \min_{t'\in[t+a,t+b]} \rho(\varphi,\xi,t)
\end{align*}

In principle, STL robustness has its domain in the field of real numbers $\mathbb{R}$, however if one knows the distribution of trajectories in which robustness will be computed,  a \emph{normalized robustness} can be considered in order to reduce the impact of outliers. In this case, using $\mu_0$ as reference distribution on trajectories, we know that signals typically take values in the interval $[-3, 3]$, hence considering the standard definition of robustness, the predicates e.g. $x_1 - 10 \geq 0$ and $x_1 -10^7\geq 0$ would have a quantitative semantics differing by orders of magnitude, while being essentially equivalent for satisfiability (on high probability trajectories w.r.t. $\mu_0$). In order to make the learning less sensitive to such outliers, we can re-scale the computation of atomic predicates' robustness as $\hat{\rho}(\pi, \xi, t) = \text{tanh}(f_{\pi}(x_1, \dots, x_n))$ so that it has domain in $(-1, 1)$. Unless differently specified, we will work with normalized robustness in this manuscript, since we take $\mu_0$ as our reference distribution on trajectories.  

\paragraph{A measure over STL formulae} can be defined via the following syntax-tree random recursive growing scheme (from \cite{stl-kernel}): 
\begin{enumerate}
    \item We start from root, forced to be an operator node. For each node, with probability $p_{\mathit{leaf}}$ we make it an atomic predicate, otherwise it will be an internal node. 
    \item In each internal (operator) node, we sample its type using a uniform distribution, then recursively sample its child or children. 
    \item We consider atomic predicates of the form $x_i \leq \theta$ or $x_i \geq \theta$. We sample randomly the variable index (the dimension of the signals is a fixed parameter), the type of inequality, and sample $\theta$ from a standard Gaussian distribution $\mathcal{N}(0,1)$.
    \item For temporal operators, we sample the right bound of the temporal interval uniformly from $\{1,2,\ldots,t_{\mathit{max}}\}$, and fix the left bound to zero. 
  \end{enumerate}

Default parameters of the above procedure are set as:  $p_{\mathit{leaf}} = 0.5$  
and $t_{\mathit{max}} = 10$.

\begin{figure}[h!]
    \centering
    \includegraphics[width=0.45\linewidth]{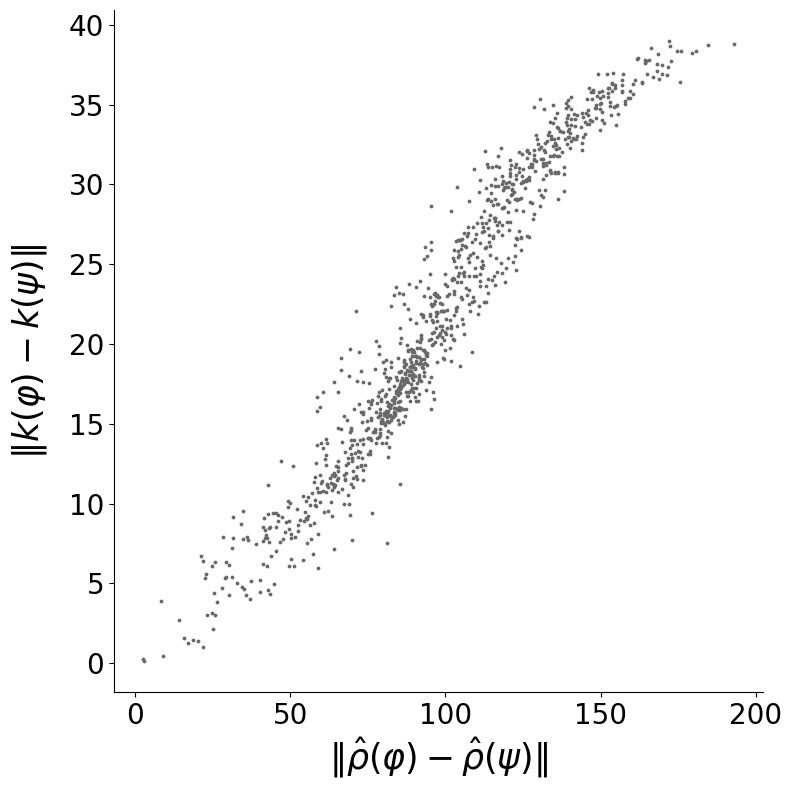}
    \includegraphics[width=0.45\linewidth]{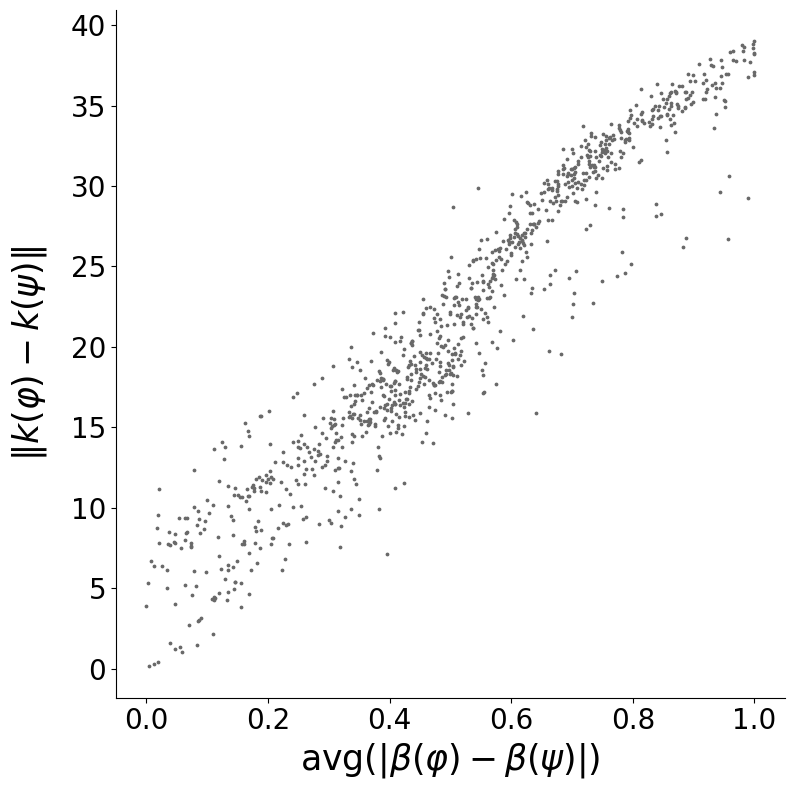}
    \vspace{0.1cm}
    \caption{$L_2$ distance between STL kernel embeddings of formulae vs (left) $L_2$ distance among their respective robustness vectors and (right) average number of Boolean satisfaction agreements, for $\num{1000}$ random STL formulae of $10$ variables.}
    \label{fig:app:semantic-similarity}
    \vspace{0.5cm}
\end{figure}

\paragraph{Semantic consistency of the STL kernel} defined in \cite{stl-kernel} is shown in Figure \ref{fig:app:semantic-similarity} where we verify that the distance of kernel embeddings is linearly correlated with: (left) the  distance among their robustness vectors, with a Pearson correlation $r$ of $\num{0.9689}$ and (right) the average number of agreements on a random set of $\num{10000}$ trajectories, with a Pearson correlation of $\num{0.9527}$.

\section{stl2vec: Deeper Insights and Additional Ablation Results}\label{app:sec:stl2vec}

\begin{figure}[h!]
    \centering
    \includegraphics[width=0.8\linewidth]{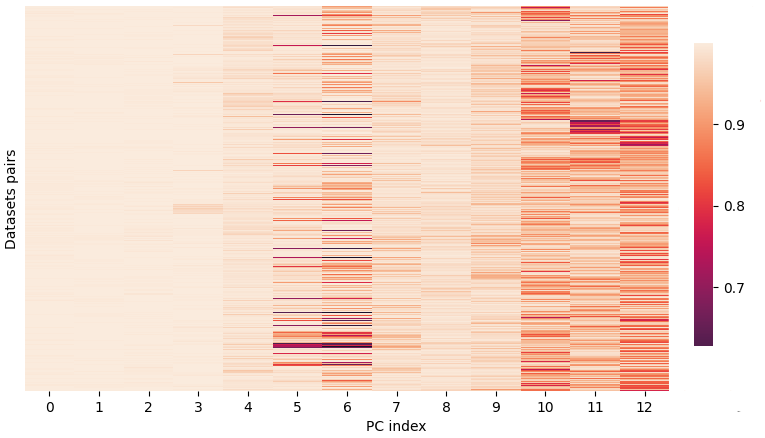}
    \vspace{0.1cm}
    \caption{Cosine similarity between corresponding PC, across all possible pairs of the $50$ datasets of STL formulae we consider.}
    \vspace{0.5cm}
    \label{fig:app:cosine-similarity}
\end{figure}

As we mentioned in Section \ref{sec:stl2vec}, we experimentally prove that, up to permutation of coordinates, the identified principal directions are almost the same across all datasets. We do so by computing the pairwise cosine similarity between corresponding PC of each possible pair of datasets, obtaining that, up to the $5^{th}$ PC, all datasets share a cosine similarity of at least $\num{0.95}$, moreover similarity stays above $\num{0.68}$ for all the $\num{13}$ considered components, with both mean and median similarity being $>\num{0.9}$ in every direction, for all possible pair of datasets, shown in Figure \ref{fig:app:cosine-similarity}.

Sampling formulae according to the distribution $\mathcal{F}$ (see \ref{app:sec:background}) allows to specify the maximum number of variables admitted in each STL requirement (hence the dimensionality of the signals over which they will be evaluated). This however do not impose that in every formulae all the possible variables will appear, for example, in our default case of signal with dimension $n=3$, we generate datasets of formulae containing requirements in which either $1$, $2$ or $3$ variables appear. 

\begin{figure}[h!]
    \centering
    \includegraphics[width=\linewidth]{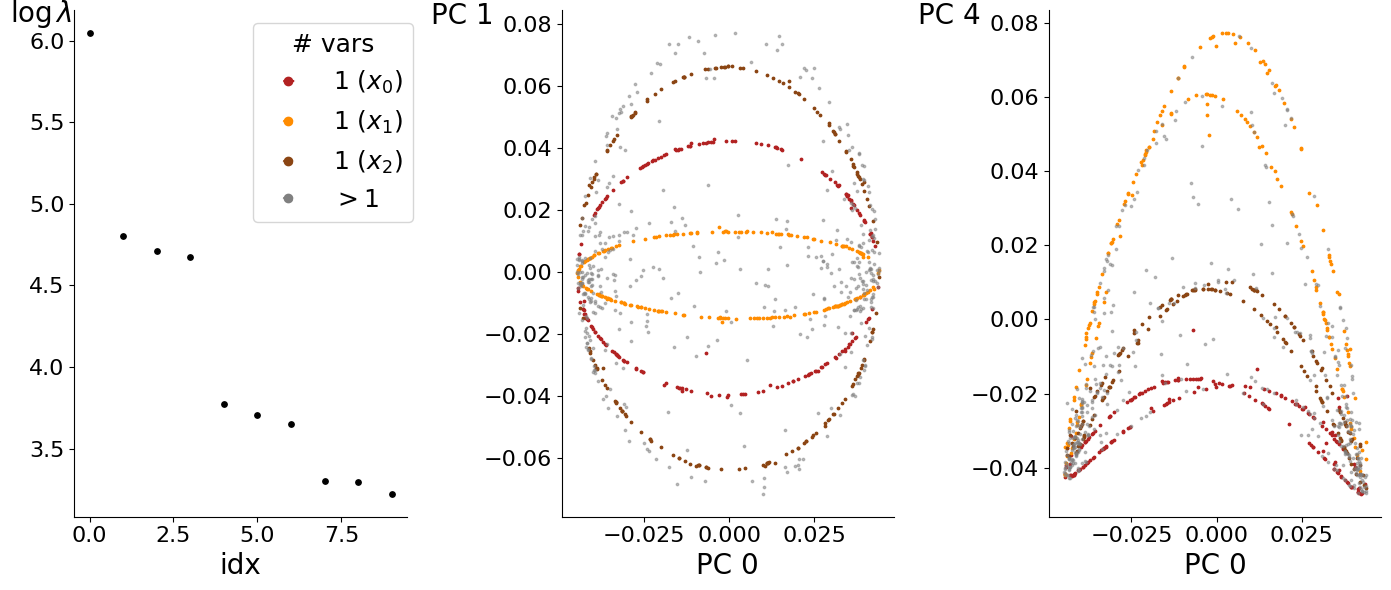}
    \vspace{0.1cm}
    \caption{Given a dataset of STL formulae with $3$ variables (left) spectrum of the covariance matrix of its Gram matrix; (center) $1^{st}$ vs $2^{nd}$ PC; (right) $1^{st}$ vs $4^{th}$ PC, in the righmost two highlighting formulae with only variable.} 
    \vspace{0.5cm}
    \label{fig:app:spectrum}
\end{figure}

Interestingly, if we plot PC$\num{0}$ against PC belonging to either the second or the third group we are not only able to individuate formulae in which only a variable appears, but also identify the involved variable, as reported in Figure \ref{fig:app:spectrum}.  Intuitively, this might depend on: (i) the fact that the explanations for the second and third groups of components hold variable-wise (suggesting that different variables are mapped to different semantic subspaces) and (ii)  the significant amount of information carried by PC$0$,  observable from the gap after PC$0$ in Figure \ref{fig:app:spectrum}. 

If we examine the behaviour of formulae obtained as per steps \ref{a:first}-\ref{a:fifth}, we observe that they have an average robustness $\approx \num{0}$ (more precisely, repeating the procedure $\num{100}$ times gives formulae with a minimum average robustness of $-\num{0.0207}\pm \num{0.0188}$ and a maximum of $\num{0.0122}\pm \num{0.0197}$). Moreover, looking at the quantiles of such robustness, we verify that, for the $\num{80}\%$ of the formulae we test, such formulae are robustly satisfied and robustly unsatisfied on a comparable number of trajectories (in some sense they are able to \emph{separate} the space of trajectories). In this case the value of the PC under analysis is the (variable-wise) similarity to formulae which are able to separate a random set of trajectories, which might be interpreted as a proxy of the ability of formulae to distinguish trajectories, when considered in each dimension separately w.r.t. variables. 

When formulae contain only one variable, the explanation for the third group of components given in steps \ref{b:first}-\ref{b:fourth} generate, for each formula $\varphi$ containing variable $i$, a quantity $\bm{\tilde{\rho}|x_i}$ which, following the notation of Section \ref{subsec:explain}, indicates the mean absolute deviation of the robustness from the value it has on the constant trajectory $\bm 0$. Indeed, $\forall k, \rho(\varphi, \xi_{ik})$ is a constant (not containig $\varphi$ variables different from $x_i$). In light of these, and recalling that PC$0$ is linearly correlated with the median robustness of formulae over trajectories, we can look at the rightmost plot of Figure \ref{fig:app:spectrum}, and in particular to points corresponding to formulae of only $1$ variable, and comment the following: formulae having a low value of PC$4$ are those whose robustness on random trajectories is very similar to the one they have on the constant trajectory $0$; up to some extent, we can say that their robustness is almost constant across all trajectories in $\mu_0$. Viceversa, formulae with a high value of PC$4$ are those whose robustness on random trajectories significantly varies across $\mu_0$. Considering now PC$0$, which represents the median robustness of formulae on signals sampled from $\mu_0$, having extreme values means having either an extremely low, or an extremely high median robustness, a situation coherent with formulae having an almost constant value of robustness (e.g. tautologies and contradictions). On the other hand, having a value of PC$0$ close to $0$ means, on most cases (cfr Section \ref{subsec:explain}), having a great variability on the robustness vector across $\mu_0$, in accordance to the meaning provided for formulae having an high value of PC$4$. 

To enforce this statement, we indeed verified that, for formulae containing only one variable, the third group of components is linearly correlated with the variance in the robustness vector of each formulae, i.e. given a formula $\varphi$ containing only one variable and considering set of $m$ random trajectories $\{\xi_k\}_{k=1}^m$ , to the variance of the vector $\bm \rho (\varphi) = \{\rho(\varphi, \xi_k)\}_{k=1}^m$. For the default case of $n=3$, considering the subset of formulae in which, respectively, only $x_0$, $x_1$ and $x_2$ appear, we have that the quantiles of the distribution of the absolute Pearson correlation coefficient (computed over $50$ independent experiments) between resp. PC$4$, PC$5$ and PC$6$ and the variance of the robustness vectors of the three subsets of formulae are: $[0.87556 0.89049 0.94194 0.95675 0.96231]$, $[0.91662 0.93629 0.95728 0.96229 0.96698]$ and $[0.72224 0.92174 0.94452 0.95222 0.96588]$.

\subsection{Supplementary plots and tables on ablations} \label{app:subsec:ablations}

\begin{figure}[h!]
    \centering
    \includegraphics[width=\linewidth, keepaspectratio]{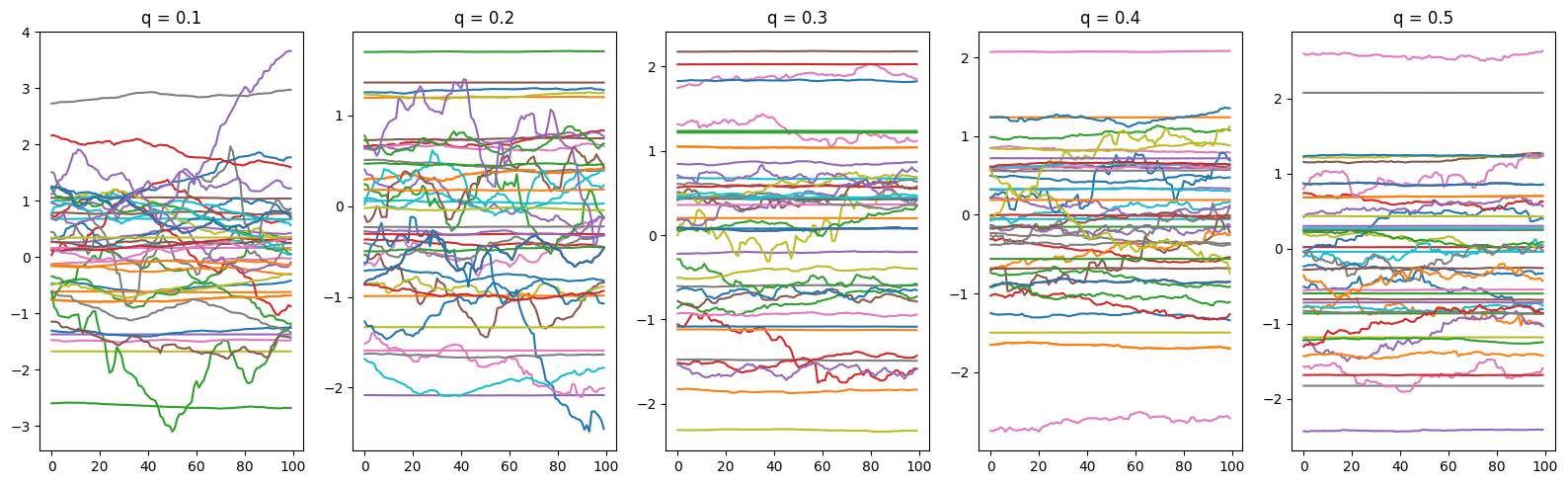}
    \vspace{0.1cm}
    \caption{Trajectories randomly sampled from $\mu_0$ varying the parameter $q$ in the sampling algorithm.}
    \vspace{0.5cm}
    \label{fig:app:signals-q}
\end{figure}

\begin{figure}[h!]
    \centering
    \includegraphics[width=\linewidth, keepaspectratio]{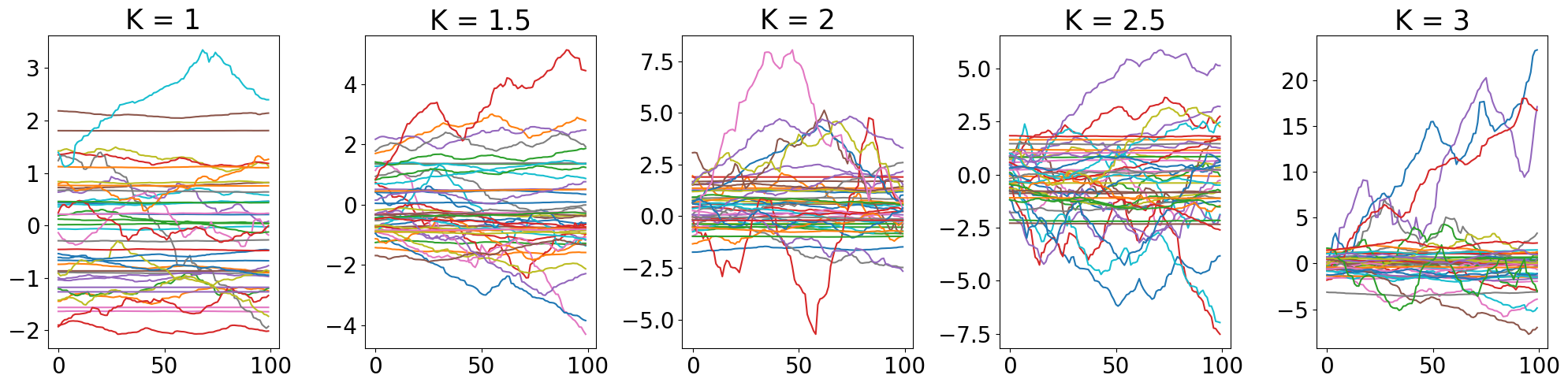}
    \vspace{0.1cm}
    \caption{Trajectories randomly sampled from $\mu_0$ varying the parameter $K$ in the sampling algorithm.}
    \label{fig:app:signals-k}
    \vspace{0.5cm}
\end{figure}

\begin{figure*}[h!]
    \centering
    \includegraphics[width=0.75\linewidth, keepaspectratio]{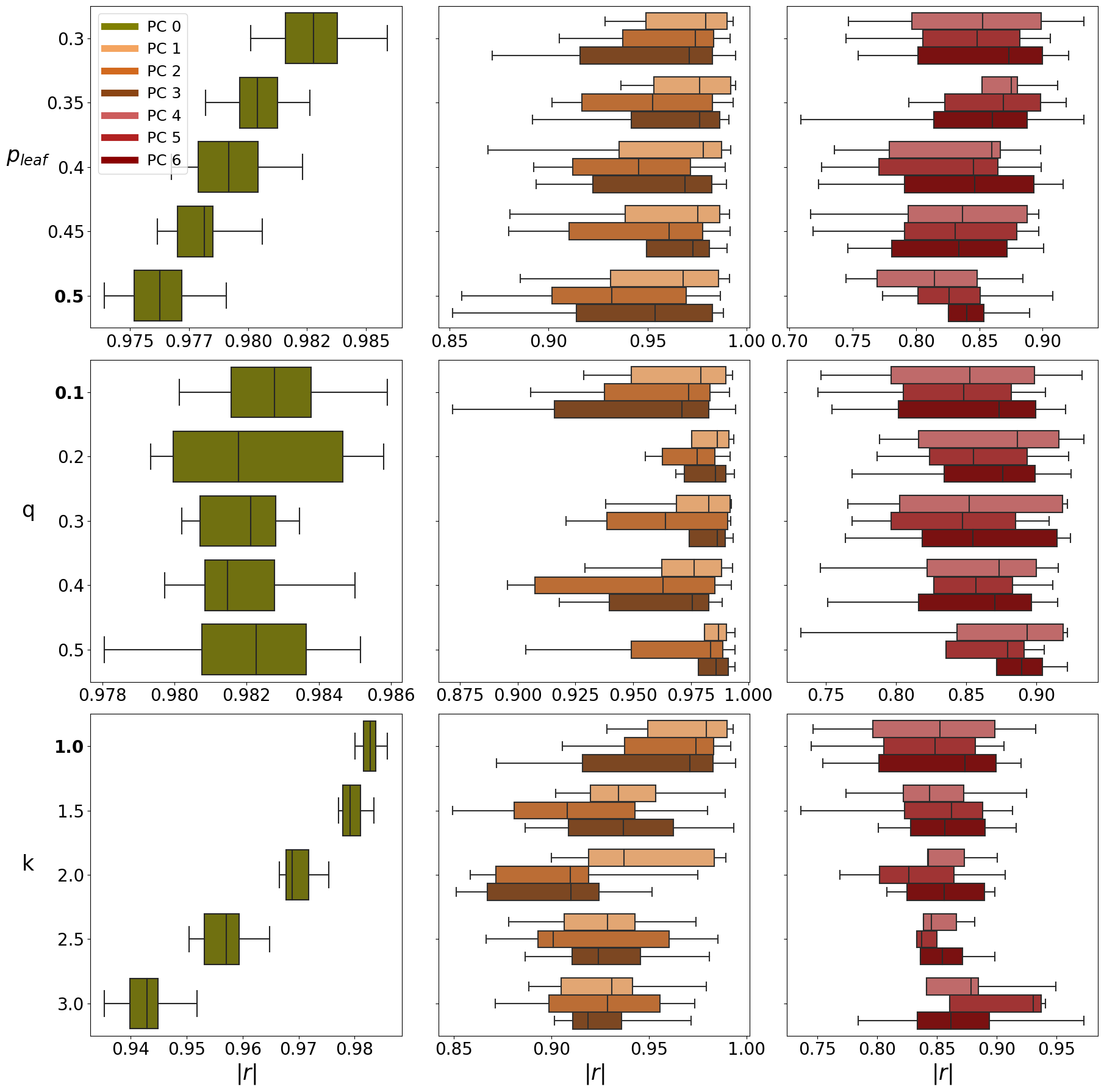}
    \vspace{0.1cm}
    \caption{Resilience of the explanations of PC to changes of the parameters (from top to bottom) $p_{\mathit{leaf}}$, $q$ and $K$ in terms of absolute Pearson Correlation Coefficient ($r$). Bold labels represent default parameters. Each column represents a different group of explained PC.}
    \label{fig:app:ablations-large}
    \vspace{0.5cm}
\end{figure*}

\paragraph{Ablations on the parameters of $\mu_0$ and $\mathcal{F}$} are extensively performed and quantiles of the experimental results over $50$ independent training datasets of STL formulae in terms of the absolute Pearson correlation coefficient $|r|$ are reported in Table \ref{tab:app:ablations-pleaf} for the parameter $p_{\mathit{leaf}}$ of the distribution $\mathcal{F}$, and Tables \ref{tab:app:ablations-q} and \ref{tab:app:ablations-k} for the parameters $q$ and $K$ of $\mu_0$, respectively. We recall that the higher the correlation coefficient, the stronger the linear correlation between its input quantities, and that it is arguably true that $|r|\geq 0.7$ indicates strong correlation. Hence we state that our explanations are resilient (or equivalently robust or stable) if, under different conditions or parameters, the correlation coefficient between our explanations and the PC remains high. All together these results are depicted in Figure \ref{fig:app:ablations-large}, where each column corresponds to a specific group of PC, as detailed in the main paper. Moreover, we also show visually how the distribution of trajectories changes when varying $q\in [0.1, 0.2, 0.3, 0.4, 0.5]$ (Figure \ref{fig:app:signals-q}) and when varying $K\in [1, 1.5, 2, 2.5, 3]$ (Figure \ref{fig:app:signals-k}).
In particular,  varying $q\in [\num{0.1}, \num{0.2}, \num{0.3}, \num{0.4}, \num{0.5}]$ yields trajectories with $[\num{17.818}, \num{31.069}, \num{40.771}, \num{46.289}, \num{51.286}]$ mean changes in their monotonicity, a visual glimpse on how the shape of signals changes under such conditions is given in Figure \ref{fig:app:signals-q}.

Differently, varying $K\in [\num{1}, \num{1.5}, \num{2}, \num{2.5}, \num{3}]$ yields trajectories with a mean total variation of $[\num{1.008}, \num{2.378}, \num{4.041}, \num{6.210}, \num{8.872}]$, a visual glimpse on how the shape of signals changes under such conditions is given in Figure \ref{fig:app:signals-q}.

Varying the parameter $p_{\mathit{leaf}}\in [\num{0.3}, \num{0.35}, \num{0.4}, \num{0.45}, \num{0.5}]$ of the formulae distribution $\mathcal{F}$ produces formulae with $[\num{28.627}, \num{12.477}, \num{7.639}, \num{5.696}, \num{4.418}]$ nodes on average.
 
\begin{figure*}[h!]
\begin{minipage}[t]{0.31\linewidth}
\centering
\resizebox{\linewidth}{!}{
\begin{tabular}{lllllll}
\toprule
{} & {} & \multicolumn{5}{c}{Absolute Pearson Correlation Coefficient} \\
\midrule
{} &  {} &  1perc &  1quart &   median &   3quart &   99perc \\
\midrule
{$p_{\mathit{leaf}}=0.3$} & {} & {} & {} & {} & {} & {} \\
\midrule
{} & PC$0$ & 0.98013 & 0.98158 & 0.98277 & 0.98380& 0.98591 \\
\hline
{} &  PC$1$ & 0.92863 & 0.94931 & 0.97932 & 0.99007 & 0.99305 \\
{} & PC$2$ & 0.90559 & 0.93758 & 0.97396 & 0.98322 & 0.99176 \\
{} & PC$3$ & 0.87180& 0.91584 & 0.97111 & 0.98275 & 0.99445 \\
\hline 
{} & PC$4$ & 0.74663 & 0.79653 & 0.85240& 0.89867 & 0.93248 \\
{} & PC$5$ & 0.74463 & 0.80540& 0.84841 & 0.88220& 0.90619 \\
{} &  PC$6$ & 0.75447 & 0.80160& 0.87320& 0.89963 & 0.92059 \\
\midrule
{$p_{\mathit{leaf}}=0.35$} & {} & {} & {} & {} & {} & {} \\
\midrule 
{} & PC$0$ & 0.97820& 0.97966 & 0.98039 & 0.98125 & 0.98262 \\
\hline
{} & PC$1$ & 0.93650& 0.95331 & 0.97622 & 0.99187 & 0.99440\\
{} & PC$2$ & 0.90182 & 0.91693 & 0.95241 & 0.98276 & 0.99321 \\
{} & PC$3$ & 0.89198 & 0.94186 & 0.97619 & 0.98638 & 0.99100\\
\hline 
{} & PC$4$ & 0.75677 & 0.88007 & 0.85209 & 0.87519 & 0.91162 \\
{} & PC$5$ & 0.79432 & 0.82293 & 0.86893 & 0.89821 & 0.91880\\
{} & PC$6$ & 0.70895 & 0.81402 & 0.86041 & 0.88777 & 0.93238 \\
\midrule 
{$p_{\mathit{leaf}}=0.4$} & {} & {} & {} & {} & {} & {} \\
\midrule 
{} & PC$0$ & 0.97675 & 0.97790& 0.97919 & 0.98043 & 0.98232 \\
\hline
{} & PC$1$ & 0.86950& 0.93556 & 0.97797 & 0.98744 & 0.99188 \\
{} & PC$2$ & 0.89265 & 0.91212 & 0.94548 & 0.97178 & 0.98912 \\
{} & PC$3$ & 0.89370& 0.92241 & 0.96895 & 0.98244 & 0.98982 \\
\hline 
{} & PC$4$ & 0.73555 & 0.77908 & 0.85982 & 0.86668 & 0.89831 \\
{} & PC$5$ & 0.72519 & 0.77056 & 0.84550& 0.86480& 0.89904 \\
{} & PC$6$ & 0.72301 & 0.79096 & 0.84639 & 0.89290& 0.91601 \\
\midrule 
{$p_{\mathit{leaf}}=0.45$} & {} & {} & {} & {} & {} & {} \\
\midrule 
{} & PC$0$ & 0.97617 & 0.97703 & 0.97816 & 0.97850& 0.98060\\
\hline 
{} & PC$1$ & 0.88061 & 0.93864 & 0.97545 & 0.98631 & 0.99142 \\
{} & PC$2$ & 0.87989 & 0.91029 & 0.96076 & 0.97777 & 0.99166 \\
{} & PC$3$ & 0.86795 & 0.94935 & 0.97294 & 0.98104 & 0.99021 \\
\hline 
{} & PC$4$ & 0.71679 & 0.79387 & 0.79651 & 0.89685 & 0.88769 \\
{} & PC$5$ & 0.61863 & 0.69071 & 0.76077 & 0.80960& 0.89715 \\
{} & PC$6$ & 0.74591 & 0.78060& 0.83401 & 0.87200& 0.90067 \\
\midrule 
{$\bm{p_{\mathit{leaf}}=\bm{0.5}}$} & {} & {} & {} & {} & {} & {} \\
\midrule 
{} & PC$0$ & 0.97393 & 0.97517 & 0.97627 & 0.97719 & 0.97908 \\
\hline
{} & PC$1$ & 0.88587 & 0.93131 & 0.96799 & 0.98576 & 0.99144 \\
{} & PC$2$ & 0.85624 & 0.90191 & 0.93203 & 0.96960& 0.98664 \\
{} & PC$3$ & 0.85171 & 0.91419& 0.95364 & 0.98283 & 0.98815 \\
\hline 
{} & PC$4$ & 0.74468 & 0.76928 & 0.81449 & 0.84850& 0.88428 \\
{} & PC$5$ & 0.77382 & 0.80144 & 0.82628 & 0.85062 & 0.90790\\
{} & PC$6$ & 0.76866 & 0.82553 & 0.83987 & 0.85343 & 0.88980\\
\bottomrule 
\end{tabular}
}
\vspace{0.1cm}
\captionof{table}{Resilience of the explanations of PC to changes of the parameter $p_{\mathit{leaf}}$ in terms of absolute Pearson Correlation Coefficient ($r$). Bold label represents the default.}
\label{tab:app:ablations-pleaf}
\end{minipage}\hfill 
\begin{minipage}[t]{0.31\linewidth}
\centering
\resizebox{\linewidth}{!}{
\begin{tabular}{lllllll}
\toprule
{} & {} & \multicolumn{5}{c}{Absolute Pearson Correlation Coefficient} \\
\midrule
{} &  {} &  1perc &  1quart &   median &   3quart &   99perc \\
\midrule 
{\textbf{q=$\bm{0.1}$}} & {} & {} & {} & {} & {} & {} \\
\midrule
{} & PC$0$ & 0.98013 & 0.98158 & 0.98277 & 0.98380& 0.98591 \\
\hline 
{} & PC$1$ & 0.92863 & 0.94931 & 0.97932 & 0.99007 & 0.99305 \\
{} & PC$2$ & 0.90559 & 0.93758 & 0.97396 & 0.98322 & 0.99176 \\
{} & PC$3$ & 0.8718 & 0.91584 & 0.97111 & 0.98275 & 0.99445 \\
\hline 
{} & PC$4$ & 0.74663 & 0.79653 & 0.8524 & 0.89867 & 0.93248 \\
{} & PC$5$ & 0.74463 & 0.8054 & 0.84841 & 0.8822 & 0.90619 \\ 
{} & PC$6$ & 0.75447 & 0.8016 & 0.8732 & 0.89963 & 0.92059 \\
\midrule
{q=$0.2$} & {} & {} & {} & {} & {} & {} \\
\midrule
{} & PC$0$ & 0.97934 & 0.97996 & 0.98177 & 0.98467 & 0.98581 \\
\hline 
{} & PC$1$ & 0.94717 & 0.97532 & 0.9866 & 0.99142 & 0.99374 \\
{} & PC$2$ & 0.95535 & 0.96256 & 0.97788 & 0.98548 & 0.9921 \\
{} & PC$3$ & 0.96849 & 0.97227 & 0.98562 & 0.99019 & 0.99378 \\
\hline 
{} & PC$4$ & 0.78842 & 0.81604 & 0.88664 & 0.91602 & 0.93367 \\
{} & PC$5$ & 0.78644 & 0.82408 & 0.85523 & 0.89326 & 0.92291 \\ 
{} & PC$6$ & 0.76877 & 0.83431 & 0.87587 & 0.89921 & 0.92446 \\
\midrule
{q=$0.3$} & {} & {} & {} & {} & {} & {} \\
\midrule
{} & PC$0$ & 0.98020& 0.9807 & 0.98212 & 0.98281 & 0.98347 \\
\hline 
{} & PC$1$ & 0.93809 & 0.9687 & 0.98291 & 0.9919 & 0.99261 \\
{} & PC$2$ & 0.92105 & 0.93859 & 0.96395 & 0.99092 & 0.99228 \\
{} & PC$3$ & 0.93887 & 0.97438 & 0.9866 & 0.98991 & 0.99346 \\
\hline 
{} & PC$4$ & 0.76595 & 0.80274 & 0.85207 & 0.91848 & 0.92204 \\
{} & PC$5$ & 0.76873 & 0.79641 & 0.84746 & 0.8853 & 0.90896 \\ 
{} & PC$6$ & 0.7639 & 0.81856 & 0.85467 & 0.91469 & 0.92424 \\
\midrule
{q=$0.4$} & {} & {} & {} & {} & {} & {} \\
\midrule
{} & PC$0$ & 0.97972 & 0.98084 & 0.98146 & 0.98278 & 0.98501 \\
\hline 
{} & PC$1$ & 0.92922 & 0.96255 & 0.97651 & 0.98824 & 0.9932 \\
{} & PC$2$ & 0.89567 & 0.9074 & 0.96294 & 0.98541 & 0.9925 \\
{} & PC$3$ & 0.91796 & 0.93964 & 0.97576 & 0.98288 & 0.98858 \\
\hline 
{} & PC$4$ & 0.74619 & 0.82221 & 0.87334 & 0.89986 & 0.9156 \\
{} & PC$5$ & 0.7323 & 0.82708 & 0.85679 & 0.88311 & 0.91152 \\ 
{} & PC$6$ & 0.75135 & 0.8161 & 0.87023 & 0.89653 & 0.91521 \\
\midrule
{q=$0.5$} & {} & {} & {} & {} & {} & {} \\
\midrule
{} & PC$0$ & 0.97806 & 0.98076 & 0.98226 & 0.98365 & 0.98517 \\
\hline 
{} & PC$1$ & 0.92160& 0.98086 & 0.98709 & 0.99047 & 0.99421 \\
{} & PC$2$ & 0.90341 & 0.94914 & 0.98357 & 0.98898 & 0.99415 \\
{} & PC$3$ & 0.94712 & 0.97837 & 0.98592 & 0.99111 & 0.99402 \\
\hline 
{} & PC$4$ & 0.73222 & 0.84336 & 0.89338 & 0.91887 & 0.92212 \\
{} & PC$5$ & 0.7471 & 0.83563 & 0.87956 & 0.90561 & 0.89109 \\ 
{} & PC$6$ & 0.75505 & 0.87159 & 0.88935 & 0.90415 & 0.92184 \\
\bottomrule 
\end{tabular}
}
\vspace{0.1cm}
\captionof{table}{Resilience of the explanations of PC to changes of the parameter $q$ in terms of absolute Pearson Correlation Coefficient ($r$). Bold label represents the default.}
\label{tab:app:ablations-q}
\end{minipage}\hfill 
\begin{minipage}[t]{0.31\linewidth}
\centering
\resizebox{\linewidth}{!}{
\begin{tabular}{lllllll}
\toprule
{} & {} & \multicolumn{5}{c}{Absolute Pearson Correlation Coefficient} \\
\midrule
{} &  {} &  1perc &  1quart &   median &   3quart &   99perc \\
\midrule
{\textbf{K=$\bm 1$}} & {} & {} & {} & {} & {} & {} \\
\midrule
{} & PC$0$ & 0.98013 & 0.98158 & 0.98277 & 0.98380& 0.98591 \\
\hline 
{} & PC$1$ & 0.92863 & 0.94931 & 0.97932 & 0.99007 & 0.99305 \\
{} & PC$2$ & 0.90559 & 0.93758 & 0.97396 & 0.98322 & 0.99176 \\
{} & PC$3$ & 0.87180& 0.91584 & 0.97111 & 0.98275 & 0.99445 \\
\hline 
{} & PC$4$ & 0.74663 & 0.79653 & 0.8524 & 0.89867 & 0.93248 \\
{} & PC$5$ & 0.74463 & 0.80540& 0.84841 & 0.88220& 0.90619 \\ 
{} & PC$6$ & 0.75447 & 0.80160& 0.87320& 0.89963 & 0.92059 \\
\midrule
{K=$1.5$} & {} & {} & {} & {} & {} & {} \\
\midrule
{} & PC$0$ & 0.97715 & 0.97789 & 0.97922 & 0.98108 & 0.98346 \\
\hline 
{} & PC$1$ & 0.90226 & 0.92017 & 0.93429 & 0.95332 & 0.98918 \\
{} & PC$2$ & 0.84947 & 0.88110& 0.90809 & 0.94287 & 0.98013 \\
{} & PC$3$ & 0.88676 & 0.90892 & 0.93689 & 0.96259 & 0.99336 \\
\hline 
{} & PC$4$ & 0.77392 & 0.82182 & 0.84394 & 0.87229 & 0.92480\\
{} & PC$5$ & 0.73610& 0.82273 & 0.86218 & 0.88827 & 0.91341 \\ 
{} & PC$6$ & 0.80080& 0.82812 & 0.85645 & 0.89027 & 0.91621 \\
\midrule
{K=$2$} & {} & {} & {} & {} & {} & {} \\
\midrule
{} & PC$0$ & 0.96662 & 0.96781 & 0.96883 & 0.97183 & 0.97535 \\
\hline 
{} & PC$1$ & 0.90005 & 0.91922 & 0.93716 & 0.98331 & 0.98935 \\
{} & PC$2$ & 0.85861 & 0.87148 & 0.90982 & 0.91897 & 0.97496 \\
{} & PC$3$ & 0.85135 & 0.86731 & 0.91016 & 0.92455 & 0.95175 \\
\hline 
{} & PC$4$ & 0.76284 & 0.84225 & 0.84267 & 0.87290& 0.90041 \\
{} & PC$5$ & 0.76866 & 0.80178 & 0.82633 & 0.86425 & 0.90697 \\ 
{} & PC$6$ & 0.80807 & 0.82483 & 0.85633 & 0.88960& 0.89829 \\
\midrule
{K=$2.5$} & {} & {} & {} & {} & {} & {} \\
\midrule
{} & PC$0$ & 0.95043 & 0.95317 & 0.95705 & 0.95935 & 0.96485 \\
\hline 
{} & PC$1$ & 0.87813 & 0.90674 & 0.92870& 0.94299 & 0.97394 \\
{} & PC$2$ & 0.86674 & 0.89319 & 0.90114 & 0.96039 & 0.98545 \\
{} & PC$3$ & 0.88667 & 0.91054 & 0.92404 & 0.94559 & 0.98084 \\
\hline 
{} & PC$4$ & 0.71614 & 0.83858 & 0.86651 & 0.84556 & 0.88150\\
{} & PC$5$ & 0.71891 & 0.83737 & 0.85000& 0.83308 & 0.87763 \\ 
{} & PC$6$ & 0.71970& 0.83594 & 0.85443 & 0.87135 & 0.89856 \\
\midrule
{K=$3$} & {} & {} & {} & {} & {} & {} \\
\midrule
{} & PC$0$ & 0.93533 & 0.93988 & 0.94295 & 0.94490& 0.95188 \\
\hline 
{} & PC$1$ & 0.88850& 0.90517 & 0.93101 & 0.94165 & 0.97934 \\
{} & PC$2$ & 0.87128 & 0.89867 & 0.92872 & 0.95559 & 0.97335 \\
{} & PC$3$ & 0.90171 & 0.91108 & 0.91888 & 0.93585 & 0.97168 \\
\hline 
{} & PC$4$ & 0.94940& 0.72640& 0.84106 & 0.88467 & 0.87857 \\
{} & PC$5$ & 0.94085 & 0.71766 & 0.93703 & 0.93077 & 0.86058 \\ 
{} & PC$6$ & 0.97282 & 0.78425 & 0.83385 & 0.86165 & 0.89378 \\
\bottomrule 
\end{tabular}
}
\vspace{0.1cm}
\captionof{table}{Resilience of the explanations of PC to changes of the parameter $K$ in terms of absolute Pearson Correlation Coefficient ($r$). Bold label represents the default.}
\label{tab:app:ablations-k}
\end{minipage}
\vspace{0.5cm}
\end{figure*}

\begin{figure*}[h!]
\begin{minipage}{0.48\linewidth}
    \centering
    \includegraphics[width=\linewidth, keepaspectratio]{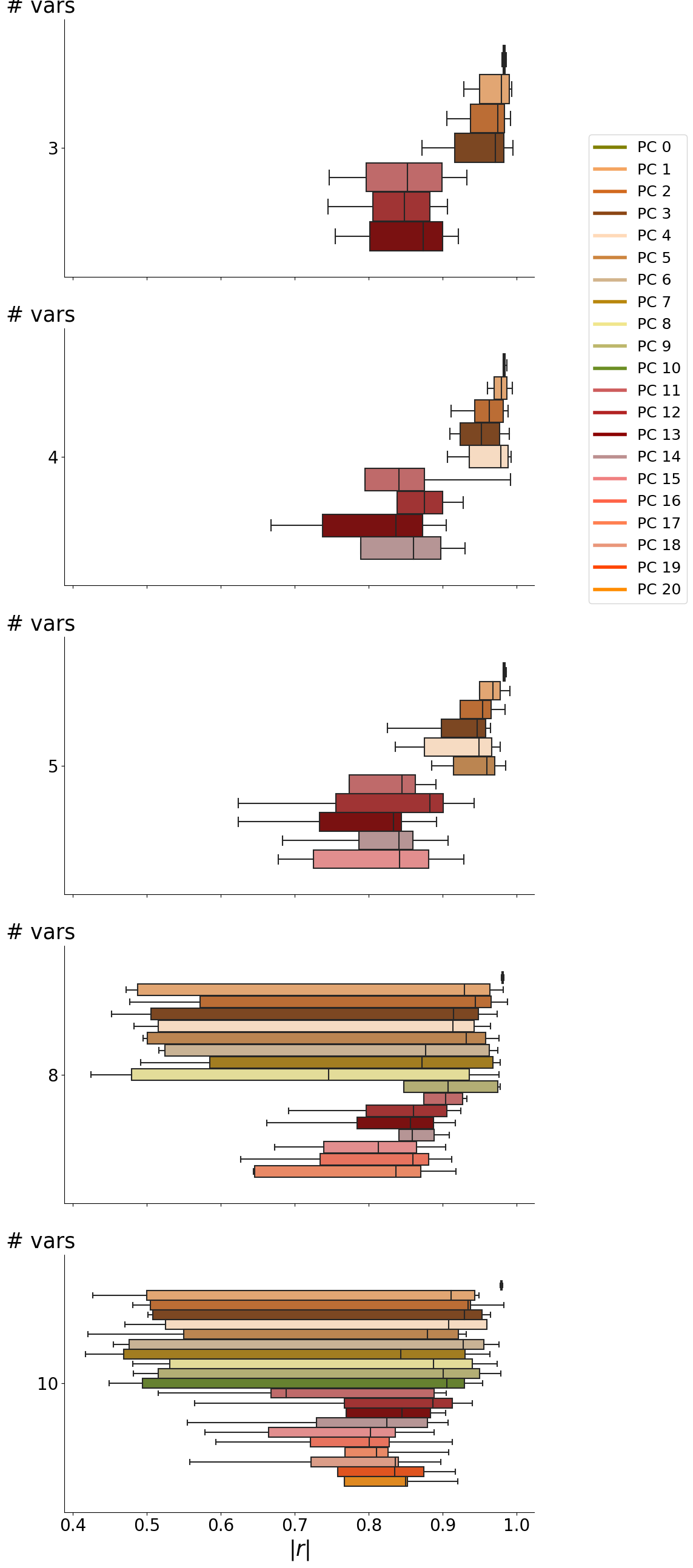}
    \vspace{0.1cm}
    \caption{Resilience of the explanations of PC to changes in the number of variables $v$ in terms of absolute Pearson Correlation Coefficient ($r$). Bold label represents the default.}
    \label{fig:app:ablation-vars}
\end{minipage}\hfill
\begin{minipage}{0.48\linewidth}
\centering
\resizebox{0.85\linewidth}{!}{
\begin{tabular}{lllllll}
\toprule
{} & {} & \multicolumn{5}{c}{Absolute Pearson Correlation Coefficient} \\
\midrule
{} &  {} &  1perc &  1quart &   median &   3quart &   99perc \\
\midrule 
{n=$3$} & {} & {} & {} & {} & {} & {} \\
\midrule
{} & PC$0$ & 0.98013 & 0.98158 & 0.98277 & 0.98380& 0.98591\\
\hline 
{} & PC$1$ & 0.92863 & 0.94931 & 0.97932 & 0.99007 & 0.99305\\
{} & PC$2$ & 0.90559 & 0.93758 & 0.97396 & 0.98322 & 0.99176\\
{} & PC$3$ & 0.87180& 0.91584 & 0.97111 & 0.98275 & 0.99445\\
\hline 
{} & PC$4$ & 0.74663 & 0.79653 & 0.85240& 0.89867 & 0.93248\\
{} & PC$5$ & 0.74463 & 0.80540& 0.84841 & 0.88220& 0.90619\\ 
{} & PC$6$ & 0.75447 & 0.80160& 0.87320& 0.89963 & 0.92059\\
\midrule
{n=$4$} & {} & {} & {} & {} & {} & {} \\
\midrule
{} & PC$0$ & 0.98121 & 0.98162 & 0.98264 & 0.98364 & 0.98641\\
\hline 
{} & PC$1$ & 0.95990& 0.96916 & 0.97932 & 0.98611 & 0.99349\\
{} & PC$2$ & 0.91144 & 0.94318 & 0.96248 & 0.98125 & 0.98804\\
{} & PC$3$ & 0.90919 & 0.92307 & 0.95212 & 0.97631 & 0.98952\\
{} & PC$4$ & 0.90625 & 0.93582 & 0.97797 & 0.98834 & 0.99215\\
\hline 
{} & PC$5$ & 0.63695 & 0.79478 & 0.84034 & 0.87544 & 0.99114\\ 
{} & PC$6$ & 0.67669 & 0.83802 & 0.87512 & 0.89959 & 0.92758\\
{} & PC$7$ & 0.66795 & 0.73766 & 0.83680& 0.87294 & 0.90475\\
{} & PC$8$ & 0.54658 & 0.78862 & 0.86033 & 0.89709 & 0.92972\\
\midrule
{n=$5$} & {} & {} & {} & {} & {} & {} \\
\midrule
{} & PC$0$ & 0.98144 & 0.98190& 0.98249 & 0.98402 & 0.98571\\
\hline 
{} & PC$1$ & 0.89876 & 0.94961 & 0.96754 & 0.97726 & 0.99040\\
{} & PC$2$ & 0.83006 & 0.92371 & 0.95373 & 0.96509 & 0.98376\\
{} & PC$3$ & 0.82498 & 0.89791 & 0.94618 & 0.95813 & 0.96402\\
{} & PC$4$ & 0.83606 & 0.87483 & 0.94841 & 0.96628 & 0.97739\\
{} & PC$5$ & 0.88513 & 0.91440& 0.95935 & 0.97050& 0.98449\\
\hline  
{} & PC$6$ & 0.56953 & 0.77330& 0.86235 & 0.84481 & 0.89054\\
{} & PC$7$ & 0.62323 & 0.75549 & 0.88228 & 0.90085 & 0.94195\\
{} & PC$8$ & 0.62335 & 0.73359 & 0.83316 & 0.84382 & 0.89163\\
{} & PC$9$ & 0.68326 & 0.78638 & 0.84060& 0.85975 & 0.90696\\
{} & PC$10$ & 0.67779 & 0.72482 & 0.84129 & 0.88077 & 0.92796\\
\midrule
{n=$8$} & {} & {} & {} & {} & {} & {} \\
\midrule
{} & PC$0$ & 0.97905 & 0.97995 & 0.98020& 0.98141 & 0.98209\\
\hline 
{} & PC$1$ & 0.47175 & 0.48785 & 0.92874 & 0.96373 & 0.98161\\
{} & PC$2$ & 0.47697 & 0.57180& 0.94356 & 0.9656 & 0.98721\\
{} & PC$3$ & 0.45185 & 0.50510& 0.91403 & 0.94779 & 0.97373\\
{} & PC$4$ & 0.48234 & 0.51503 & 0.91335 & 0.94197 & 0.96440\\
{} & PC$5$ & 0.49499 & 0.50071 & 0.93159 & 0.95793 & 0.97597\\
{} & PC$6$ & 0.52452 & 0.51580& 0.87629 & 0.96244 & 0.97390\\
{} & PC$7$ & 0.49167 & 0.58510& 0.87171 & 0.96754 & 0.97743\\
{} & PC$8$ & 0.42473 & 0.47906 & 0.84548 & 0.93547 & 0.97621\\
\hline  
{} & PC$9$ & 0.63559 & 0.97438 & 0.84722 & 0.90692 & 0.97725\\
{} & PC$10$ & 0.64099 & 0.87443 & 0.90373 & 0.92677 & 0.93281\\
{} & PC$11$ & 0.69137 & 0.79606 & 0.86069 & 0.90556 & 0.92388\\
{} & PC$12$ & 0.66200& 0.78390& 0.85613 & 0.88724 & 0.91673\\
{} & PC$13$ & 0.76264 & 0.84077 & 0.85901 & 0.88838 & 0.90876\\
{} & PC$14$ & 0.67269 & 0.73934 & 0.81307 & 0.86439 & 0.90376\\
{} & PC$15$ & 0.62642 & 0.73389 & 0.85955 & 0.88097 & 0.91192\\
{} & PC$16$ & 0.64398 & 0.64527 & 0.83660& 0.86977 & 0.91728\\
\midrule
{n=$10$} & {} & {} & {} & {} & {} & {} \\
\midrule
{} & PC$0$ & 0.97742 & 0.97805 & 0.97870& 0.97976 & 0.98080\\
\hline 
{} & PC$1$ & 0.42716 & 0.49973 & 0.91094 & 0.94342 & 0.94915\\
{} & PC$2$ & 0.48106 & 0.50458 & 0.93737 & 0.93397 & 0.98210\\
{} & PC$3$ & 0.50804 & 0.50143 & 0.92919 & 0.95265 & 0.96412\\
{} & PC$4$ & 0.46997 & 0.52490& 0.90786 & 0.95952 & 0.95984\\
{} & PC$5$ & 0.42036 & 0.54990& 0.87938 & 0.92131 & 0.93165\\
{} & PC$6$ & 0.47568 & 0.45466 & 0.9276 & 0.95564 & 0.97594\\
{} & PC$7$ & 0.41713 & 0.46831 & 0.84279 & 0.93017 & 0.96365\\
{} & PC$8$ & 0.48081 & 0.53094 & 0.88774 & 0.94009 & 0.97344\\
{} & PC$9$ & 0.48213 & 0.5155 & 0.90067 & 0.94938 & 0.97792\\
{} & PC$10$ & 0.44915 & 0.49401 & 0.90513 & 0.92924 & 0.95337\\
\hline  
{} & PC$11$ & 0.51493 & 0.68832 & 0.88797 & 0.90447 & 0.66811\\
{} & PC$12$ & 0.56419 & 0.76710& 0.88635 & 0.91304 & 0.93961\\
{} & PC$13$ & 0.5774 & 0.76963 & 0.84459 & 0.88325 & 0.90342\\
{} & PC$14$ & 0.55473 & 0.72911 & 0.82447 & 0.87913 & 0.90738\\
{} & PC$15$ & 0.57856 & 0.66471 & 0.80235 & 0.83566 & 0.88848\\
{} & PC$16$ & 0.59342 & 0.72112 & 0.80012 & 0.82713 & 0.91291\\
{} & PC$17$ & 0.51717 & 0.76745 & 0.81029 & 0.82562 & 0.90772\\
{} & PC$18$ & 0.55766 & 0.72197 & 0.83611 & 0.84005 & 0.89689\\
{} & PC$19$ & 0.54322 & 0.75785 & 0.83455 & 0.87388 & 0.91681\\
{} & PC$20$ & 0.59524 & 0.76662 & 0.84997 & 0.85242 & 0.92023\\
\bottomrule 
\end{tabular}
}
\vspace{0.1cm}
\captionof{table}{Resilience of the explanations of PC to changes in the number of variables $n$ in terms of absolute Pearson Correlation Coefficient ($r$). Bold label represents the default.}
\label{tab:app:ablations-v}
\end{minipage}
\end{figure*}

\paragraph{Ablations on the number of variables $n$ of STL formulae} (being $\num{3}$ the minimum and $\num{10}$ the maximum) are performed as well, resulting in the following: medians of the absolute linear correlation coefficient $|r|$ for the first PC are all $> \num{0.97}$, for the second group of PC they are all $>\num{0.84}$ and for the third $>\num{0.8}$, hence showing resilience of the explanations also to the change of the dimensionality of the signals. Quantiles of the distribution of $|r|$ across $50$ different datasets for each dimension of signals are reported in Table \ref{tab:app:ablations-v} and Figure \ref{fig:app:ablation-vars}. We remark here that, as reported in Section \ref{sec:stl2vec} (in particualar Table \ref{tab:xai-variance}), when the number of variables is higher then $\num{5}$ we are providing an interpretation for more than the $\num{95}\%$ of the variance in the data.

\begin{figure*}[h!]
\begin{minipage}{0.48\linewidth}
    \centering
    \includegraphics[width=0.85\linewidth, keepaspectratio]{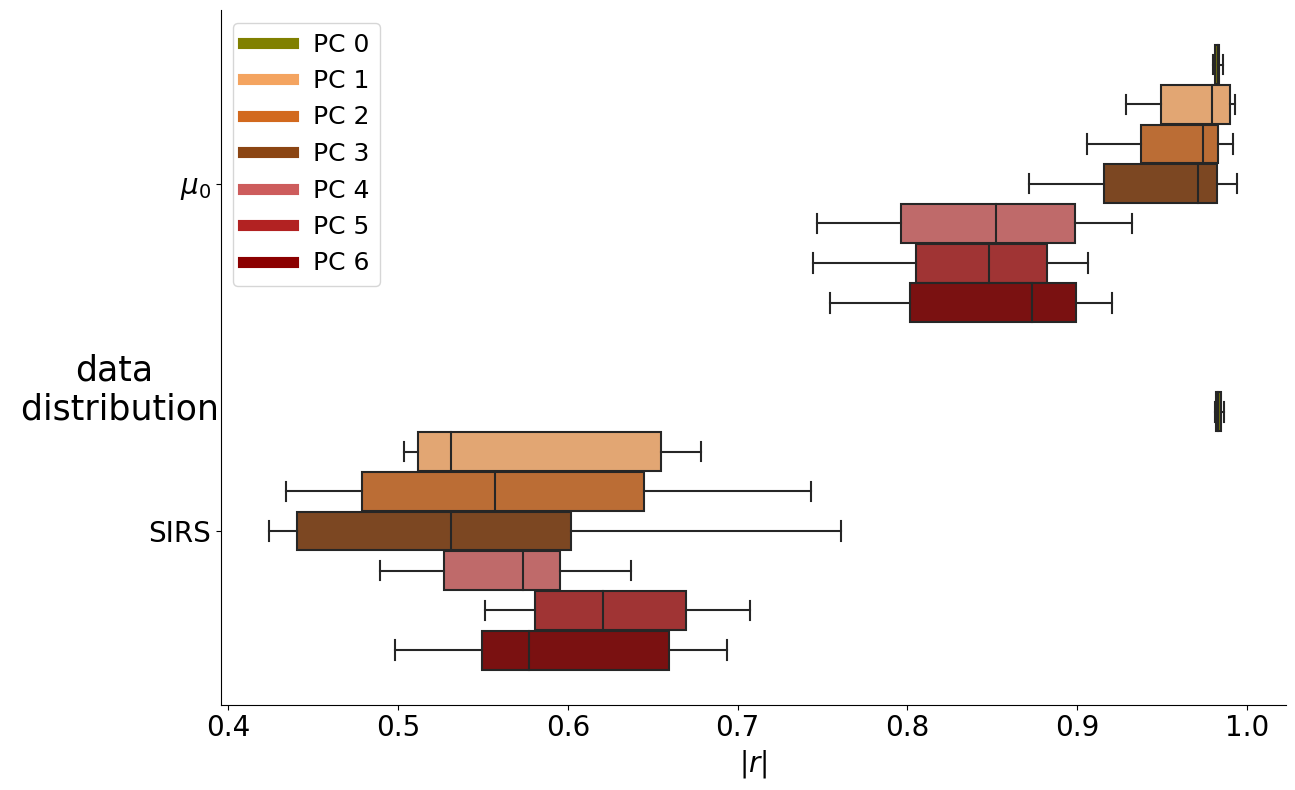}
    \vspace{0.1cm}
    \caption{Resilience of the explanations of PC to changes in the trajectory distribution, in terms of absolute Pearson Correlation Coefficient ($r$).}
    \label{fig:app:ablation-dist}
\end{minipage}\hfill
\begin{minipage}{0.48\linewidth}
\centering
\resizebox{0.9\linewidth}{!}{
\begin{tabular}{llllll}
\toprule
{} & \multicolumn{5}{c}{Absolute Pearson Correlation Coefficient} \\
\midrule
{} &  1perc &  1quart &   median &   3quart &   99perc \\
\midrule
PC$0$ & 0.98119 & 0.98201 & 0.98327 & 0.98500& 0.98657\\
\hline 
PC$1$ & 0.50325 & 0.51159 & 0.53107 & 0.65458 & 0.67820\\
PC$2$ & 0.43412 & 0.47881 & 0.55730& 0.64470& 0.74303\\
PC$3$ & 0.42409 & 0.44063 & 0.53108 & 0.60180& 0.76070\\
\hline 
PC$4$ & 0.48938 & 0.52717 & 0.57353 & 0.59549 & 0.63707\\
PC$5$ & 0.55139 & 0.58062 & 0.62054 & 0.66962 & 0.70748\\ 
PC$6$ & 0.49812 & 0.54959 & 0.57693 & 0.65955 & 0.69351\\
\bottomrule 
\end{tabular}
}
\vspace{0.1cm}
\captionof{table}{Resilience of the explanations of PC to changes in the trajectory distribution in terms of absolute Pearson Correlation Coefficient ($r$). Reported results are computed on trajectories sampled from the SIRS stochastic model.}
\label{tab:app:ablations-sirs}
\end{minipage}
\vspace{0.5cm}
\end{figure*}

\paragraph{Ablation on trajectory distribution} is done by considering the SIRS compartmental model, which is $3$-dimensional, and that can be simulated with the SSA algorithm. Quantiles of the distribution of the results over $50$ runs of the experiment are reported in Table \ref{tab:app:ablations-sirs} and Figure \ref{fig:app:ablation-dist}. Being signals of the SIRS model and those sampled from $\mu_0$ qualitatively very different, and having that the STL kernel (hence stl2vec) imposes a statistical filter on the similarity of formulae, given by the distribution according to which the kernel itself is computed, we do not expect our explanations to perfectly translate among trajectory distributions. However, we still witness a moderate linear correlation between PC and their candidate statistical descriptors, with a peak of correlation $>0.9$ for the first component. 

\section{Applications: Additional Results}\label{app:sec:experimental}

The experiments are implemented in Python exploiting the PyTorch \cite{pytorch} library for GPU acceleration. For sampling trajectories using SSA, the library StochPy \cite{stochpy} has been used. 
\subsection{More results on Predictive Power of stl2vec embeddings}\label{subsec:app:predictive-power}

Here we report the results of experiments made for predicting (i) robustness on single trajectories; (ii) average robustness of a stochastic system and (iii) satisfaction probability of a stochastic system. For each task, we perform ridge regression in the space of either plain STL kernel embeddings or stl2vec finite-dimensional explicit representations. We measure the performance both in terms of Relative Error (RE) and Absolute Error (AE), and we investigate how these indexes change when varying the number of retained components, for datasets of STL formulae of increasing complexity (from $n=3$ to $n=10$ variables) and for datasets of trajectories coming from different stochastic processes, i.e. Immigration ($1$-dim), Isomerization ($2$-dim) and Transcription ($3$-dim) from the suite of experiments of \cite{stl-kernel}, and SIRS ($3$-dim) epidemiological model. As a side note, we also highlight the results obtained when retaining only the coordinates that we are able to statistically describe, which are $1 + 2\cdot n$, remarking that being ridge regression a linear method, if features are interpretable, then the whole methodology is interpretable (hereafter, we will refer to this setting as \emph{interpretable case}). Unless differently specified, we use datasets of $1000$ formulae for constructing the embeddings, and we report average results over $100$ independent experiments. We fix $\mu_0$ with its default parameters as measure on the space of trajectories (i.e. for computing the kernel embeddings).

\begin{figure*}[h!]
\begin{minipage}{0.48\linewidth}
\centering
    \includegraphics[width=0.7\linewidth]{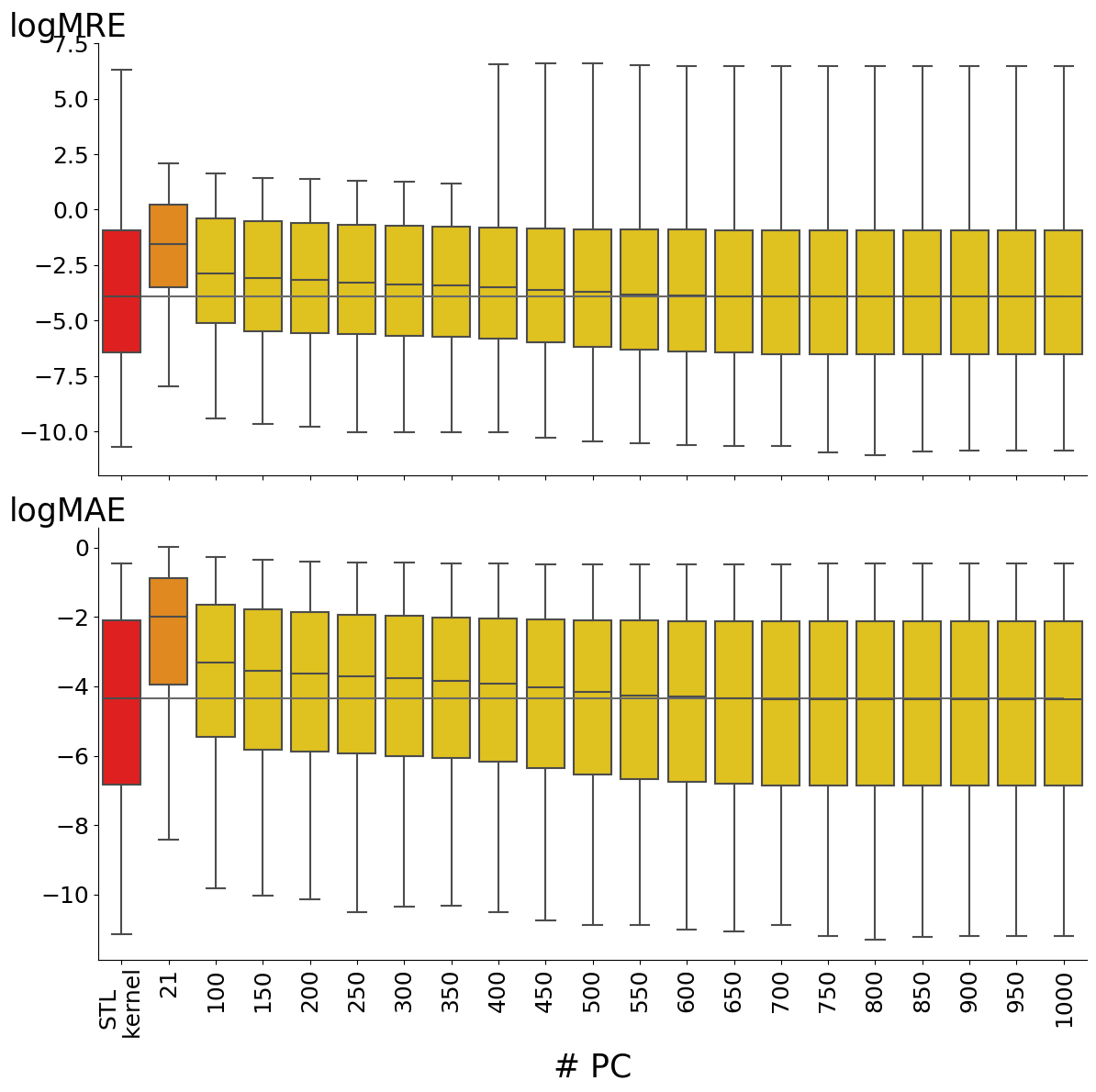}
    \vspace{0.1cm}
    \caption{Mean of the quantiles for $RE$ and $AE$ over $\num{100}$ experiments for predicting robustness on single trajectories $\rho$, varying the number of retained PC, on signals of dimension $10$.}
    \label{fig:app:singeltraj-boxplot}
\end{minipage}\hfill
\begin{minipage}{0.48\linewidth}
\centering    
\resizebox{0.85\linewidth}{!}{
\begin{tabular}{llllll|llll}
\toprule
{} & {} & \multicolumn{4}{c}{relative error (RE) } & \multicolumn{4}{c}{absolute error (AE)} \\
\midrule
{} &  {} &  1quart &   median &   3quart &  99perc &  1quart &   median &   3quart &   99perc \\
\midrule
{n=$3$} & \multicolumn{9}{c}{} \\ 
\midrule 
{} &  STL kernel & 0.00610& 0.02009 & 0.07813 & 2.90724 & 0.00419 & 0.01289 & 0.04102 & 0.28975\\ 
{} & stl2vec($7$) & 0.10159 & 0.21606 & 0.42622 & 8.16004 & 0.06679 & 0.13745 & 0.24162 & 0.68894\\
{} & stl2vec($\num{250}$) & 0.00794 & 0.02413 & 0.08513 & 3.14041 & 0.00551 & 0.01558 & 0.04482 & 0.28849\\
{} & stl2vec($\num{500}$) & 0.00575 & 0.01982 & 0.07775 & 2.90352 & 0.00399 & 0.01259 & 0.04070& 0.28965\\ 
\midrule 
{n=$4$} & \multicolumn{9}{c}{} \\ 
\midrule 
{} &  STL kernel & 0.00728 & 0.02428 & 0.09258 & 3.39204& 0.00505 & 0.01486 & 0.04821 & 0.31121\\
{} & stl2vec($9$) & 0.06679 & 0.13745 & 0.24162 & 0.68894 & 0.06446 & 0.13107 & 0.22821 & 0.62797\\
{} & stl2vec($\num{250}$) & 0.01123 & 0.03160& 0.10416 & 3.59948 & 0.00767 & 0.02019 & 0.05369 & 0.31286\\
{} & stl2vec($\num{500}$) & 0.00683 & 0.02395 & 0.09214 & 3.39061 & 0.00468 & 0.01447 & 0.04809 & 0.31096\\ 
\midrule
{n=$5$} & \multicolumn{9}{c}{} \\ 
\midrule 
{} &  STL kernel & 0.00801 & 0.02902 & 0.10852 & 3.97237 & 0.00538 & 0.01755 & 0.05715 & 0.35581\\ 
{} & stl2vec($11$) & 0.10261 & 0.21785 & 0.41231 & 7.20532 & 0.06370& 0.13245 & 0.23023 & 0.63934\\
{} & stl2vec($\num{250}$) & 0.01235 & 0.03740& 0.12187 & 4.10909 & 0.00834 & 0.02337 & 0.06309 & 0.35657\\
{} & stl2vec($\num{500}$) & 0.00765 & 0.02861 & 0.10795 & 3.95479 & 0.00515 & 0.01739 & 0.05725 & 0.35442\\ 
\midrule
{n=$8$} & \multicolumn{9}{c}{} \\ 
\midrule 
{} &  STL kernel & 0.01084 & 0.03839 & 0.14478 & 4.95479 & 0.00726 & 0.02372 & 0.07494 & 0.44167\\ 
{} & stl2vec($17$) & 0.10993 & 0.23077 & 0.42760& 8.17373 & 0.06930& 0.14162 & 0.23602 & 0.68824\\
{} & stl2vec($\num{250}$) & 0.01666 & 0.04974 & 0.16411 & 5.40267 & 0.01129 & 0.03100& 0.08318 & 0.45375\\
{} & stl2vec($\num{500}$) & 0.01039 & 0.03796 & 0.14511 & 4.97415 & 0.00695 & 0.02356 & 0.07484 & 0.44400\\ 
\midrule
{n=$10$} & \multicolumn{9}{c}{} \\ 
\midrule 
{} &  STL kernel & 0.01138 & 0.04141 & 0.15630&  5.36983 & 0.00742 & 0.02523 & 0.08240& 0.48631\\ 
{} & stl2vec($21$) & 0.10509 & 0.22659 & 0.43039 & 7.77285 & 0.06660& 0.13701 & 0.23335 & 0.66586\\
{} & stl2vec($\num{250}$) & 0.01905 & 0.05742 & 0.18647 & 5.86343 & 0.01313 & 0.03541 & 0.09263 & 0.50065\\
{} & stl2vec($\num{500}$) & 0.01083 & 0.04121 & 0.15560&  5.47765 & 0.00710& 0.02525 & 0.08162 & 0.48309\\ 
\bottomrule 
\end{tabular}
}
\vspace{0.1cm}
\captionof{table}{Mean of quantiles for RE and AE over $\num{100}$ experiments for prediction of robustness on single trajectories $\rho$, changing the number of variables in the dataset of formulae.}
\label{tab:app:single_traj}
\end{minipage}
\vspace{0.5cm}
\end{figure*}

\paragraph{Robustness on Single Trajectories} experiments consist in learning the function $\rho:\varphi\mapsto\rho(\varphi,\xi)$. Results for what concerns variations on the dimensionality of signals sampled from $\mu_0$ are reported in Figure \ref{fig:app:singeltraj-boxplot} and Table \ref{tab:app:single_traj}. For both reported errors, we observe that the performance of the regression algorithm improves until the dimensionality of stl2vec embeddings reaches a number of components which is roughly half the size of the original training datasets, after which it stabilizes to values comparable to that of plain STL kernel regression (see Figure \ref{fig:app:singeltraj-boxplot}). Performance of the interpretable case is significantly worse than that of the other dimensionality reported.

\begin{figure*}[h!]
\begin{minipage}{0.29\linewidth}
\centering
    \includegraphics[width=\linewidth]{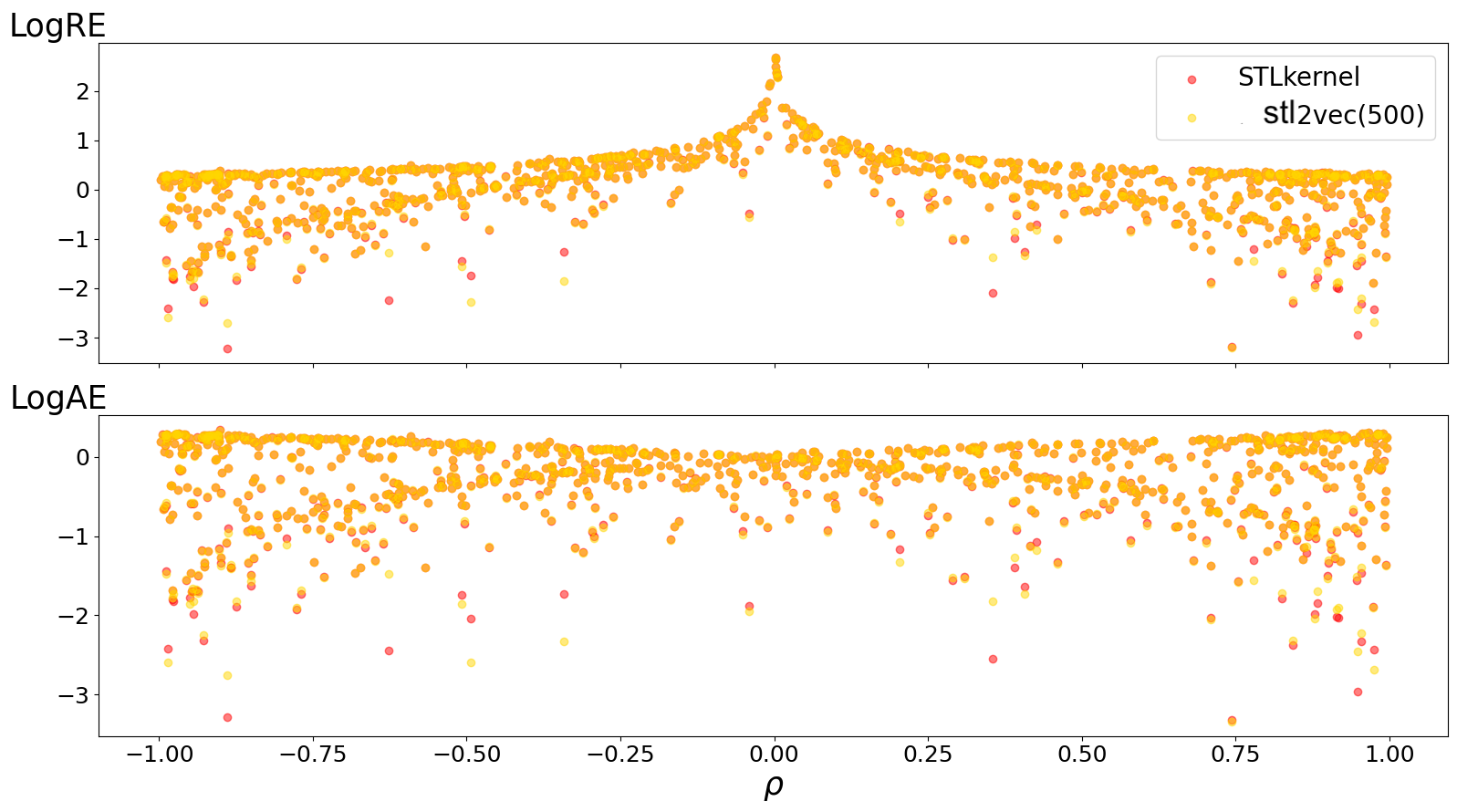}
    \vspace{0.1cm}
    \caption{Distribution of AE and RE of both STL kernel embeddings and stl2vec representation of dimension $500$, against ground truth robustness on single trajectories $\rho$, for a single random experiment.}
    \label{fig:app:single_exp_single_traj}
\end{minipage}\hfill
\begin{minipage}{0.29\linewidth}
\centering
    \includegraphics[width=\linewidth]{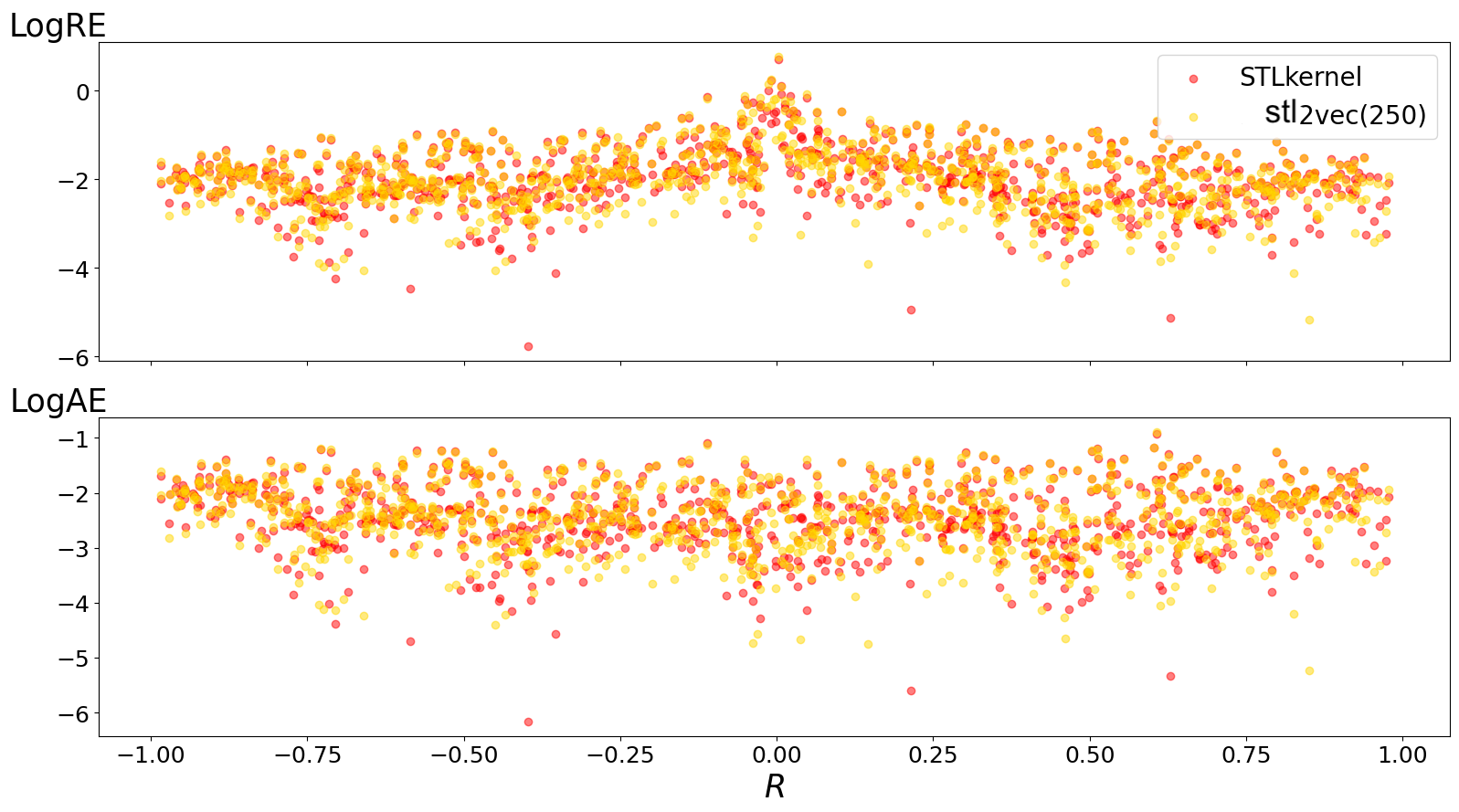}
    \vspace{0.1cm}
    \caption{Distribution of AE and RE of both STL kernel embeddings and stl2vec representation of dimension $250$, against ground truth average robustness $R$, for a single random experiment.}
    \label{fig:app:single_exp_avgrob}
\end{minipage}\hfill
\begin{minipage}{0.29\linewidth}
\centering
    \includegraphics[width=\linewidth]{fig/10-var-single-exp-avgrob.png}
    \vspace{0.1cm}
    \caption{Distribution of AE and RE of both STL kernel embeddings and stl2vec representation of dimension $300$, against ground truth satisfaction probability $S$, for a single random experiment.}
    \label{fig:app:single_exp_satprob}
\end{minipage}
\vspace{0.5cm}
\end{figure*}

If we focus on a single random experiment, we verify that both relative and absolute error of stl2vec, when keeping $500$ dimensions (as mentioned earlier), follow the same distribution of respectively relative and absolute error of implicit STL kernel embeddings. We visually show this in Figure \ref{fig:app:single_exp_single_traj}, and we numerically verify it using the Kolmogorov-Smirnov statistical test: providing as null hypothesis the equality between the error distribution of stl2vec($500$) and STL kernel outputs a p-value of $0.9999$ for both AE and RE.

\begin{figure*}[h!]
\begin{minipage}{0.48\linewidth}
\centering
    \includegraphics[width=0.7\linewidth]{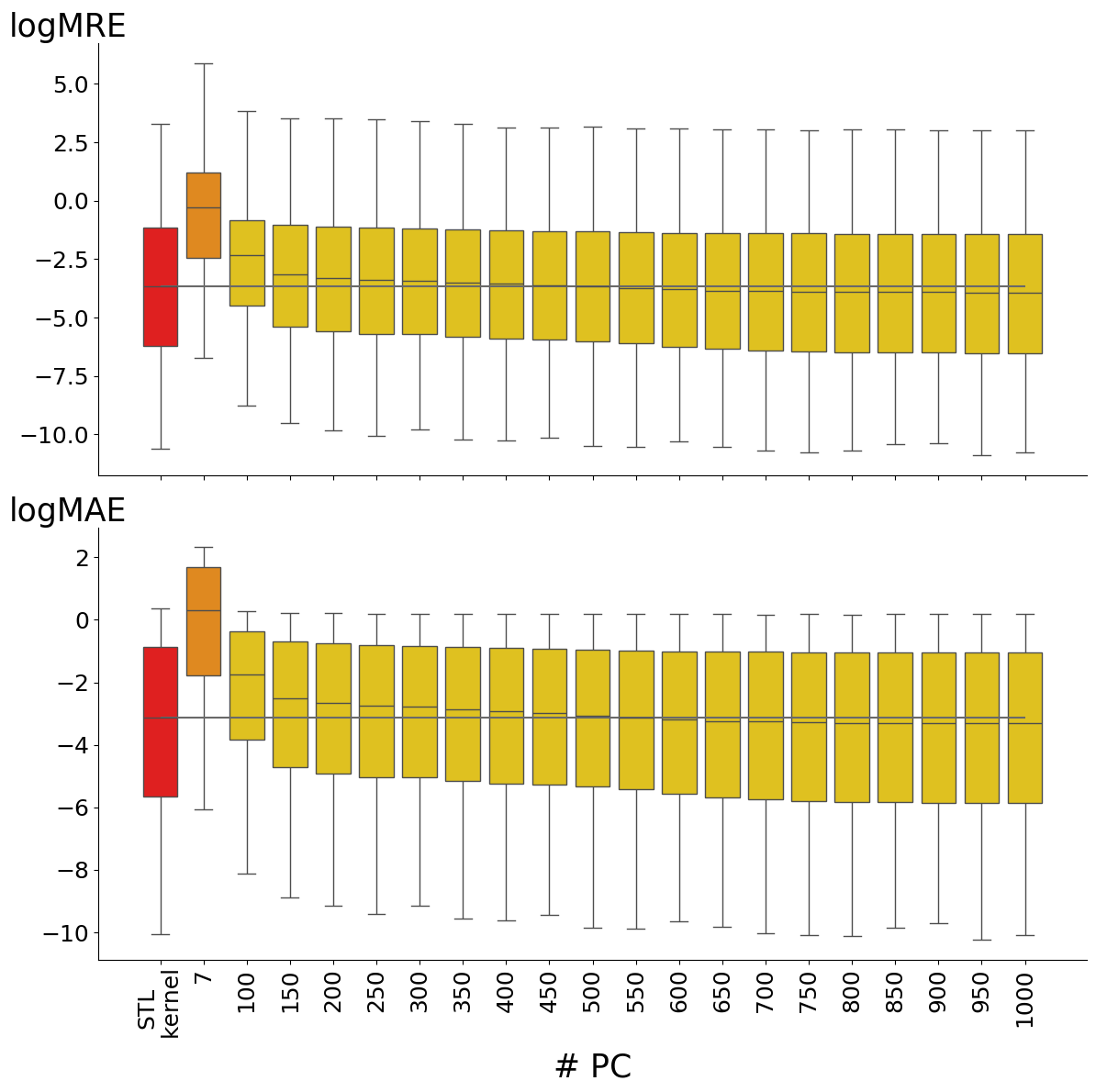}
    \vspace{0.1cm}
    \caption{Mean of the quantiles for $RE$ and $AE$ over $\num{100}$ experiments for predicting robustness on single trajectories $\rho$, on signals sampled from the SIRS model.}
    \label{fig:app:singeltraj-boxplot-other}
\end{minipage}\hfill
\begin{minipage}{0.48\linewidth}
\centering    
\resizebox{0.85\linewidth}{!}{
\begin{tabular}{llllll|llll}
\toprule
{} & {} & \multicolumn{4}{c}{relative error (RE) } & \multicolumn{4}{c}{absolute error (AE)} \\
\midrule
{} &  {} &  1quart &   median &   3quart &  99perc &  1quart &   median &   3quart &   99perc \\
\midrule
{Immigration} & \multicolumn{9}{c}{} \\ 
\midrule 
{} &  STL kernel & 0.00846 & 0.02337 & 0.08263 & 3.34974 & 0.00512 & 0.01208 & 0.03265 & 0.26956\\
{} & stl2vec($\num{250}$) &0.01131 & 0.03073 & 0.10354 & 3.94368 & 0.00708 & 0.01591 & 0.04141 & 0.27248\\
{} & stl2vec($\num{500}$) & 0.00829 & 0.02345 & 0.08611 & 3.45799 & 0.00511 & 0.01201 & 0.03372 & 0.26619\\ 
\midrule
{Isomerization} & \multicolumn{9}{c}{} \\ 
\midrule 
{} &  STL kernel & 0.00773 & 0.02728 & 0.10229 & 2.70708 & 0.0073 & 0.02382 & 0.07407 & 0.57839\\ 
{} & stl2vec($\num{250}$) & 0.01737 & 0.04351 & 0.13504 & 3.02634 & 0.01621 & 0.03909 & 0.09956 & 0.60761\\
{} & stl2vec($\num{500}$) & 0.00763 & 0.02701 & 0.10232 & 2.6718 & 0.0072 & 0.02377 & 0.07435 & 0.5797\\ 
\midrule 
{Transcription} & \multicolumn{9}{c}{} \\ 
\midrule 
{} &  STL kernel & 0.00924 & 0.03303 & 0.10723 & 2.70434 & 0.02292 & 0.07017 & 0.18515 & 0.96171\\ 
{} & stl2vec($\num{250}$) & 0.01845 & 0.05462 & 0.1567 & 3.17236 & 0.04536 & 0.11887 & 0.26418 & 1.00227\\
{} & stl2vec($\num{500}$) & 0.00962 & 0.03353 & 0.10921 & 2.74337 & 0.02396 & 0.07197 & 0.18696 & 0.9564\\ 
\midrule
{SIRS} & \multicolumn{9}{c}{} \\ 
\midrule 
{} &  STL kernel & 0.00772 & 0.02582 & 0.09225 & 1.41988 & 0.01362 & 0.04376 & 0.14283 & 0.92352\\ 
{} & stl2vec($\num{250}$) & 0.01246 & 0.03385 & 0.10293 & 1.26477 & 0.02409 & 0.06393 & 0.17317 & 0.83707\\
{} & stl2vec($\num{500}$) & 0.00917 & 0.02532 & 0.07942 & 1.14463 & 0.01769 & 0.04689 & 0.13455 & 0.79238\\ 
\bottomrule 
\end{tabular}
}
\vspace{0.1cm}
\captionof{table}{Mean of quantiles for RE and AE over $\num{100}$ experiments for prediction of robustness on single trajectories $\rho$, changing stochastic models.}
\label{tab:app:single_traj-other}
\end{minipage}
\vspace{0.5cm}
\end{figure*}

For what concerns tests done on trajectories coming from other stochastic processes, the same considerations hold and results are reported in Table \ref{tab:app:single_traj-other} and Figure \ref{fig:app:singeltraj-boxplot-other}. 

\begin{figure*}[h!]
\begin{minipage}{0.48\linewidth}
    \centering
    \includegraphics[width=0.7\linewidth]{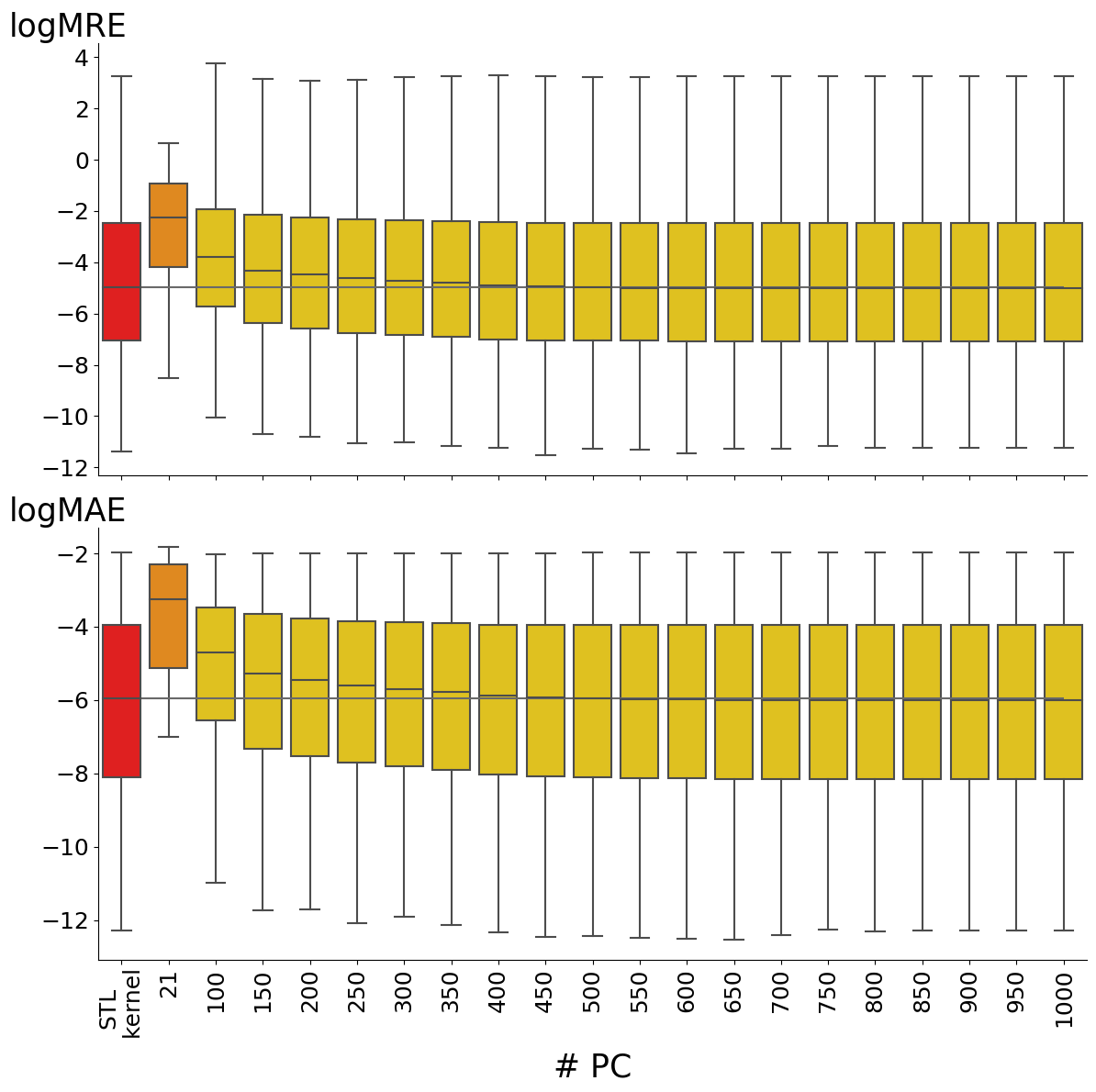}
    \vspace{0.1cm}
    \caption{Mean of the quantiles for $RE$ and $AE$ over $\num{100}$ experiments for predicting average robustness $R$, varying the number of retained PC, on signals of dimension $10$.}
    \label{fig:app:avgrob-boxplot}
\end{minipage}\hfill
\begin{minipage}{0.48\linewidth}
\centering    
\centering    
\resizebox{0.85\linewidth}{!}{
\begin{tabular}{llllll|llll}
\toprule
{} & {} & \multicolumn{4}{c}{relative error (RE) } & \multicolumn{4}{c}{absolute error (AE)} \\
\midrule
{} &  {} &  1quart &   median &   3quart &  99perc &  1quart &   median &   3quart &   99perc \\
\midrule
{n=$3$} & \multicolumn{9}{c}{} \\ 
\midrule 
{} &  STL kernel & 0.00302 & 0.00683 & 0.01797 & 0.49188 & 0.00107 & 0.00258 & 0.00706 & 0.05591\\ 
{} & stl2vec($7$) & 0.05318 & 0.10619 & 0.18066 & 1.92817 & 0.02015 & 0.03888 & 0.06335 & 0.13579\\
{} & stl2vec($\num{250}$) & 0.00305 & 0.00693 & 0.01829 & 0.47836 & 0.00109 & 0.00261 & 0.00717 & 0.05596\\
{} & stl2vec($\num{500}$) & 0.00292 & 0.00662 & 0.01770& 0.47532 & 0.00103 & 0.00250& 0.00702 & 0.05618\\ 
\midrule 
{n=$4$} & \multicolumn{9}{c}{} \\ 
\midrule 
{} &  STL kernel & 0.00306 & 0.00715 & 0.02041 & 0.50400& 0.00109 & 0.00277 & 0.00773 & 0.05496\\ 
{} & stl2vec($9$) & 0.04711 & 0.09311 & 0.15601 & 1.71762 & 0.01722 & 0.03363 & 0.05514 & 0.12168\\
{} & stl2vec($\num{250}$) & 0.00315 & 0.00754 & 0.02124 & 0.51606 & 0.00115 & 0.00287 & 0.00783 & 0.05463\\
{} & stl2vec($\num{500}$) & 0.00291 & 0.00685 & 0.02005 & 0.50314 & 0.00104 & 0.00264 & 0.00763 & 0.05522\\ 
\midrule
{n=$5$} & \multicolumn{9}{c}{} \\ 
\midrule 
{} &  STL kernel & 0.00322 & 0.00776 & 0.02279 & 0.52315 & 0.00116 & 0.00303 & 0.00841 & 0.05461\\ 
{} & stl2vec($11$) & 0.04092 & 0.08143 & 0.13798 & 1.52205 & 0.01492 & 0.02934 & 0.05063 & 0.11536\\
{} & stl2vec($\num{250}$) & 0.00365 & 0.00877 & 0.02505 & 0.55300& 0.00138 & 0.00338 & 0.00867 & 0.05503\\
{} & stl2vec($\num{500}$) & 0.00308 & 0.00743 & 0.02249 & 0.52253 & 0.00111 & 0.00292 & 0.00829 & 0.05480\\ 
\midrule
{n=$8$} & \multicolumn{9}{c}{} \\ 
\midrule 
{} &  STL kernel & 0.00318 & 0.00825 & 0.02675 & 0.54230& 0.00115 & 0.00331 & 0.00961 & 0.05799\\ 
{} & stl2vec($17$) & 0.03444 & 0.06818 & 0.11623 & 1.38049 & 0.01216 & 0.02492 & 0.04470& 0.10214\\
{} & stl2vec($\num{250}$) & 0.00466 & 0.01118 & 0.03244 & 0.61653 & 0.00186 & 0.00436 & 0.01034 & 0.05765\\
{} & stl2vec($\num{500}$) & 0.00313 & 0.00813 & 0.02702 & 0.52959 & 0.00113 & 0.00325 & 0.00952 & 0.05833\\ 
\midrule
{n=$10$} & \multicolumn{9}{c}{} \\ 
\midrule 
{} &  STL kernel & 0.00366 & 0.00936 & 0.03002 & 0.56771 & 0.00135 & 0.00375 & 0.01047 & 0.05858\\ 
{} & stl2vec($21$) & 0.03036 & 0.06282 & 0.10814 & 1.28478 & 0.01081 & 0.02313 & 0.04296 & 0.09811\\
{} & stl2vec($\num{250}$) & 0.00548 & 0.01318 & 0.03643 & 0.67124 & 0.00219 & 0.00507 & 0.01165 & 0.05895\\
{} & stl2vec($\num{500}$) & 0.00348 & 0.00912 & 0.02985 & 0.57566 & 0.00127 & 0.00367 & 0.01029 & 0.05876\\ 
\bottomrule 
\end{tabular}
}
\vspace{0.1cm}
\captionof{table}{Mean of quantiles for RE and AE over $\num{100}$ experiments for prediction of average robustness $R$, changing the number of variables in the dataset of formulae.}
\label{tab:app:avg_rob}
\end{minipage}
\vspace{0.5cm}
\end{figure*}

\paragraph{Average Robustness} experiments consist in learning the function $\mathbb E_{\xi\sim\mu_0} [\rho(\varphi,\xi)]$, approximated by the experimental average $R:\varphi\mapsto \frac{\sum_j \rho(\varphi, \xi_j)}{m}$, with $\varphi\in \mathcal{F}$ and $\{\xi_j\in \mathcal{T}\}_{j=1}^m$. Results for what concerns variations on the dimensionality of signals sampled from $\mu_0$ are reported in Figure \ref{fig:app:avgrob-boxplot} and Table \ref{tab:app:avg_rob}. In terms of both RE and AE, we observe that the performance of ridge regression improves until the number of retained components is roughly $250$ (which is a quarter of the original dimensionality of the training dataset), after which it stabilizes to values comparable to that of plain STL kernel regression (see Figure \ref{fig:app:avgrob-boxplot}). Performance of the interpretable case is again worse than that of the other dimensionality reported, but still it obtains a relative error $\leq 0.10$ in all tested cases, hence yielding acceptable performance. 

If we focus on a single random experiment, we verify that both relative and absolute error of stl2vec when keeping $250$ dimensions (as mentioned earlier) follow the same distribution of respectively relative and absolute error of implicit STL kernel embeddings. We visually show this in Figure \ref{fig:app:single_exp_avgrob}, and we numerically verify it using the Kolmogorov-Smirnov statistical test: providing as null hypothesis the equality between the error distribution of stl2vec($250$) and STL kernel outputs a p-value of $0.2877$ for RE and of $0.5728$ for AE.

\begin{figure*}[h!]
\begin{minipage}{0.48\linewidth}
\centering
    \includegraphics[width=0.7\linewidth]{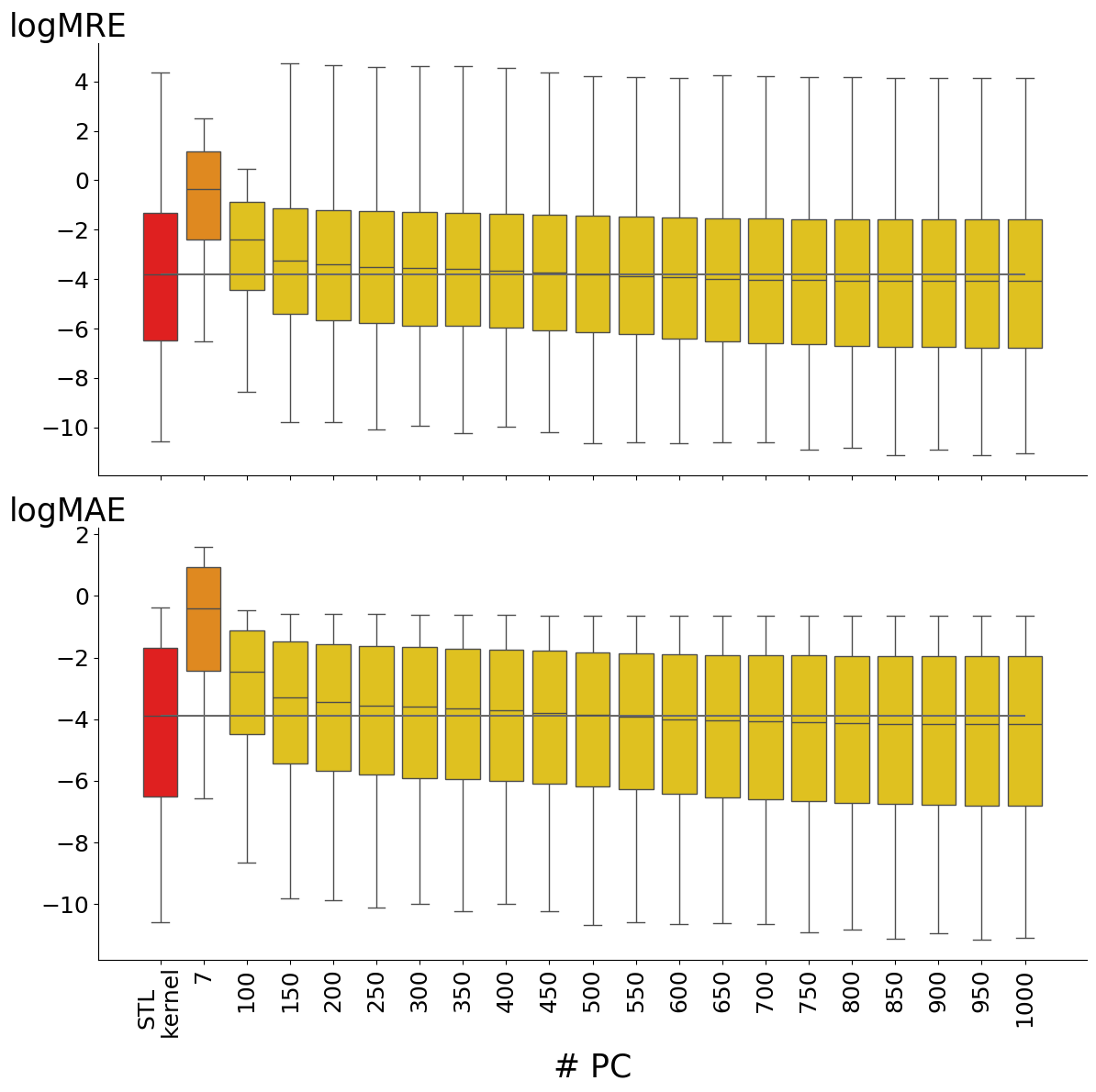}
    \vspace{0.1cm}
    \caption{Mean of the quantiles for $RE$ and $AE$ over $\num{100}$ experiments for average robustness $R$, varying the number of retained PC, on signals sampled from the SIRS model.}
    \label{fig:app:avgrob-boxplot-other}
\end{minipage}\hfill
\begin{minipage}{0.48\linewidth}
\centering    
\resizebox{0.85\linewidth}{!}{
\begin{tabular}{llllll|llll}
\toprule
{} & {} & \multicolumn{4}{c}{relative error (RE) } & \multicolumn{4}{c}{absolute error (AE)} \\
\midrule
{} &  {} &  1quart &   median &   3quart &  99perc &  1quart &   median &   3quart &   99perc \\
\midrule
{Immigration} & \multicolumn{9}{c}{} \\ 
\midrule 
{} &  STL kernel & 0.00504 & 0.01427 & 0.0478 & 1.83794 & 0.00278 & 0.00732 & 0.01928 & 0.15045\\
{} & stl2vec($\num{250}$) & 0.00624 & 0.0176 & 0.05681 & 2.08241 & 0.0034 & 0.00902 & 0.02307 & 0.15098\\
{} & stl2vec($\num{500}$) & 0.00463 & 0.01363 & 0.04695 & 1.79559 & 0.00249 & 0.00698 & 0.0191 & 0.14985\\ 
\midrule
{Isomerization} & \multicolumn{9}{c}{} \\ 
\midrule 
{} &  STL kernel & 0.00228 & 0.00815 & 0.03302 & 1.06605 & 0.00607 & 0.01942 & 0.06112 & 0.47101\\ 
{} & stl2vec($\num{250}$) & 0.00565 & 0.01511 & 0.04724 & 1.25824 & 0.01496 & 0.03756 & 0.09153 & 0.49388 \\
{} & stl2vec($\num{500}$) & 0.00221 & 0.00817 & 0.03294 & 1.05738 & 0.00592 & 0.01934 & 0.06059 & 0.46637\\ 
\midrule 
{Transcription} & \multicolumn{9}{c}{} \\ 
\midrule 
{} &  STL kernel & 0.0178 & 0.06401 & 0.22543 & 7.67721 & 0.01399 & 0.04239 & 0.11736 & 0.69872\\ 
{} & stl2vec($\num{250}$) & 0.02914 & 0.08566 & 0.27821 & 8.3278 & 0.02356 & 0.05973 & 0.14082 & 0.70263\\
{} & stl2vec($\num{500}$) & 0.01795 & 0.06545 & 0.22875 & 7.77321 & 0.01419 & 0.04347 & 0.11841 & 0.69761\\ 
\midrule
{SIRS} & \multicolumn{9}{c}{} \\ 
\midrule 
{} &  STL kernel & 0.00629 & 0.02209 & 0.07593 & 1.19013 & 0.00608 & 0.02052 & 0.06493 & 0.43494\\ 
{} & stl2vec($\num{250}$) & 0.01162 & 0.03026 & 0.08979 & 1.28718 & 0.01129 & 0.02868 & 0.07669 & 0.38096\\
{} & stl2vec($\num{500}$) & 0.00822 & 0.02235 & 0.06859 & 1.0287 & 0.00797 & 0.021 & 0.05864 & 0.348011\\ 
\bottomrule 
\end{tabular}
}
\vspace{0.1cm}
\captionof{table}{Mean of quantiles for RE and AE over $\num{100}$ experiments for prediction of average robustness $R$, changing stochastic model.}
\label{tab:app:avgrob-other}
\end{minipage}
\vspace{0.5cm}
\end{figure*}

For what concerns tests done on trajectories coming from other stochastic processes, the same considerations hold and results are reported in Table \ref{tab:app:avgrob-other} and Figure \ref{fig:app:avgrob-boxplot-other}. 

\begin{figure*}[h!]
\begin{minipage}{0.48\linewidth}
    \centering
    \includegraphics[width=0.7\linewidth]{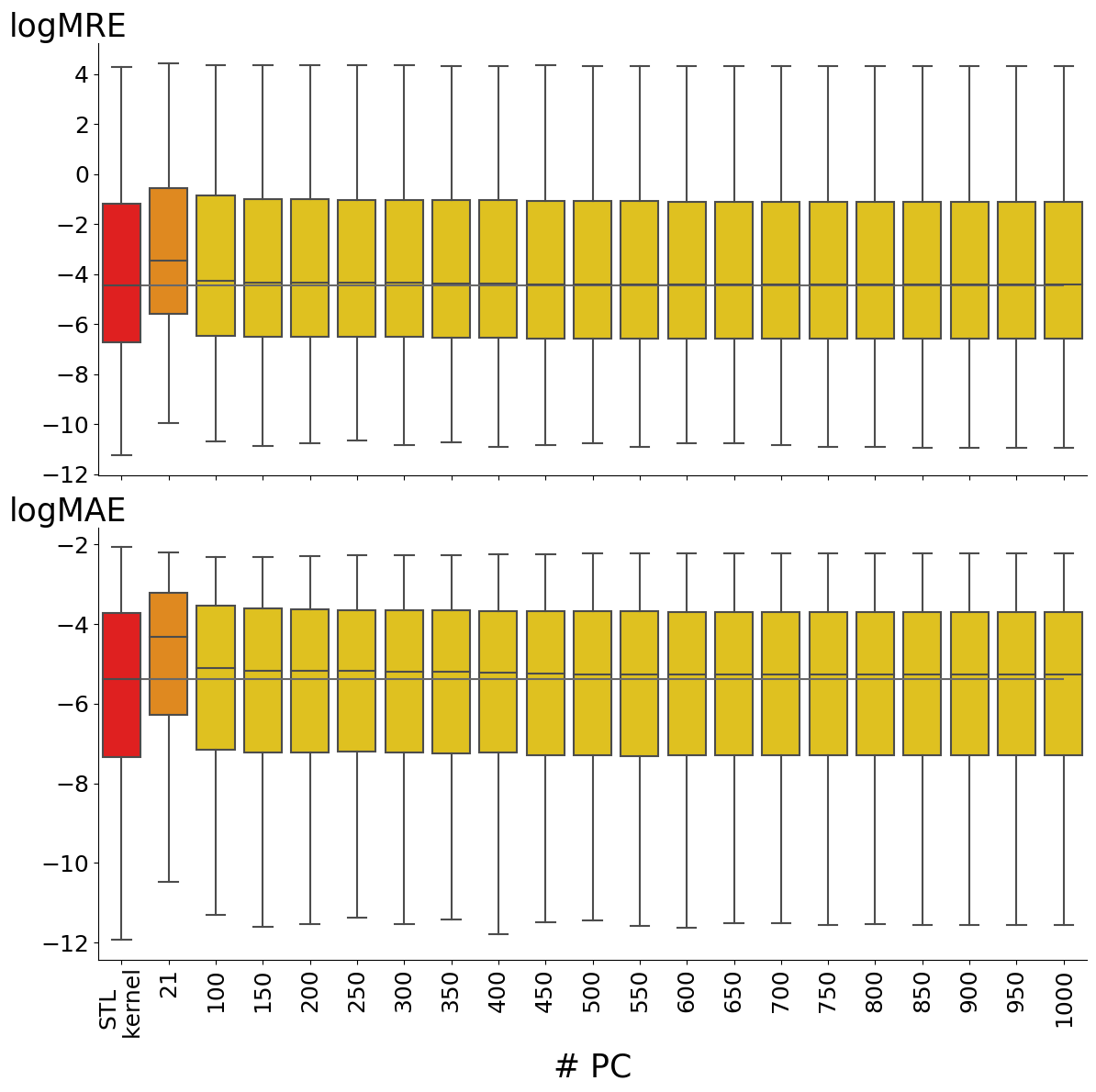}
    \vspace{0.1cm}
    \caption{Mean of the quantiles for $RE$ and $AE$ over $\num{100}$ experiments for predicting satisfaction probability $S$, varying the number of retained PC, on signals of dimension $10$.}
    \label{fig:app:satprob-boxplot}
\end{minipage}\hfill
\begin{minipage}{0.48\linewidth}
\centering    
\centering    
\resizebox{0.85\linewidth}{!}{
\begin{tabular}{llllll|llll}
\toprule
{} & {} & \multicolumn{4}{c}{relative error (RE) } & \multicolumn{4}{c}{absolute error (AE)} \\
\midrule
{} &  {} &  1quart &   median &   3quart &  99perc &  1quart &   median &   3quart &   99perc \\
\midrule
{n=$3$} & \multicolumn{9}{c}{} \\ 
\midrule 
{} &  STL kernel & 0.00447  & 0.01164  & 0.03653 & 11.18914 & 0.00198 & 0.00466 & 0.00925 & 0.06686\\ 
{} & stl2vec($7$) & 0.01330&  0.03164 & 0.07862 & 28.37474 & 0.00641 & 0.01325 & 0.02253 & 0.08385\\
{} & stl2vec($\num{250}$)  & 0.00538  & 0.01298  & 0.03743 & 28.58705 & 0.00258 & 0.00564 & 0.01074 & 0.06254\\
{} & stl2vec($\num{500}$) & 0.00496  & 0.01203  & 0.03567 & 27.12305 & 0.00239 & 0.00522 & 0.01016 & 0.06321\\ 
\midrule 
{n=$4$} & \multicolumn{9}{c}{} \\ 
\midrule 
{} &  STL kernel & 0.00455 & 0.01192 & 0.03766 & 6.38769 & 0.00211 & 0.00480& 0.00956 & 0.06398\\ 
{} & stl2vec($9$) &  0.01056  & 0.02475  & 0.06345 & 24.70682 & 0.00509 & 0.01070& 0.01865 & 0.07369\\
{} & stl2vec($\num{250}$) & 0.00521  & 0.01280& 0.03613 & 23.54337 & 0.00255 & 0.00560& 0.01054 & 0.05823\\
{} & stl2vec($\num{500}$) & 0.00496  & 0.01230& 0.03480& 23.12045 & 0.00245 & 0.00536 & 0.01003 & 0.05825\\ 
\midrule
{n=$5$} & \multicolumn{9}{c}{} \\ 
\midrule 
{} &  STL kernel & 0.00483 & 0.01270& 0.03888 & 4.24863 & 0.00218 & 0.00506 & 0.01005 & 0.06238\\ 
{} & stl2vec($11$) & 0.00917  & 0.02198  & 0.05864 & 23.14694 & 0.00445 & 0.00957 & 0.01685 & 0.06820\\
{} & stl2vec($\num{250}$) & 0.00540& 0.01314  & 0.03614 & 23.80843 & 0.00259 & 0.00573 & 0.01079 & 0.05743\\
{} & stl2vec($\num{500}$) & 0.00524  & 0.01274  & 0.03511 & 21.51802 & 0.00255 & 0.00553 & 0.01037 & 0.05648\\ 
\midrule
{n=$8$} & \multicolumn{9}{c}{} \\ 
\midrule 
{} &  STL kernel & 0.00521 & 0.01371 & 0.04177 & 2.08428 & 0.00243 & 0.00552 & 0.01094 & 0.05364 \\ 
{} & stl2vec($17$) & 0.00801  & 0.01963  & 0.05118 & 19.19273 & 0.00395 & 0.00842 & 0.01513 & 0.05591\\
{} & stl2vec($\num{250}$) & 0.00607  & 0.01462  & 0.03953 & 16.94946 & 0.00298 & 0.00637 & 0.01174 & 0.05233\\
{} & stl2vec($\num{500}$) & 0.00546  & 0.01321  & 0.03656 & 14.15122 & 0.00266 & 0.00579 & 0.01066 & 0.04996\\ 
\midrule
{n=$10$} & \multicolumn{9}{c}{} \\ 
\midrule 
{} &  STL kernel & 0.00524 & 0.01407 & 0.04272 & 1.66716 & 0.00248 & 0.00566 & 0.01119 & 0.0521 \\ 
{} & stl2vec($21$) & 0.00825  & 0.01963 & 0.05107 & 18.66501 & 0.00409 & 0.00855 & 0.01534 & 0.05164\\
{} & stl2vec($\num{250}$) & 0.00622  & 0.01500& 0.03963 & 16.56051 & 0.00301 & 0.00659 & 0.01208 & 0.04833\\
{} & stl2vec($\num{500}$) & 0.00558  & 0.01353  & 0.03687 & 13.21197 & 0.00271 & 0.00592 & 0.01092 & 0.04641\\ 
\bottomrule
\end{tabular}
}
\vspace{0.1cm}
\captionof{table}{Mean of quantiles for RE and AE over $\num{100}$ experiments for prediction of satisfaction probability $S$, changing the number of variables in the dataset of formulae.}
\label{tab:app:sat_prob}
\end{minipage}
\vspace{0.5cm}
\end{figure*}

\paragraph{Satisfaction Probability} experiments consist in learning the function $\mathbb E_{\xi\sim\mu_0} [s(\varphi,\xi)]$, approximated by the statistical average $S:\varphi\mapsto \frac{\sum_j s(\varphi, \xi_j)}{m}$, with $\varphi\in \mathcal{F}$ and $\{\xi_j\in \mathcal{T}\}_{j=1}^m$. Results for what concerns variations on the dimensionality of signals sampled from $\mu_0$ are reported in Figure \ref{fig:app:satprob-boxplot} and Table \ref{tab:app:sat_prob}. For both RE and AE, we observe that the performance of ridge regression improves until the number of retained components is roughly $300$ (which is less than a third of the original dimensionality of the training dataset), after which it stabilizes to values comparable to that of plain STL kernel regression (see Figure \ref{fig:app:satprob-boxplot}). Performance of the interpretable case is again worse than that of the other dimensionality reported, with a relative error $\sim 0.25$ in all tested cases.

If we focus on a single random experiment, we verify that both relative and absolute error of stl2vec when keeping $300$ dimensions (as mentioned earlier) follow the same distribution of respectively relative and absolute error of implicit STL kernel embeddings. We visually show this in Figure \ref{fig:app:single_exp_satprob}, and we numerically verify it using the Kolmogorov-Smirnov statistical test: providing as null hypothesis the equality between the error distribution of stl2vec($300$) and STL kernel outputs a p-value of $0.9542$ for RE and of $0.8881$ for AE.

\begin{figure*}[h!]
\begin{minipage}{0.48\linewidth}
\centering
    \includegraphics[width=0.7\linewidth]{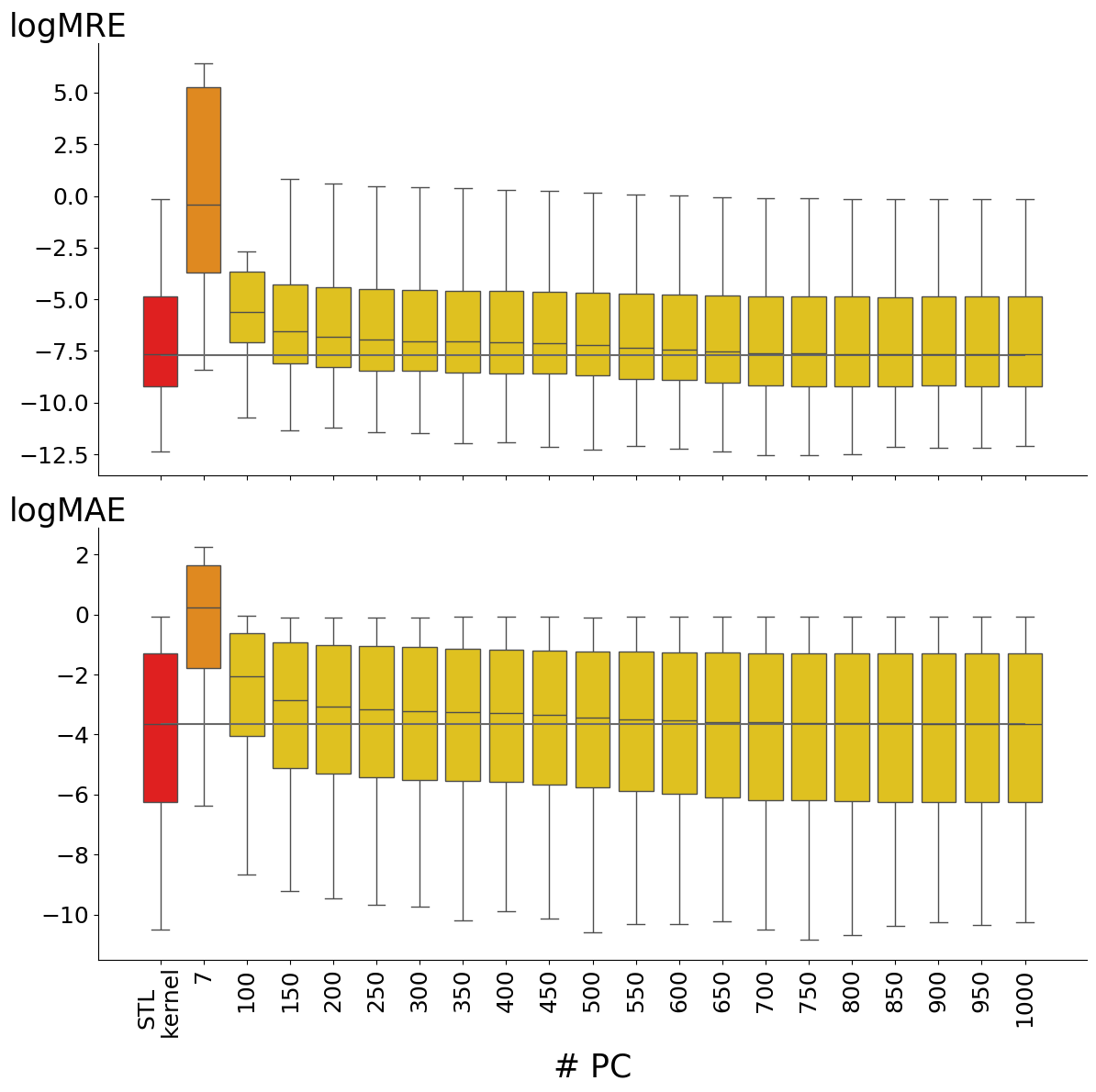}
    \vspace{0.1cm}
    \caption{Mean of the quantiles for $RE$ and $AE$ over $\num{100}$ experiments for predicting satisfaction probability $S$, varying the number of retained PC, on signals sampled from the SIRS model.}
    \label{fig:app:satprob-boxplot-other}
\end{minipage}\hfill
\begin{minipage}{0.48\linewidth}
\centering    
\resizebox{0.85\linewidth}{!}{
\begin{tabular}{llllll|llll}
\toprule
{} & {} & \multicolumn{4}{c}{relative error (RE) } & \multicolumn{4}{c}{absolute error (AE)} \\
\midrule
{} &  {} &  1quart &   median &   3quart &  99perc &  1quart &   median &   3quart &   99perc \\
\midrule
{Immigration} & \multicolumn{9}{c}{} \\ 
\midrule 
{} &  STL kernel & 0.00301 & 0.0248 & 1.29873 & 5.70658 & 0.0002 & 0.00061 & 0.00138 & 0.00981\\
{} & stl2vec($\num{250}$) & 0.00313 & 0.02495 & 1.4116 & 5.93454 & 0.00021 & 0.00063 & 0.00141 & 0.0098\\
{} & stl2vec($\num{500}$) & 0.00301 & 0.02465 & 1.2887 & 5.72107 & 0.0002 & 0.00061 & 0.00138 & 0.00979\\ 
\midrule
{Isomerization} & \multicolumn{9}{c}{} \\ 
\midrule 
{} &  STL kernel & 0.00285 & 0.04328 & 2.02743 & 8.37949 & 0.00021 & 0.00057 & 0.00167 & 0.01114\\ 
{} & stl2vec($\num{250}$) & 0.00367 & 0.04723 & 2.80182 & 9.52695 & 0.00031 & 0.00074 & 0.00187 & 0.01101 \\
{} & stl2vec($\num{500}$) & 0.00288 & 0.0433 & 2.0656 & 8.47736 & 0.00021 & 0.00058 & 0.00167 & 0.01109\\ 
\midrule 
{Transcription} & \multicolumn{9}{c}{} \\ 
\midrule 
{} &  STL kernel & 0.0178 & 0.06401 & 0.22543 & 7.67721 & 0.01399 & 0.04239 & 0.11736 & 0.69872\\ 
{} & stl2vec($\num{250}$) & 0.02914 & 0.08566 & 0.27821 & 8.3278 & 0.02356 & 0.05973 & 0.14082 & 0.70263\\
{} & stl2vec($\num{500}$) & 0.01795 & 0.06545 & 0.22875 & 7.77321 & 0.01419 & 0.04347 & 0.11841 & 0.69761\\ 
\midrule
{SIRS} & \multicolumn{9}{c}{} \\ 
\midrule 
{} &  STL kernel & 0.00629 & 0.02209 & 0.07593 & 1.19013 & 0.00608 & 0.02052 & 0.06493 & 0.43494\\ 
{} & stl2vec($\num{250}$) & 0.01162 & 0.03026 & 0.08979 & 1.28718 & 0.01129 & 0.02868 & 0.07669 & 0.38096\\
{} & stl2vec($\num{500}$) & 0.00822 & 0.02235 & 0.06859 & 1.0287 & 0.00797 & 0.021 & 0.05864 & 0.348011\\ 
\bottomrule 
\end{tabular}
}
\vspace{0.1cm}
\captionof{table}{Mean of quantiles for RE and AE over $\num{100}$ experiments for prediction of satisfaction probability $S$, changing stochastic model.}
\label{tab:app:satprob-other}
\end{minipage}
\vspace{0.5cm}
\end{figure*}

For what concerns tests done on trajectories coming from other stochastic processes, the same considerations hold and results are reported in Table \ref{tab:app:satprob-other} and Figure \ref{fig:app:satprob-boxplot-other}.

\subsection{More Results on Conditional Generation of Trajectories}\label{subsec:app:cvae-results}

\paragraph{Ablation on the number of components} to keep for injecting the semantics of the requirements inside the Conditional Variational Autoencoder (CVAE) are carried out. Here, we report results of extensive experiments we have performed to assess the use of stl2vec semantic representations as conditioning vectors inside a CVAE model. After a hyperparameter tuning phase, we established a CVAE architecture composed of a $6$-layered $1$d convolutional encoder, having dimensions $[400, 400, 256, 256, 128, 64]$, latent space of dimension $32$ and decoder composed of $6$ layers of $1$d transposed convolutions, having the same dimensions of the encoder, but in reverse order. The decoder's last layer is a $1$-d convolution that maps the decoded signals back to their original dimensions. Between (de)convolutional layers (except the last) batch normalization and Rectified Linear Unit (ReLU) activation layers are present; semantic conditioning vectors are projected into the appropriate space using $1$-layer Feed Forward Network with ReLU activation, and summed to either the convolved input trajectories (in the encoder), or the latent vector (in the decoder). 

We train the network using the Adam optimizer \cite{adam} with a learning rate of $\num{0.0005}$ for $\num{800}$ epochs, adopting a cyclic annealing schedule for $\beta$ of Equation (\ref{eq:cvae-loss}) \cite{cyclic-beta}. Signals have been scaled in $[-\num{1}, \num{1}]$ in training for enhancing stability, and scaled back during the evaluation phase.

We investigate the goodness of our CVAE model varying the number of retained dimension in stl2vec representations, and also using the plain STL kernel embedding as conditioner. As performance indexes, we use average robustness $R$ and satisfaction probability $S$ of each test formula on a batch of $1000$ decoded trajectories. In each setting we perform $30$ experiments, and we report the mean quantiles of the distribution of the predicted average robustness $R_{\mathit{cond}}$ and the predicted satisfaction probability $S_{\mathit{cond}}$ (on independent test sets of $300$ formulae). 

\begin{figure*}[h!]
\begin{minipage}{0.48\linewidth}
    \centering
    \includegraphics[width=\linewidth]{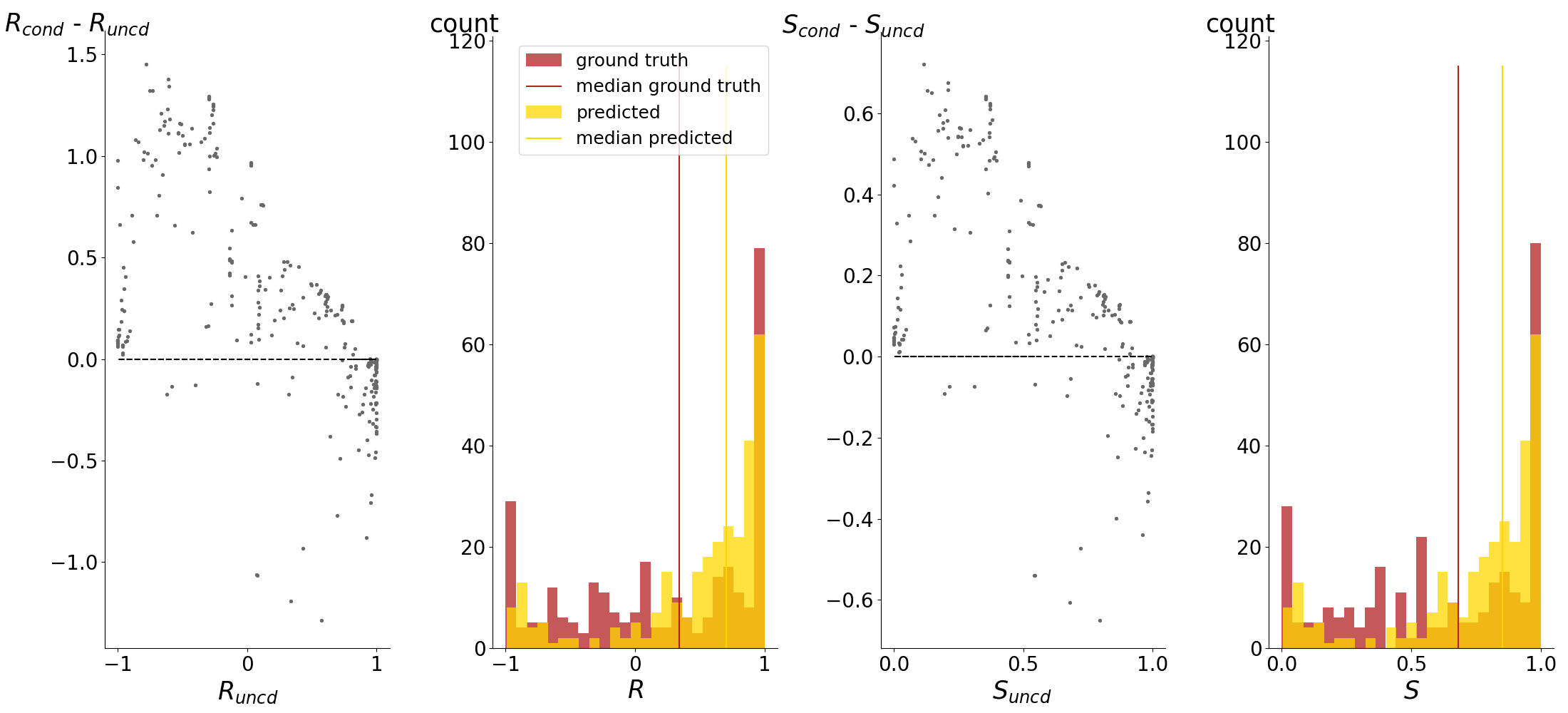}
    \vspace{0.1cm}
    \caption{Results of a random experiment for the conditional generation of trajectories using CVAE, in terms of average robustness and satisfaction probability, conditioning on stl2vec embeddings of dimension $10$.}
    \label{fig:app:cvae-10}
\end{minipage}\hfill
\begin{minipage}{0.48\linewidth}
\centering
\resizebox{\linewidth}{!}{
\begin{tabular}{cccccc}
    \toprule
    & 1perc & 1quart & median & 3quart & 99perc\\
    \midrule
    $R_{\mathit{uncd}}$  & -0.9994 $\pm$ 0.0004 & -0.5128 $\pm$ 0.0123 & 0.0869 $\pm$ 0.0104 & 0.7321 $\pm$ 0.0046 & 1.0 $\pm$ 0.0\\
    $R_{\mathit{cond}}$  & -0.9357   $\pm$ 0.0005& 0.3018  $\pm$ 0.0076 & 0.7008  $\pm$ 0.0083 & 0.8935  $\pm$ 0.0027 & 1.0  $\pm$ 0.0 \\
    \midrule 
    $S_{\mathit{uncd}}$  & 3.23e-04 $\pm$ 0.0003 & 0.2290 $\pm$ 0.0076& 0.5243 $\pm$ 0.0051 & 0.8122 $\pm$ 0.0022 & 1.0 $\pm$ 0.0 \\
    $S_{\mathit{cond}}$  & 0.0321  $\pm$ 0.0027 & 0.6509  $\pm$ 0.0038 & 0.8305  $\pm$ 0.0042 & 0.9468  $\pm$ 0.0013 & 1.0  $\pm$ 0.0 \\
    \bottomrule
\end{tabular}
}
\vspace{0.1cm}
\captionof{table}{Mean and standard deviation of quantiles of the distributions of $R_{\mathit{uncd}}$ (resp. $S_{\mathit{uncd}}$) and $R_{\mathit{cond}}$ (resp. $S_{\mathit{cond}}$), over $\num{300}$ test formulae, averaged over $30$ experiments, conditioning on stl2vec embeddings of dimension $10$.}
\label{tab:app:cvae-quantiles-10}
\end{minipage}
\vspace{0.5cm}
\end{figure*}

\begin{figure*}
\begin{minipage}{0.48\linewidth}
    \centering
    \includegraphics[width=\linewidth]{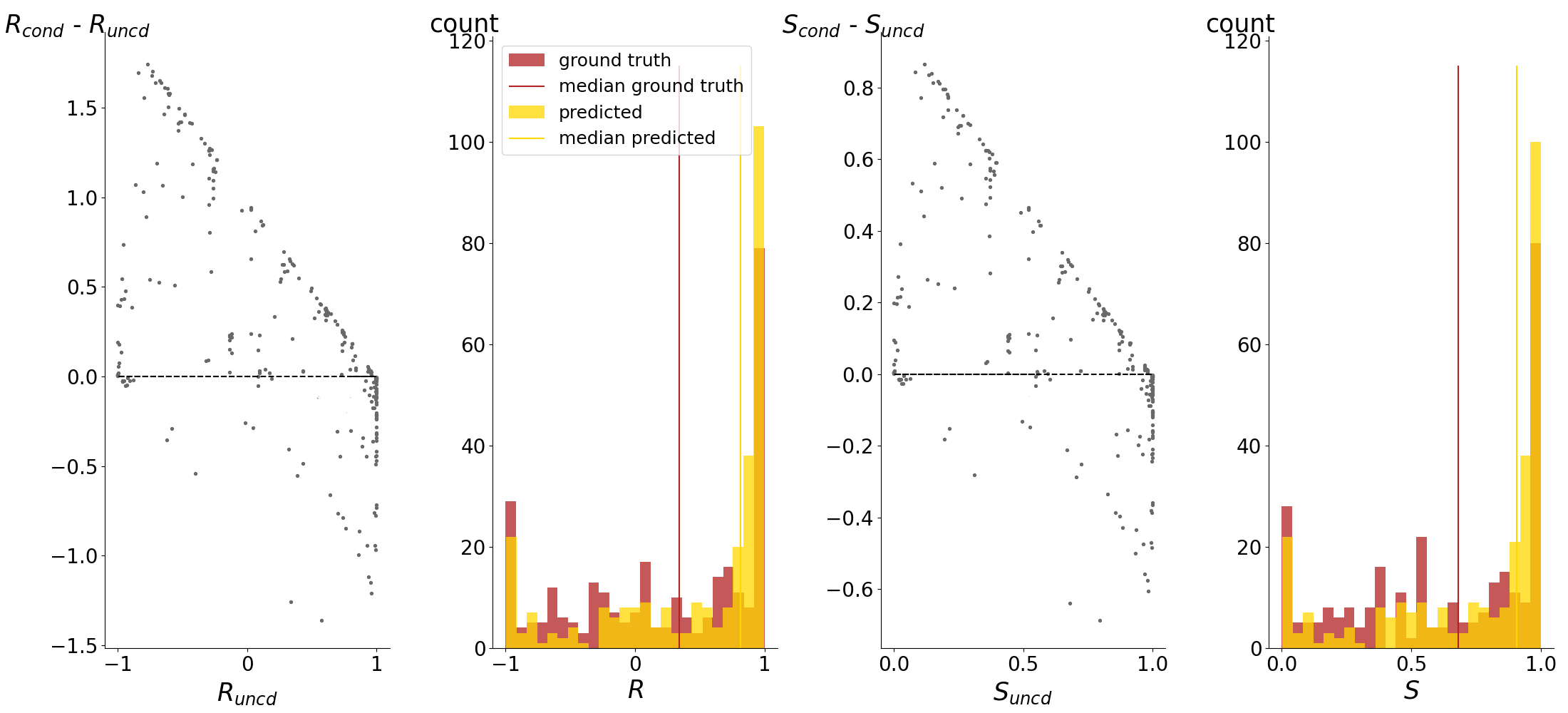}
    \vspace{0.1cm}
    \caption{Results of a random experiment for the conditional generation of trajectories using CVAE, in terms of average robustness and satisfaction probability, conditioning on stl2vec embeddings of dimension $50$.}
    \label{fig:app:cvae-50}
\end{minipage}\hfill
\begin{minipage}{0.48\linewidth}
\centering
\resizebox{\linewidth}{!}{
\begin{tabular}{cccccc}
    \toprule
    & 1perc & 1quart & median & 3quart & 99perc\\
    \midrule
    $R_{\mathit{uncd}}$  & -0.9994 $\pm$ 0.0004 & -0.5128 $\pm$ 0.0123 & 0.0869 $\pm$ 0.0104 & 0.7321 $\pm$ 0.0046 & 1.0 $\pm$ 0.0 \\
    $R_{\mathit{cond}}$  & -0.9905  $\pm$ 0.0016 & 0.0493  $\pm$ 0.0134 & 0.8134  $\pm$ 0.0043 & 0.9684  $\pm$ 0.0014&0.9933  $\pm$ 0.0010\\
    \midrule 
    $S_{\mathit{uncd}}$  & 3.23e-04 $\pm$ 0.0003 & 0.2290 $\pm$ 0.0076& 0.5243 $\pm$ 0.0051 & 0.8122 $\pm$ 0.0022 & 1.0 $\pm$ 0.0\\
    $S_{\mathit{cond}}$  & 0.0048  $\pm$ 0.0008 & 0.5247  $\pm$ 0.0067 & 0.9065  $\pm$ 0.0022 & 0.9843  $\pm$ 0.0007 & 0.9983  $\pm$ 0.0006 \\
    \bottomrule
\end{tabular}
}
    \vspace{0.1cm}
\captionof{table}{Mean and standard deviation of quantiles of the distributions of $R_{\mathit{uncd}}$ (resp. $S_{\mathit{uncd}}$) and $R_{\mathit{cond}}$ (resp. $S_{\mathit{cond}}$), over $\num{300}$ test formulae, averaged over $30$ experiments, conditioning on stl2vec embeddings of dimension $50$.}
\label{tab:app:cvae-quantiles-50}
\end{minipage}
    \vspace{0.5cm}
\end{figure*}

\begin{figure*}
\begin{minipage}{0.48\linewidth}
    \centering
    \includegraphics[width=\linewidth]{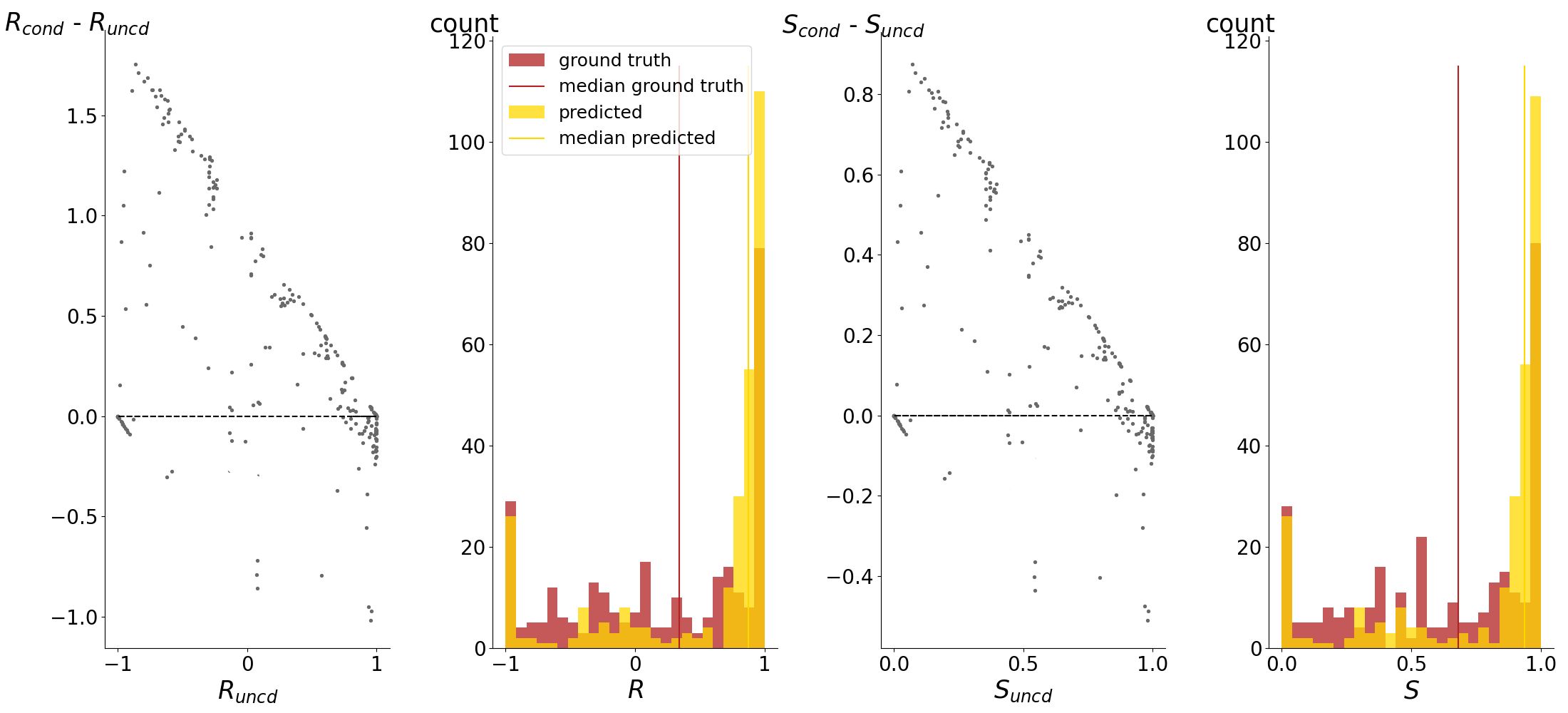}
    \vspace{0.1cm}
    \caption{Results of a random experiment for the conditional generation of trajectories using CVAE, in terms of average robustness and satisfaction probability, conditioning on stl2vec embeddings of dimension $100$.}
    \label{fig:app:cvae-100}
\end{minipage}\hfill
\begin{minipage}{0.48\linewidth}
\centering
\resizebox{\linewidth}{!}{
\begin{tabular}{cccccc}
    \toprule
    & 1perc & 1quart & median & 3quart & 99perc\\
    \midrule
    $R_{\mathit{uncd}}$  & -0.9994 $\pm$ 0.0004 & -0.5128 $\pm$ 0.0123 & 0.0869 $\pm$ 0.0104 & 0.7321 $\pm$ 0.0046 & 1.0 $\pm$ 0.0 \\
    $R_{\mathit{cond}}$  & -1.0  $\pm$ 0.0&  0.3015 $\pm$ 0.0113 & 0.8737 $\pm$ 0.0033 & 0.9667  $\pm$ 0.0024 & 1.0  $\pm$ 0.0 \\
    \midrule 
    $S_{\mathit{uncd}}$  & 3.23e-04 $\pm$ 0.0003 & 0.2290 $\pm$ 0.0076& 0.5243 $\pm$ 0.0051 & 0.8122 $\pm$ 0.0022 & 1.0 $\pm$ 0.0\\
    $S_{\mathit{cond}}$  & 0.0  $\pm$ 0.0 & 0.6508  $\pm$ 0.0057 & 0.9363  $\pm$ 0.0017 & 0.9927  $\pm$ 0.0010& 1.0  $\pm$ 0.0 \\
    \bottomrule
\end{tabular}
}
    \vspace{0.1cm}
\captionof{table}{Mean and standard deviation of quantiles of the distributions of $R_{\mathit{uncd}}$ (resp. $S_{\mathit{uncd}}$) and $R_{\mathit{cond}}$ (resp. $S_{\mathit{cond}}$), over $\num{300}$ test formulae, averaged over $30$ experiments, conditioning on stl2vec embeddings of dimension $100$.}
\label{tab:app:cvae-quantiles-100}
\end{minipage}
    \vspace{0.5cm}
\end{figure*}

\begin{figure*}
\begin{minipage}{0.48\linewidth}
    \centering
    \includegraphics[width=\linewidth]{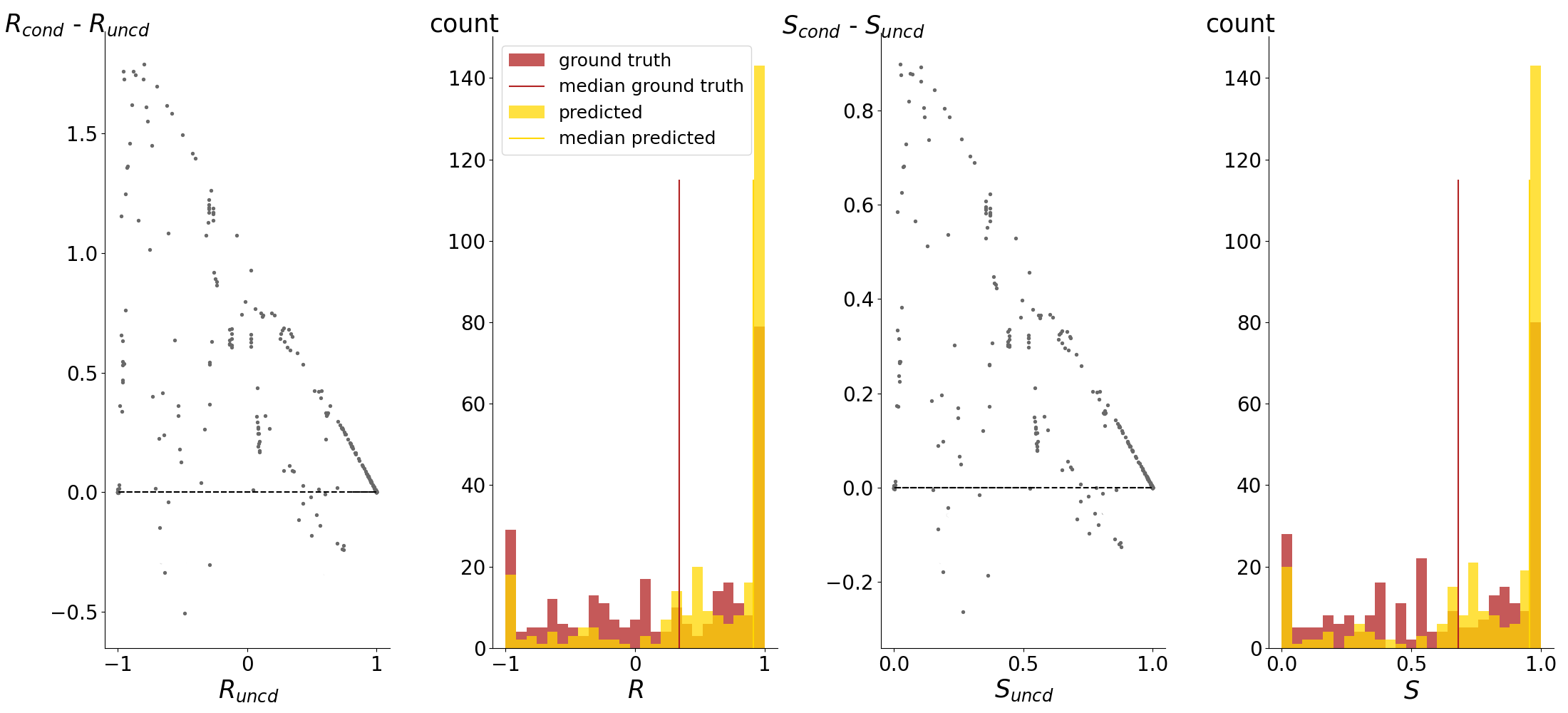}
    \vspace{0.1cm}
    \caption{Results of a random experiment for the conditional generation of trajectories using CVAE, in terms of average robustness and satisfaction probability, conditioning on stl2vec embeddings of dimension $250$.}
    \label{fig:app:cvae-250}
\end{minipage}
\begin{minipage}{0.48\linewidth}
\centering
\resizebox{\linewidth}{!}{
\begin{tabular}{cccccc}
    \toprule
    & 1perc & 1quart & median & 3quart & 99perc\\
    \midrule
    $R_{\mathit{uncd}}$  & -0.9994 $\pm$ 0.0004 & -0.5128 $\pm$ 0.0123 & 0.0869 $\pm$ 0.0104 & 0.7321 $\pm$ 0.0046 & 1.0 $\pm$ 0.0 \\
    $R_{\mathit{cond}}$  & -1.0  $\pm$ 0.0 & -0.6157  $\pm$ 0.0086 & 0.9030  $\pm$ 0.0043& 1.0  $\pm$ 0.0 & 1.0  $\pm$ 0.0 \\
    \midrule 
    $S_{\mathit{uncd}}$  & 3.23e-04 $\pm$ 0.0003 & 0.2290 $\pm$ 0.0076& 0.5243 $\pm$ 0.0051 & 0.8122 $\pm$ 0.0022 & 1.0 $\pm$ 0.0\\
    $S_{\mathit{cond}}$  & 0.0 $\pm$ 0.0 & 0.1923 $\pm$ 0.0045 & 0.9515 $\pm$ 0.0021 & 1.0  $\pm$ 0.0 & 1.0  $\pm$ 0.0\\
    \bottomrule
\end{tabular}
}
    \vspace{0.1cm}
\captionof{table}{Mean and standard deviation of quantiles of the distributions of $R_{\mathit{uncd}}$ (resp. $S_{\mathit{uncd}}$) and $R_{\mathit{cond}}$ (resp. $S_{\mathit{cond}}$), over $\num{300}$ test formulae, averaged over $30$ experiments, conditioning on stl2vec embeddings of dimension $250$.}
\label{tab:app:cvae-quantiles-250}
\end{minipage}
    \vspace{0.5cm}
\end{figure*}

\begin{figure*}
\begin{minipage}{0.48\linewidth}
    \centering
    \includegraphics[width=\linewidth]{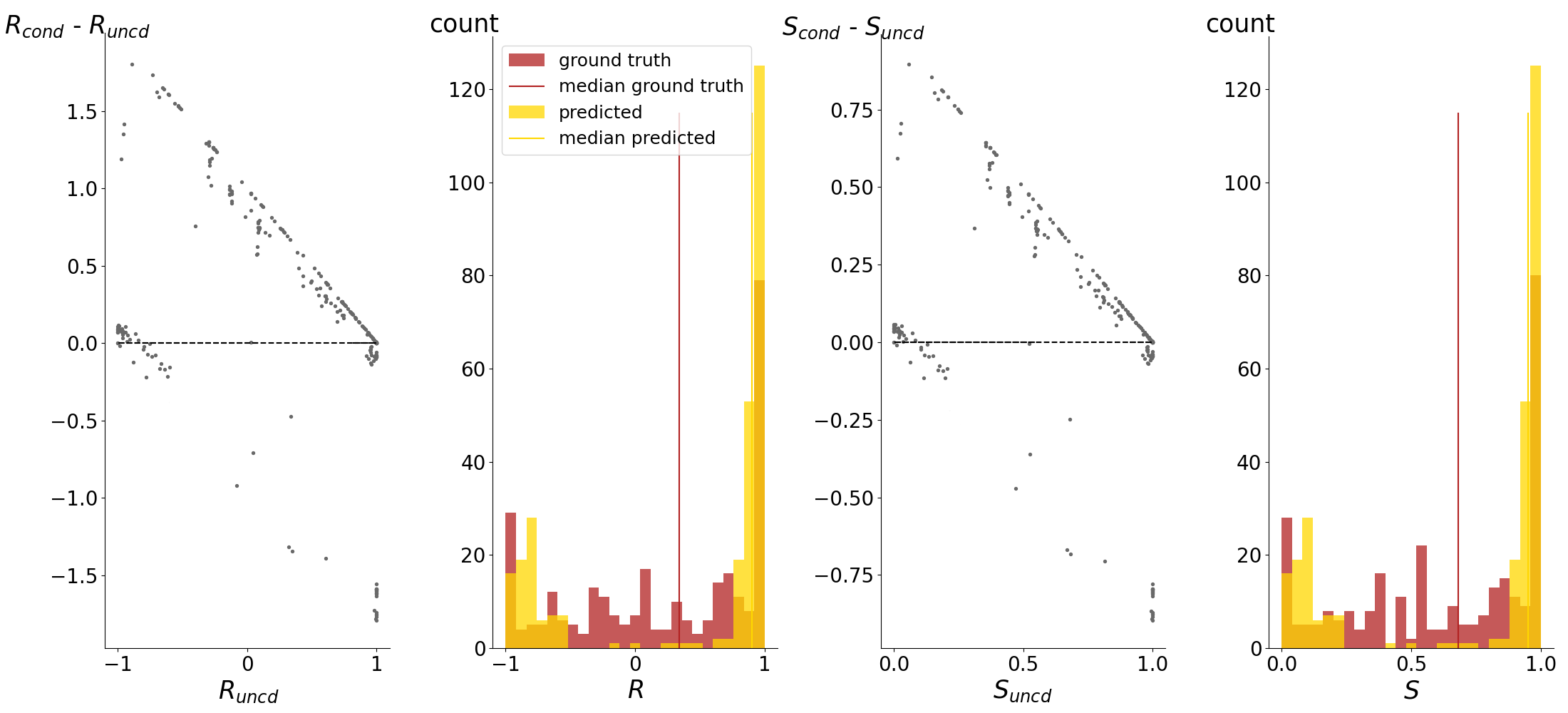}
    \vspace{0.1cm}
    \caption{Results of a random experiment for the conditional generation of trajectories using CVAE, in terms of average robustness and satisfaction probability, conditioning on stl2vec embeddings of dimension $500$.}
    \label{fig:app:cvae-500}
\end{minipage}\hfill
\begin{minipage}{0.48\linewidth}
\centering
\resizebox{\linewidth}{!}{
\begin{tabular}{cccccc}
    \toprule
    & 1perc & 1quart & median & 3quart & 99perc\\
    \midrule
    $R_{\mathit{uncd}}$  & -0.9994 $\pm$ 0.0004 & -0.5128 $\pm$ 0.0123 & 0.0869 $\pm$ 0.0104 & 0.7321 $\pm$ 0.0046 & 1.0 $\pm$ 0.0\\
    $R_{\mathit{cond}}$  & -1.0  $\pm$ 0.0 & -0.6282  $\pm$ 0.0087 & 0.9023  $\pm$ 0.0026 & 1.0  $\pm$ 0.0 & 1.0  $\pm$ 0.0 \\
    \midrule 
    $S_{\mathit{uncd}}$  & 3.23e-04 $\pm$ 0.0003 & 0.2290 $\pm$ 0.0076& 0.5243 $\pm$ 0.0051 & 0.8122 $\pm$ 0.0022 & 1.0 $\pm$ 0.0\\
    $S_{\mathit{cond}}$  & 0.0  $\pm$ 0.0 & 0.1859  $\pm$ 0.0044 & 0.9510  $\pm$ 0.0013& 1.0  $\pm$ 0.0 & 1.0  $\pm$ 0.0 \\
    \bottomrule
\end{tabular}
}
    \vspace{0.1cm}
\captionof{table}{Mean and standard deviation of quantiles of the distributions of $R_{\mathit{uncd}}$ (resp. $S_{\mathit{uncd}}$) and $R_{\mathit{cond}}$ (resp. $S_{\mathit{cond}}$), over $\num{300}$ test formulae, averaged over $30$ experiments, conditioning on stl2vec embeddings of dimension $500$.}
\label{tab:app:cvae-quantiles-500}
\end{minipage}
    \vspace{0.5cm}
\end{figure*}

\begin{figure*}
\begin{minipage}{0.48\linewidth}
    \centering
    \includegraphics[width=\linewidth]{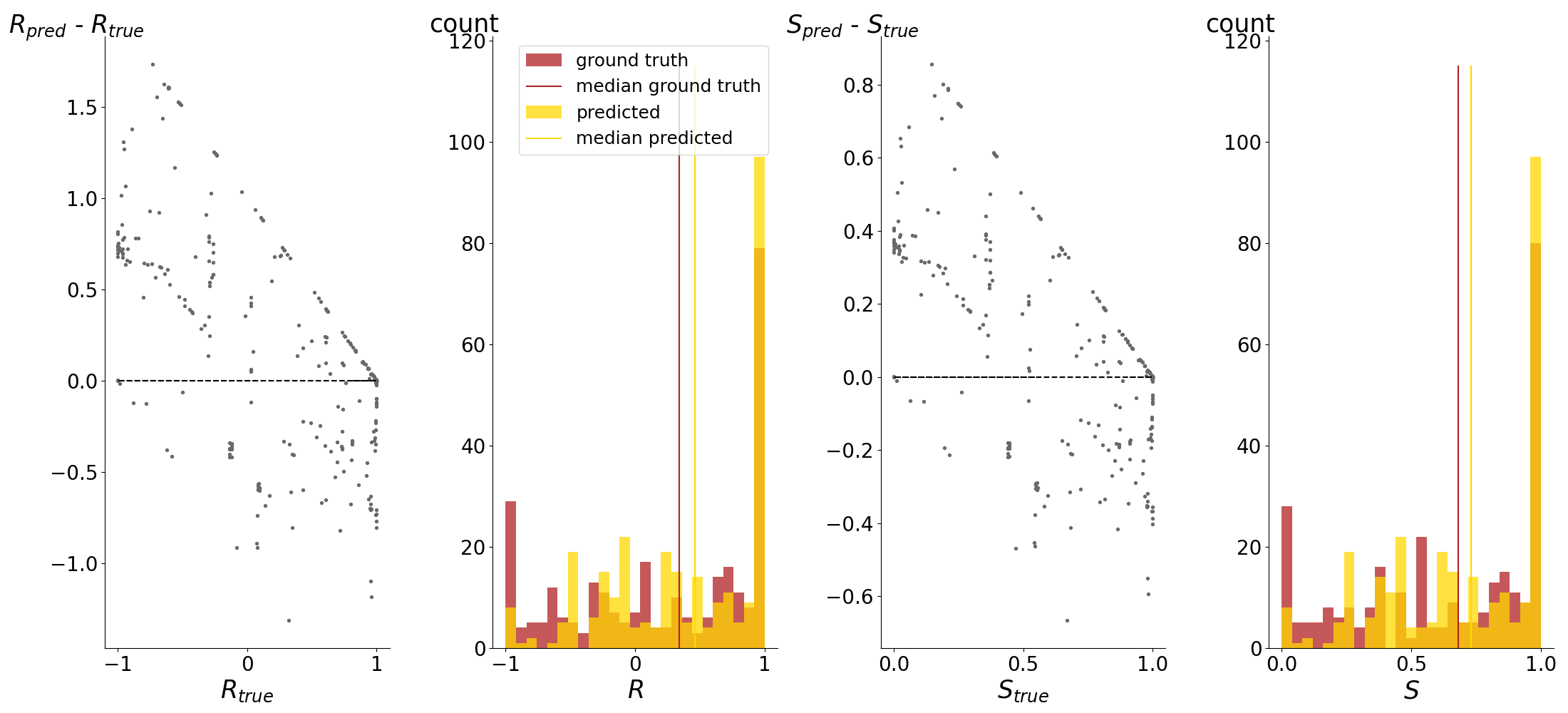}
    \vspace{0.1cm}
    \caption{Results of a random experiment for the conditional generation of trajectories using CVAE, in terms of average robustness and satisfaction probability, conditioning on STL kernel embeddings of dimension $1000$.}
    \label{fig:app:cvae-full}
\end{minipage}\hfill
\begin{minipage}{0.48\linewidth}
\centering
\resizebox{\linewidth}{!}{
\begin{tabular}{cccccc}
    \toprule
    & 1perc & 1quart & median & 3quart & 99perc\\
    \midrule
    $R_{\mathit{uncd}}$  & -0.9994 $\pm$ 0.0004 & -0.5128 $\pm$ 0.0123 & 0.0869 $\pm$ 0.0104 & 0.7321 $\pm$ 0.0046 & 1.0 $\pm$ 0.0\\
    $R_{\mathit{cond}}$  & -0.9999  $\pm$ 0.0004 & -0.0846  $\pm$ 0.0108 & 0.4657  $\pm$ 0.0099 & 0.9969  $\pm$ 0.0009 & 1.0  $\pm$ 0.0 \\
    \midrule 
    $S_{\mathit{uncd}}$  & 3.23e-04 $\pm$ 0.0003 & 0.2290 $\pm$ 0.0076& 0.5243 $\pm$ 0.0051 & 0.8122 $\pm$ 0.0022 & 1.0 $\pm$ 0.0\\
    $S_{\mathit{cond}}$  & 0.0001 $\pm$ 0.0002 & 0.4577  $\pm$ 0.0054 & 0.7300  $\pm$ 0.005& 0.9984  $\pm$ 0.0005 & 1.0 $\pm$ 0.0 \\
    \bottomrule
\end{tabular}
}
    \vspace{0.1cm}
\captionof{table}{Mean and standard deviation of quantiles of the distributions of $R_{\mathit{uncd}}$ (resp. $S_{\mathit{uncd}}$) and $R_{\mathit{cond}}$ (resp. $S_{\mathit{cond}}$), over $\num{300}$ test formulae, averaged over $30$ experiments, conditioning on STL kernel embeddings of dimension $1000$.}
\label{tab:app:cvae-quantiles-full}
\end{minipage}
    \vspace{0.5cm}
\end{figure*}

In \cref{tab:app:cvae-quantiles-10,tab:app:cvae-quantiles-50,tab:app:cvae-quantiles-100,tab:app:cvae-quantiles-250,tab:app:cvae-quantiles-500} we report such statistics for stl2vec embeddings of dimension $[10, 50, 100, 250, 500]$, respectively, while in table \ref{tab:app:cvae-quantiles-full} we test the performance of full implicit STL kernel embeddings in the same setting. We compare $R_{\mathit{cond}}$ and $S_{\mathit{cond}}$ against average robustness and satisfaction probability of each of our test formulae on a bunch of $10000$ trajectories randomly sampled form the SIRS model (denoted as $R_{\mathit{uncd}}$ and $S_{\mathit{uncd}}$, respectively). 

Moreover, in \cref{fig:app:cvae-10,fig:app:cvae-50,fig:app:cvae-100,fig:app:cvae-250,fig:app:cvae-500,fig:app:cvae-full} we show both the distribution of $R_{\mathit{cond}}$ and $S_{\mathit{cond}}$ (compared to that of $R_{\mathit{uncd}}$ and $S_{\mathit{uncd}}$) across test formulae of a single random experiment, as well as the correlation between $R_{\mathit{uncd}}$ (resp., $S_{\mathit{uncd}}$) and the difference in average robustness $R_{\mathit{cond}} - R_{\mathit{uncd}}$ (resp., $S_{\mathit{cond}} - S_{\mathit{uncd}}$).  In such plots, a point above $0$ indicates that $R_{\mathit{cond}} > R_{\mathit{uncd}}$  (resp., $S_{\mathit{cond}} >S_{\mathit{uncd}}$), hence that the satisfaction of conditioning requirements is more robust (resp. higher) for trajectories generated by CVAE, w.r.t. those generated with SSA on the SIRS model.  Being $R\in [-1, 1]$ (resp. $S\in [0, 1]$), the maximum possible increase is $1 - R_{\mathit{uncd}}$ (resp. $1 - S_{\mathit{uncd}}$), i.e. the increase is linearly anti-correlated with the ground truth, hence producing triangular-like shapes in the mentioned figures (first and third subplot of each, starting from the left). Increasing the number of dimensions retained by stl2vec improves the performance until representations reach a dimensionality of $250$. Interestingly, this performance is better than that of full STL kernel embeddings: we believe this is due to the fact that implicit embeddings contain redundant and noisy information, which might make the learning process of CVAE more difficult.

\paragraph{Examples of requirements} imposed to the output of CVAE are reported in Table \ref{tab:app:requirements}. We recall that the SIRS model \cite{sirs} is an epidemiological model in which a population of $N$ individuals is assigned to compartments, namely Susceptible, Infected and Recovered. Letters S, I, R denote the number of people in each compartment, with S$+$I$+$R$=$N at each time. Each susceptible individual can be infected, and then recover; a recovered individual becomes susceptible again after a certain amount of time. In our experiments we set $N=100$, and simulate for $33$ timesteps. Transitions among states are governed by first order differential equations. We can give an intuitive description to some of the requirements listed in Table \ref{tab:app:requirements}, to highlight the significance of our methodology: property $1$ and $2$ assure that, from a certain time on, the number of infected individual (resp. recovered) will be higher than that of recovered individuals (resp. infected); property $5$ impose constraints on I and R, in an initial phase of the epidemic spreading; property $9$ describes instead the evolution on the number of recovered individuals, while property $10$ of both recovered and susceptible. We also report a property (namely number $12$) containing only Boolean operators.

\begin{table*}[h!]
\centering
\resizebox{0.85\linewidth}{!}{
\begin{tabular}{llll}
\toprule
{} & STL Requirement & $R_{\mathit{uncd}}$ / $R_{\mathit{cond}}$ & $S_{\mathit{uncd}}$ / $S_{\mathit{cond}}$ \\
\midrule
1 & $F_{[0, 6]} (G_{[0, 10]} (I\geq 20 \wedge R\leq 13))$ & -0.9509 / 0.8128 & 0.0267 / 0.9150 \\
\hline
2 & $F_{[0, 6]} (G_{[0, 10]} (I\leq 37 \wedge R\geq 48))$ & -0.4967 / 0.9951 & 0.2608 /1.0000 \\
\hline
3 & $F_{[0, 20]} (S\geq 60)$ & -0.2628 / 0.9029 & 0.3722 / 0.9500 \\
\hline
4 & $F_{[0, 2]} (G (I\leq 60 \wedge R\geq 37))$ & -0.6202 / 0.9573 & 0.1944 / 0.9990 \\
\hline
5 & $F_{[0, 12]} (I\geq 20 \wedge R\leq 28)$ & 0.0904 / 0.9834 & 0.5541 / 0.9930 \\
\hline
6 & $F_{[0, 2]} (G_{[0, 15]} (I\leq 60 \wedge R\geq 37))$ & -0.5941 / 0.9999 & 0.2137 / 1.0000 \\
\hline
7 & $F_{[0, 17]} (G_{[0, 7]} (I\geq 77 \wedge R\leq 71))$ & -0.6982 / 0.9992 & 0.1569 / 1.0000 \\
\hline
8 & $F_{[0, 6]} (I\leq 23 \wedge R\geq 62)$ & -0.7960 / 0.9983 & 0.1058 / 0.9990 \\
\hline
9 & $G(R\geq 46) \wedge (F_{0, 2} R\leq 67)$ & -0.8687 / 0.8655 & 0.0630 / 0.9420\\
\hline
10 & $G_{[0, 17]} ((G_{[0, 14]} R\geq 14) \wedge R\leq 52) \wedge (F_{[0, 7]} (G_{[0, 22]} S\leq 50) \wedge S\leq 20)$ & -0.9274 / 0.7723 & 0.0348 / 0.8930 \\
\hline
11 &  $G_{[0, 8]} ((F_{[0, 6]} I\geq 85) \wedge I\leq 60) \wedge (G_{[0, 12]} (F_{[0, 15]} S\leq 40))$ & -0.9061 / 0.9685 & 0.0474 / 0.9960 \\
\hline
12 & $I\geq 42 \wedge (S\leq 23 \wedge S\geq 10)$ & -0.8902 / 0.4700 & 0.0590 / 0.7450 \\
\hline
13 &  $F(\neg R\geq 19) \wedge (S\leq \vee F(G_{[0, 13]} (I\leq 53 \wedge I\geq 80)))$ & -0.8725 / 0.9927 & 0.0641 / 0.9990 \\
\bottomrule
\end{tabular}
}
    \vspace{0.1cm}
\caption{Examples of requirements for the SIRS model, to constrain the generation of trajectories of the CVAE model.}
\label{tab:app:requirements}
    \vspace{0.5cm}
\end{table*}

\begin{figure*}[h!]
    \centering 
    \includegraphics[width=0.23\textwidth]{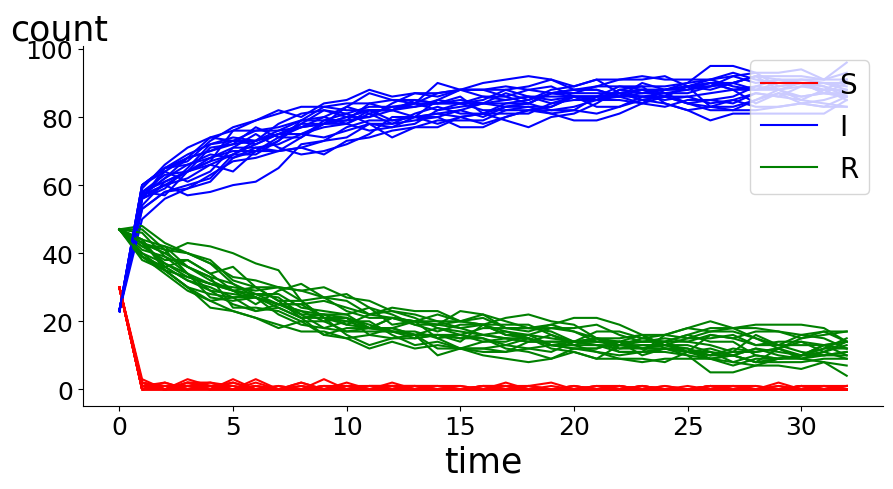}
    \includegraphics[width=0.23\textwidth]{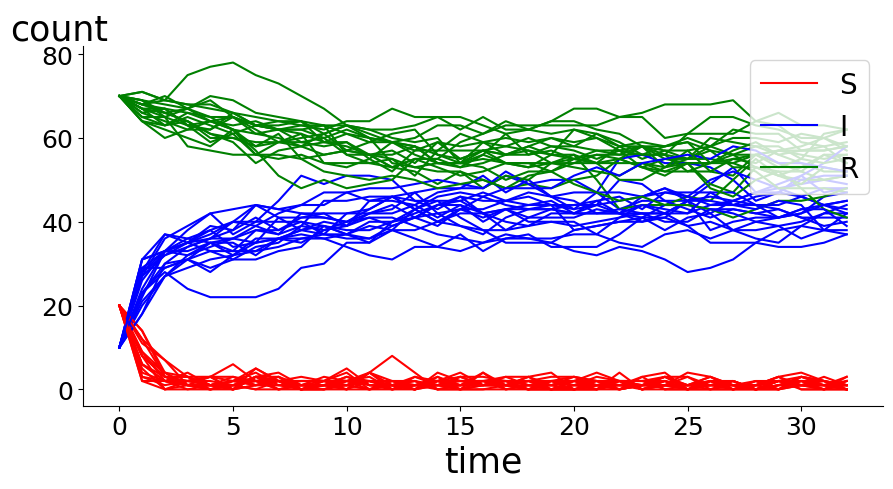}
    \includegraphics[width=0.23\textwidth]{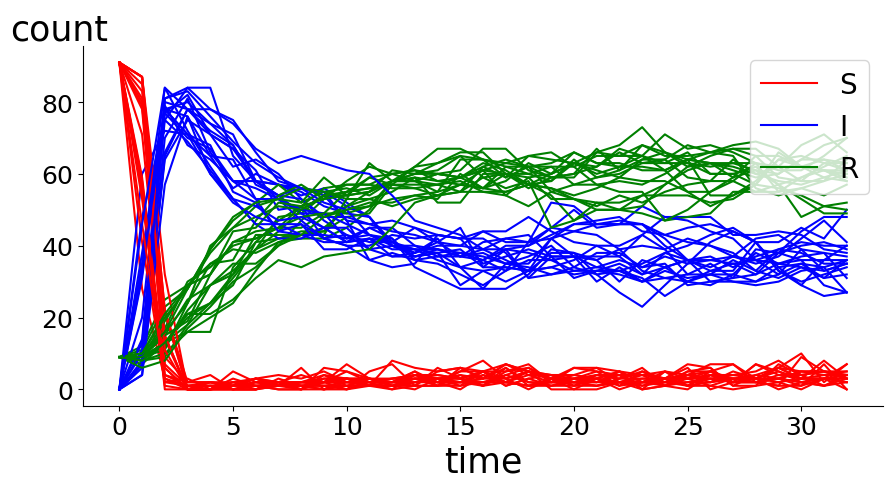}
    \includegraphics[width=0.23\textwidth]{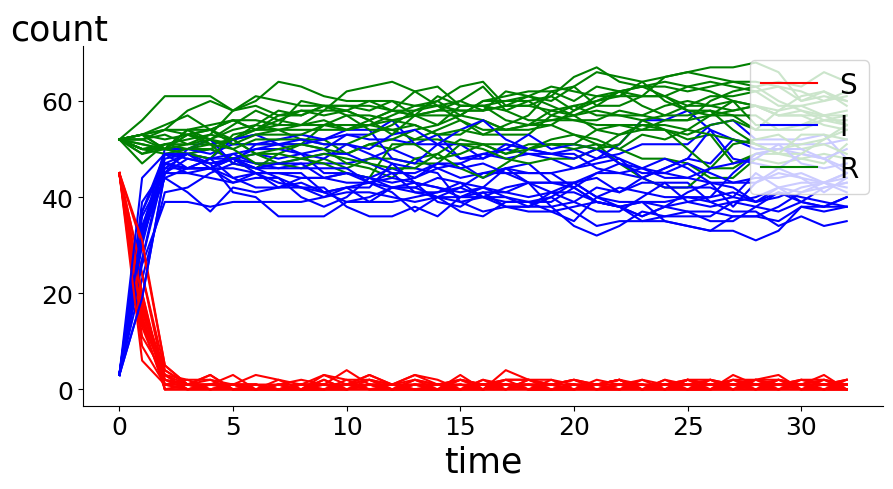}
    \vspace{0.1cm}
    \caption{Examples of trajectories sampled form the SIRS model, starting from different initial states.}
    \label{fig:app:sirs}
    \vspace{0.5cm}
\end{figure*}

In Figure \ref{fig:app:sirs} we graphically show trajectories sampled from the SIRS model, simulated via SSA. 

\begin{figure*}[h!]
    \centering 
    \includegraphics[width=0.33\textwidth]{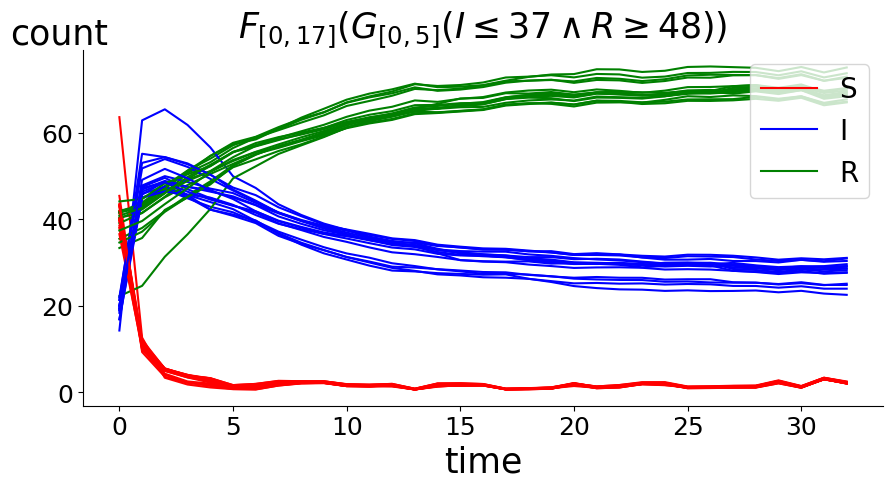}
    \includegraphics[width=0.33\textwidth]{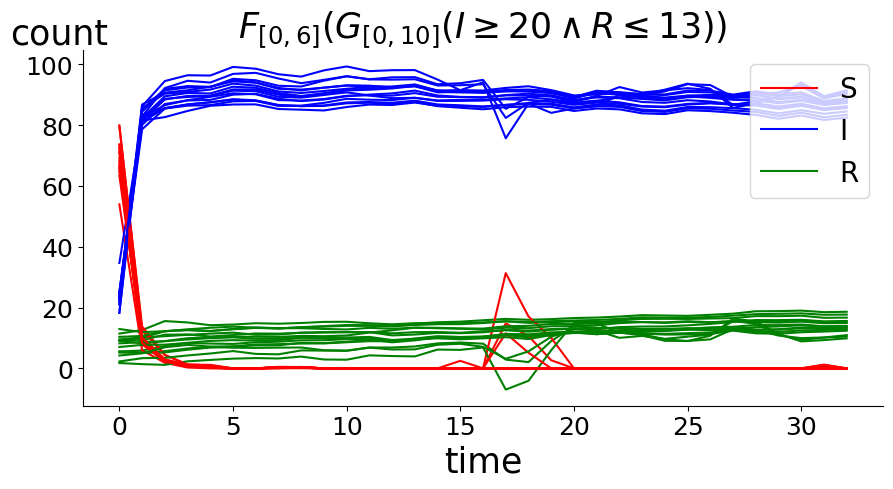}
    \includegraphics[width=0.33\textwidth]{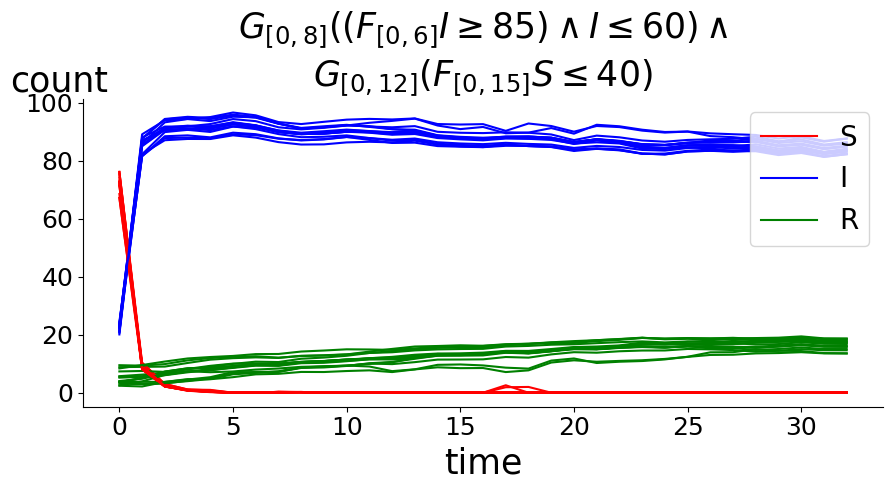}
    \includegraphics[width=0.33\textwidth]{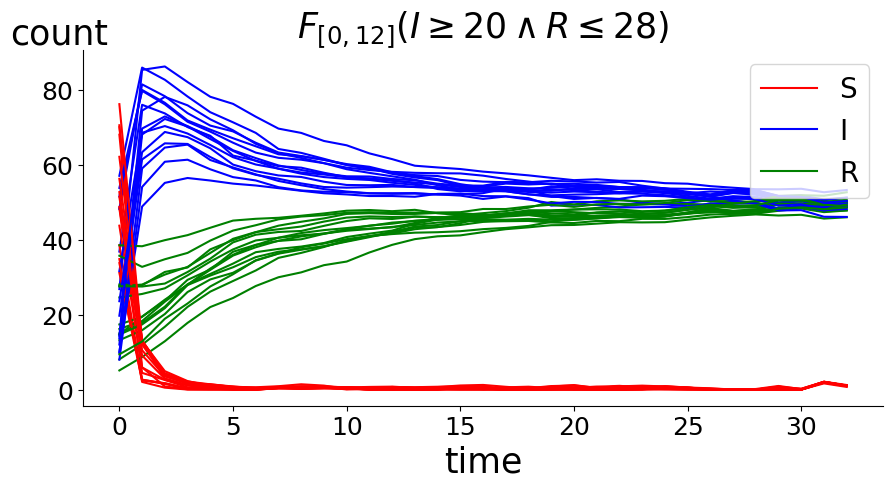}
    \includegraphics[width=0.33\textwidth]{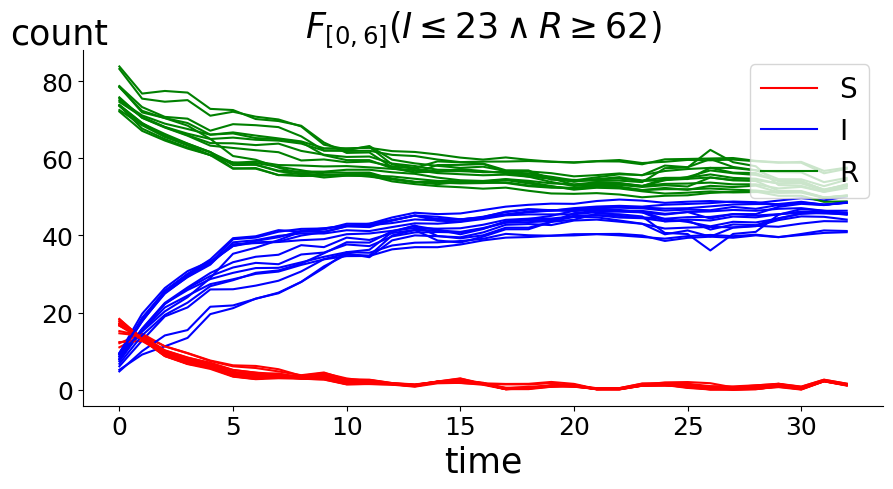}
    \includegraphics[width=0.33\textwidth]{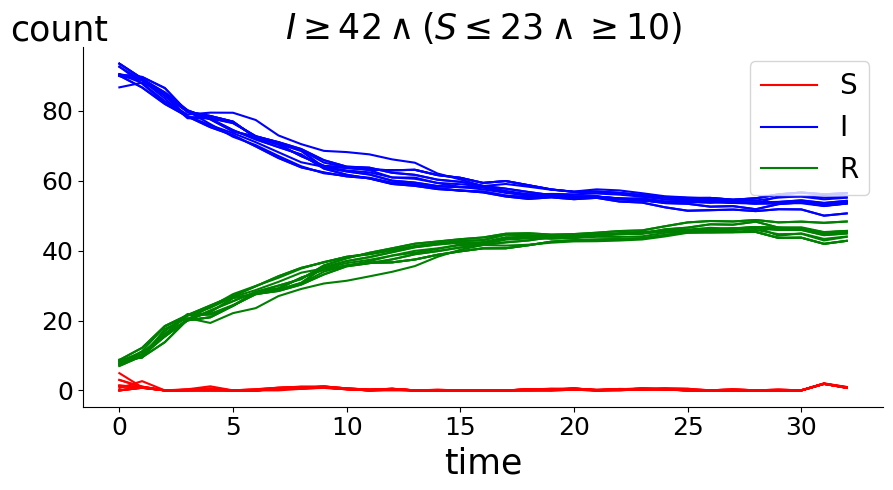}
    \vspace{0.1cm}
    \caption{Examples of trajectories generated by our CVAE model, imposing different STL requirements (reported as titles of the subplots) in the form of stl2vec embeddings of dimension $250$.}
    \vspace{0.5cm}
    \label{fig:app:sat-test}
\end{figure*}

Moreover, in Figure \ref{fig:app:sat-test}, we show a sample of trajectories generated by the CVAE model, when conditioned on the requirement specified as title of each subplot (only trajectories satisfying the conditioning properties are shown). From Figure \ref{fig:app:sat-test}, we can assess the similarity of the distribution of trajectories sampled from the SIRS model and those generated by our CVAE. 

\begin{figure*}[h!]
    \centering 
    \includegraphics[width=0.2\textwidth]{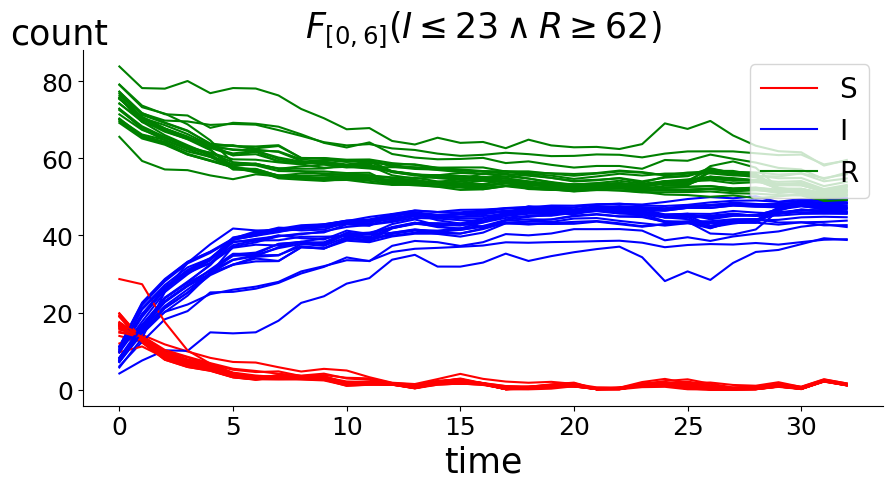}
    \includegraphics[width=0.2\textwidth]{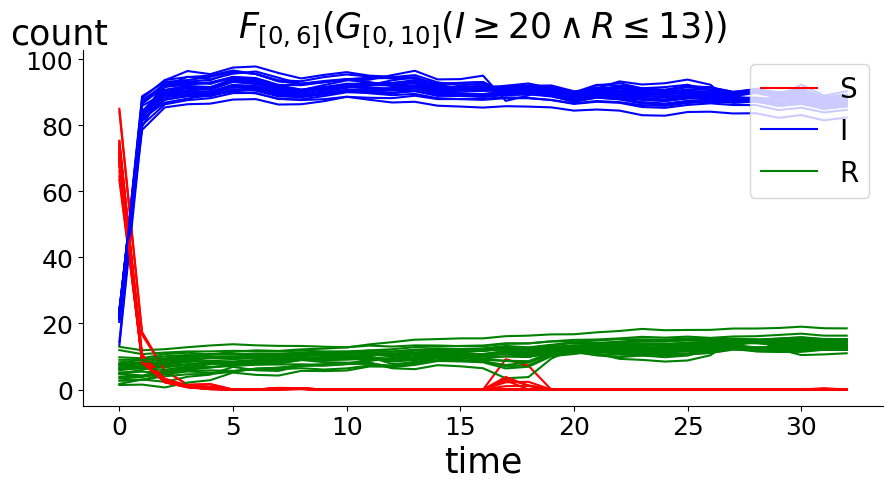}
    \includegraphics[width=0.2\textwidth]{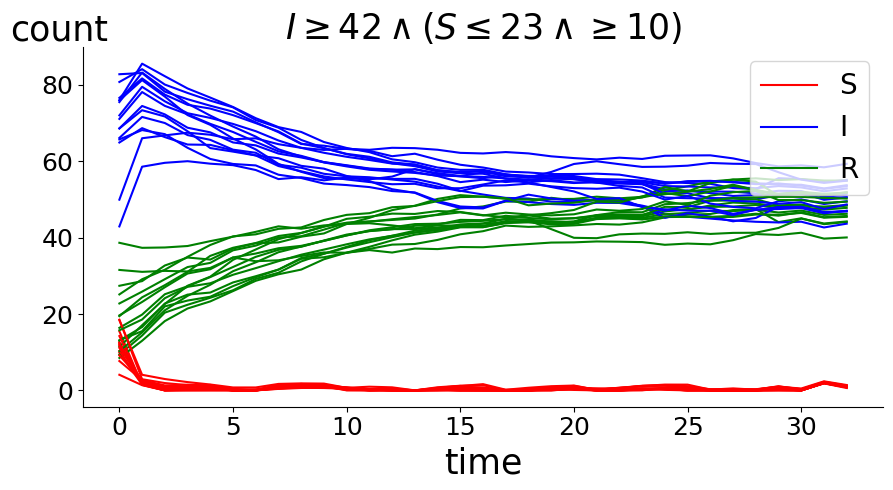}
    \includegraphics[width=0.2\textwidth]{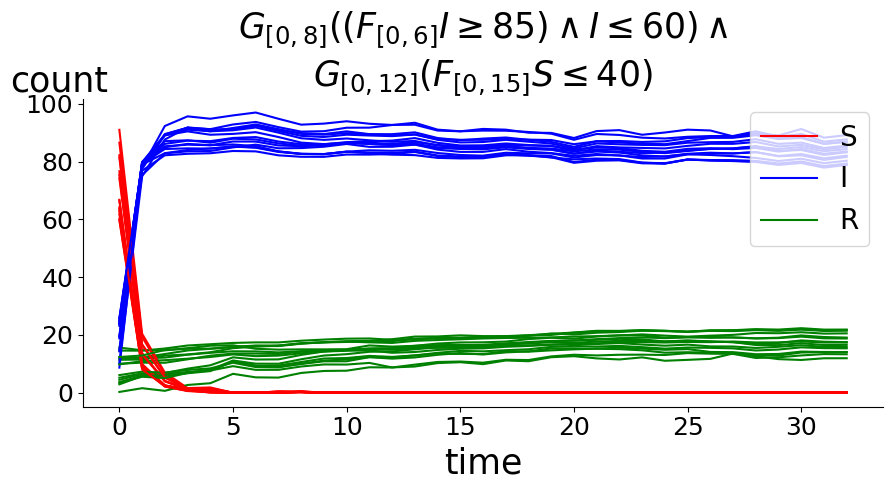}
    \vspace{0.1cm}
    \caption{Examples of trajectories generated by our CVAE model, not satisfying the input STL requirements, reported as titles of each subplot (provided in the form of stl2vec embeddings of dimension $250$).}
    \vspace{0.5cm}
    \label{fig:app:unsat-test}
\end{figure*}

Finally, we show in Figure \ref{fig:app:unsat-test} trajectories generated by our CVAE model which do not satisfy the input STL requirement (reported as title of each subplot). From Figure \ref{fig:app:unsat-test}, we can assess that generated trajectories are still distributed like those of the SIRS model (see Figure \ref{fig:app:sirs}), confirming that the autoencoder has learnt a meaningful representation of SIRS trajectories. 

